\documentclass{egpubl}
\JournalSubmission

\usepackage{xcolor}
\newcommand{\dc}[1]{}
\newcommand{\op}[1]{}

\usepackage[utf8]{inputenc}
\usepackage[T1]{fontenc}
\usepackage{dfadobe}
\usepackage{cite}
\usepackage{url}

\BibtexOrBiblatex
\electronicVersion
\PrintedOrElectronic

\usepackage{graphicx}

\usepackage{egweblnk}
\usepackage{booktabs}
\usepackage{amsmath, amsfonts, amssymb}
\usepackage{nicefrac}
\usepackage{microtype}
\usepackage{subcaption}
\usepackage{multirow}
\usepackage[ruled,vlined]{algorithm2e}
\usepackage{enumitem}
\usepackage{pifont}
\usepackage{etoolbox}
\makeatletter
\patchcmd{\@maketitle}{\newpage}{\par}{}{}
\makeatother

\newcommand{\method}{LumiVid}

\title[HDR Video Generation via Latent Alignment]%
{HDR Video Generation via Latent Alignment with Logarithmic Encoding}

\author[N. Ken Korem et al.]{%
\begin{tabular}[t]{@{}c@{}}%
\let\\=\tabularnewline
Naomi Ken Korem$^{1}$ \quad Mohamed Oumoumad$^{1,2}$ \quad Omer Hagage$^{1}$ \quad Amir Gam$^{1}$\\[2pt]
Matan Ben Yosef$^{1}$ \quad Harel Cain$^{1}$ \quad Urska Jelercic$^{1}$ \quad Ofir Bibi$^{1}$ \quad
Yaron Inger$^{1}$\\[2pt]  Or Patashnik$^{3}$ \quad Daniel Cohen-Or$^{3}$\\[6pt]
$^{1}$Lightricks \quad $^{2}$Gear Productions \quad $^{3}$Tel Aviv University%
\end{tabular}}

\begin{document}

\teaser{\noindent
\begin{minipage}{\textwidth}
    \centering
    \setlength{\tabcolsep}{1pt}
    \begin{tabular}{c cc c cc}
    & \multicolumn{2}{c}{\small\textbf{Image}} && \multicolumn{2}{c}{\small\textbf{Video}} \\
    \raisebox{12pt}{\rotatebox{90}{\scriptsize\textbf{SDR (Input)}}} &
    \includegraphics[width=0.24\textwidth]{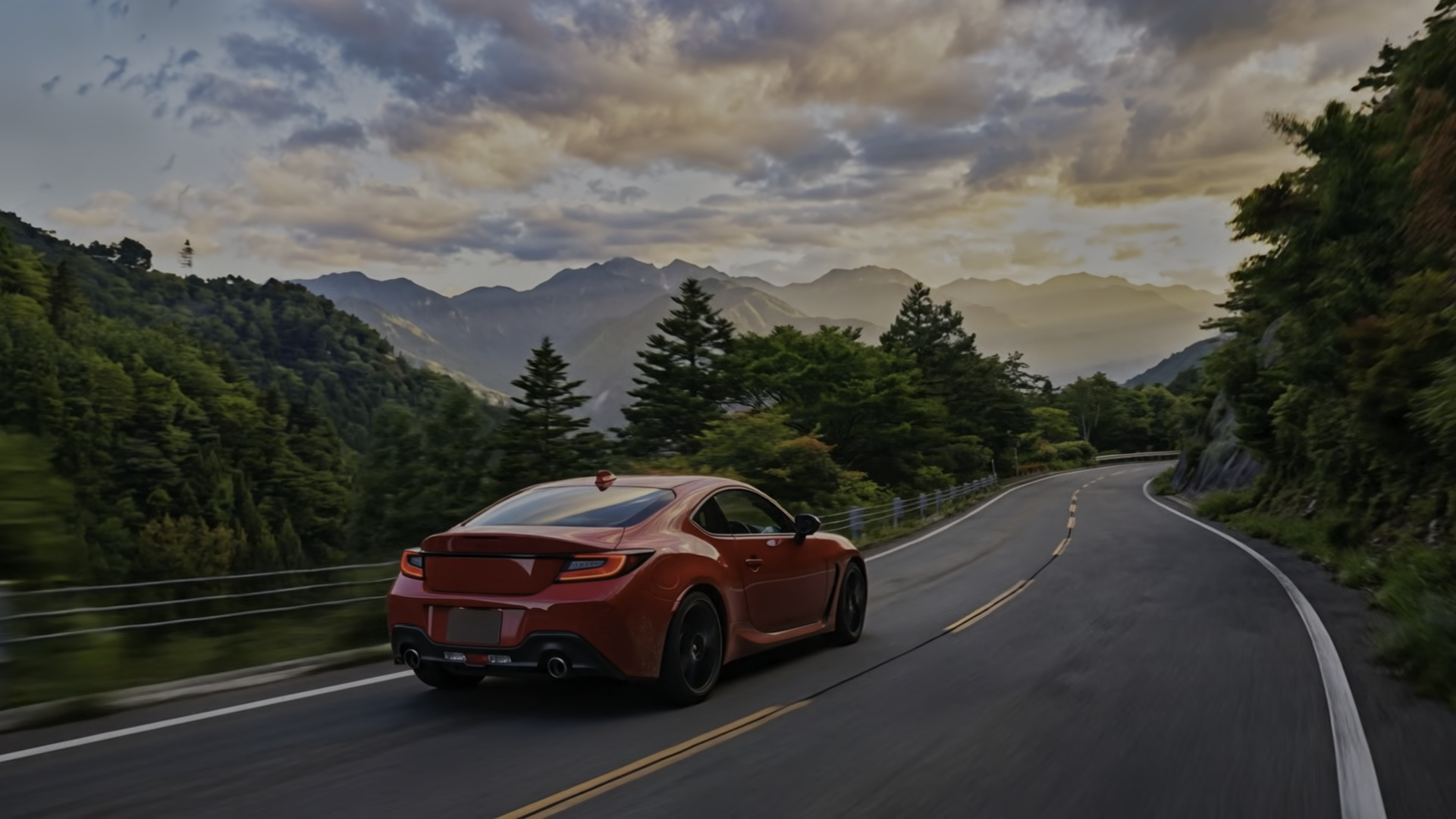} &
    \includegraphics[width=0.24\textwidth]{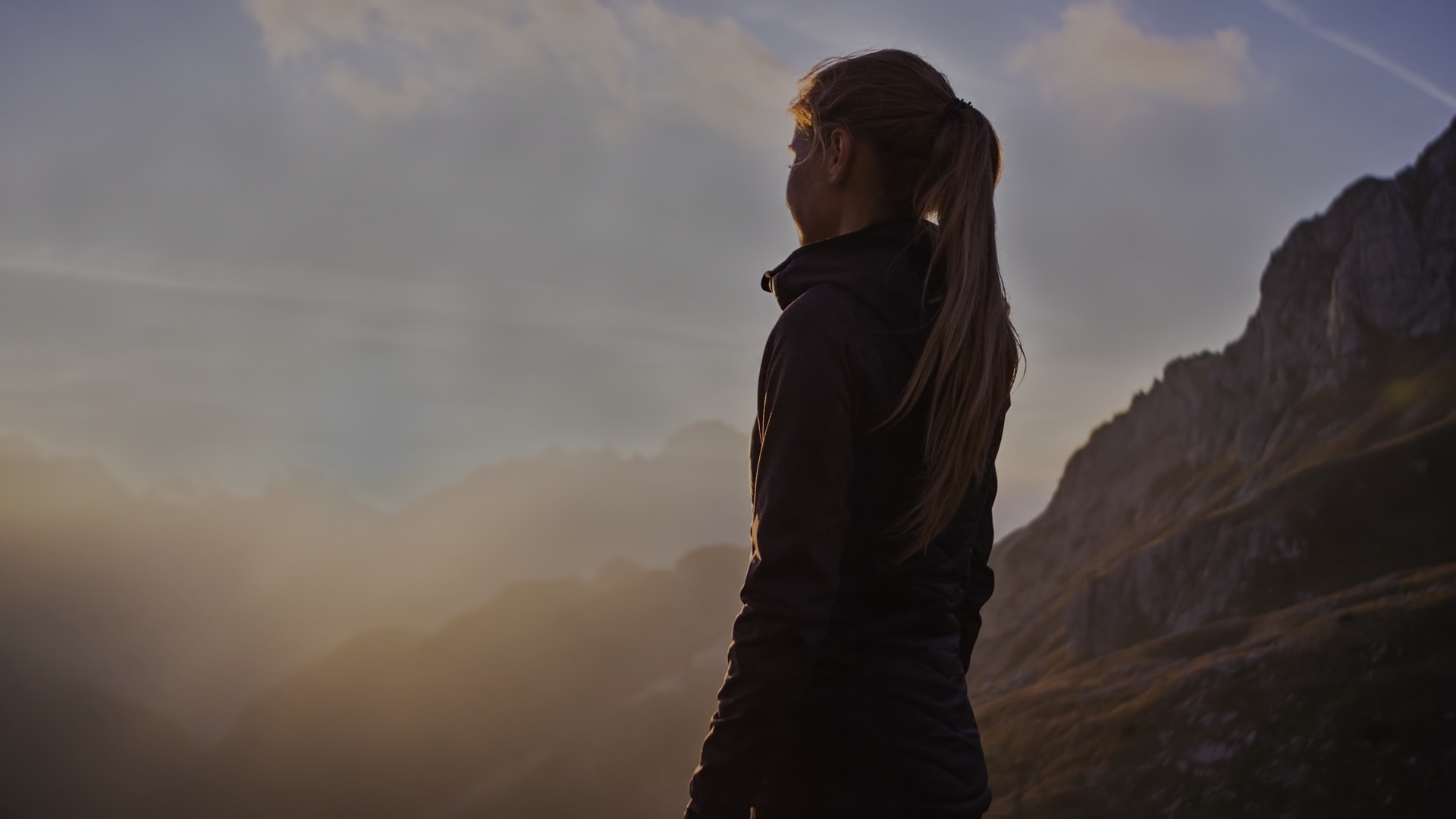} &&
    \includegraphics[width=0.24\textwidth]{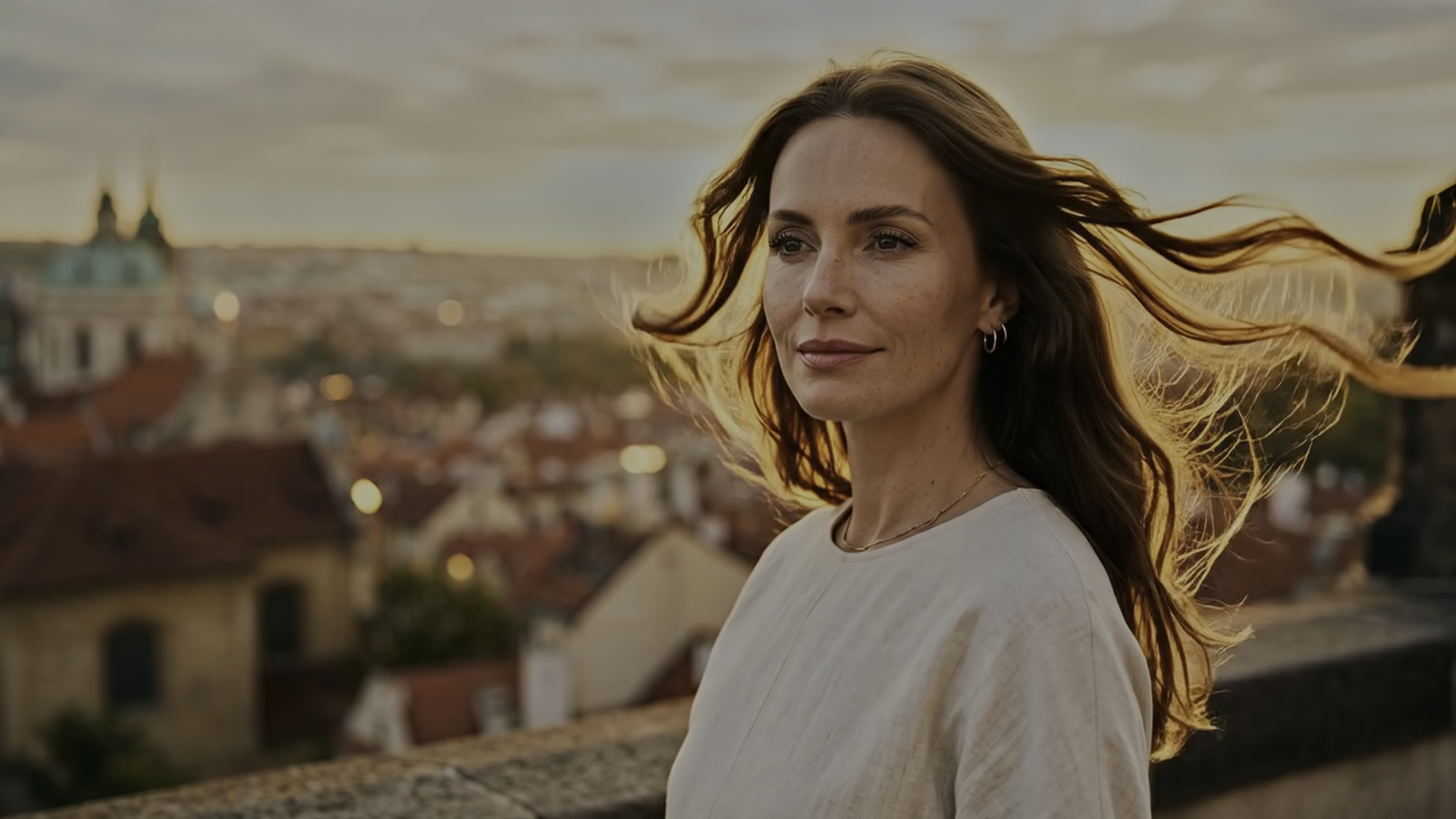} &
    \includegraphics[width=0.24\textwidth]{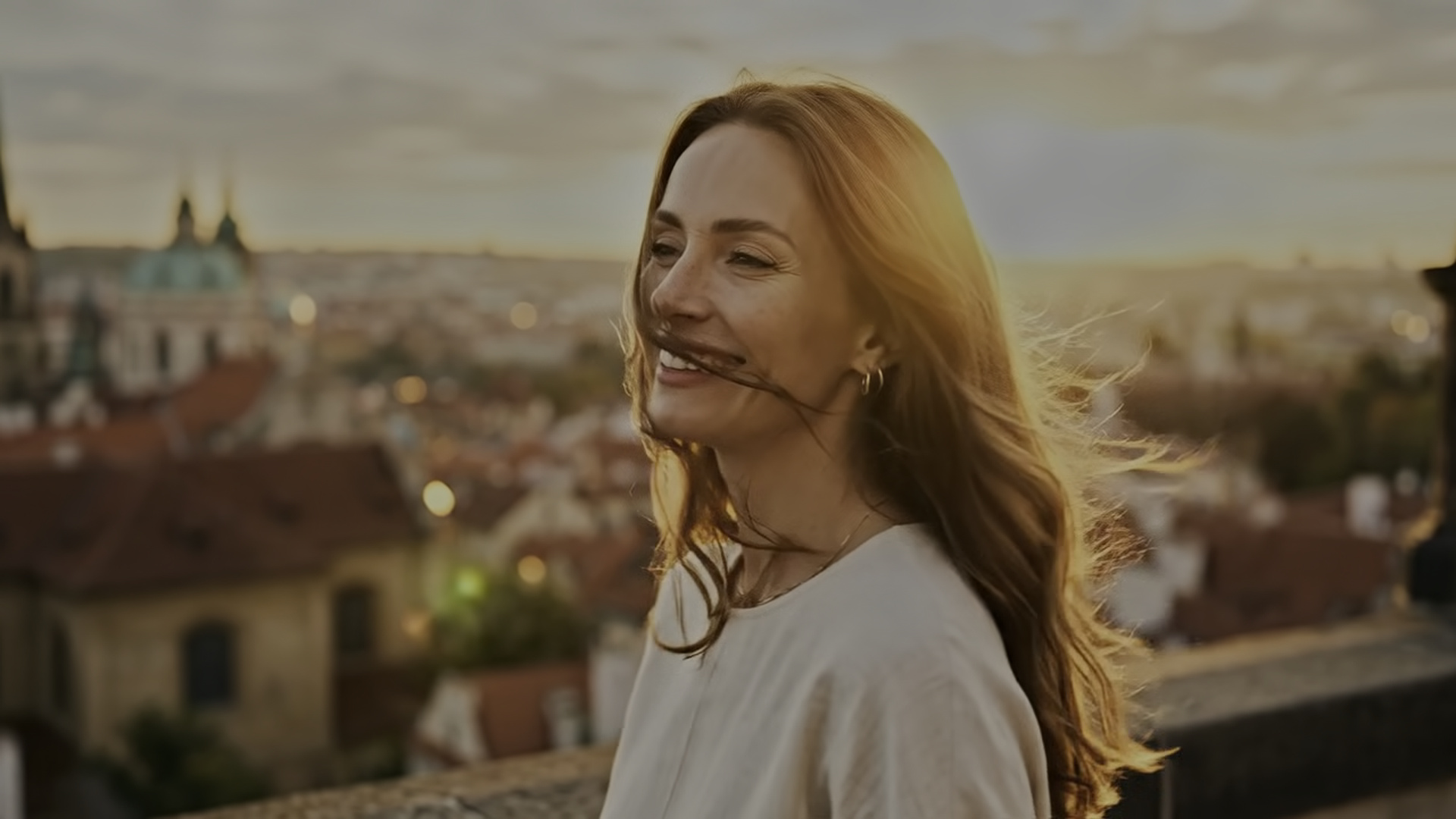} \\
    \raisebox{12pt}{\rotatebox{90}{\scriptsize\textbf{HDR (Ours)}}} &
    \includegraphics[width=0.24\textwidth]{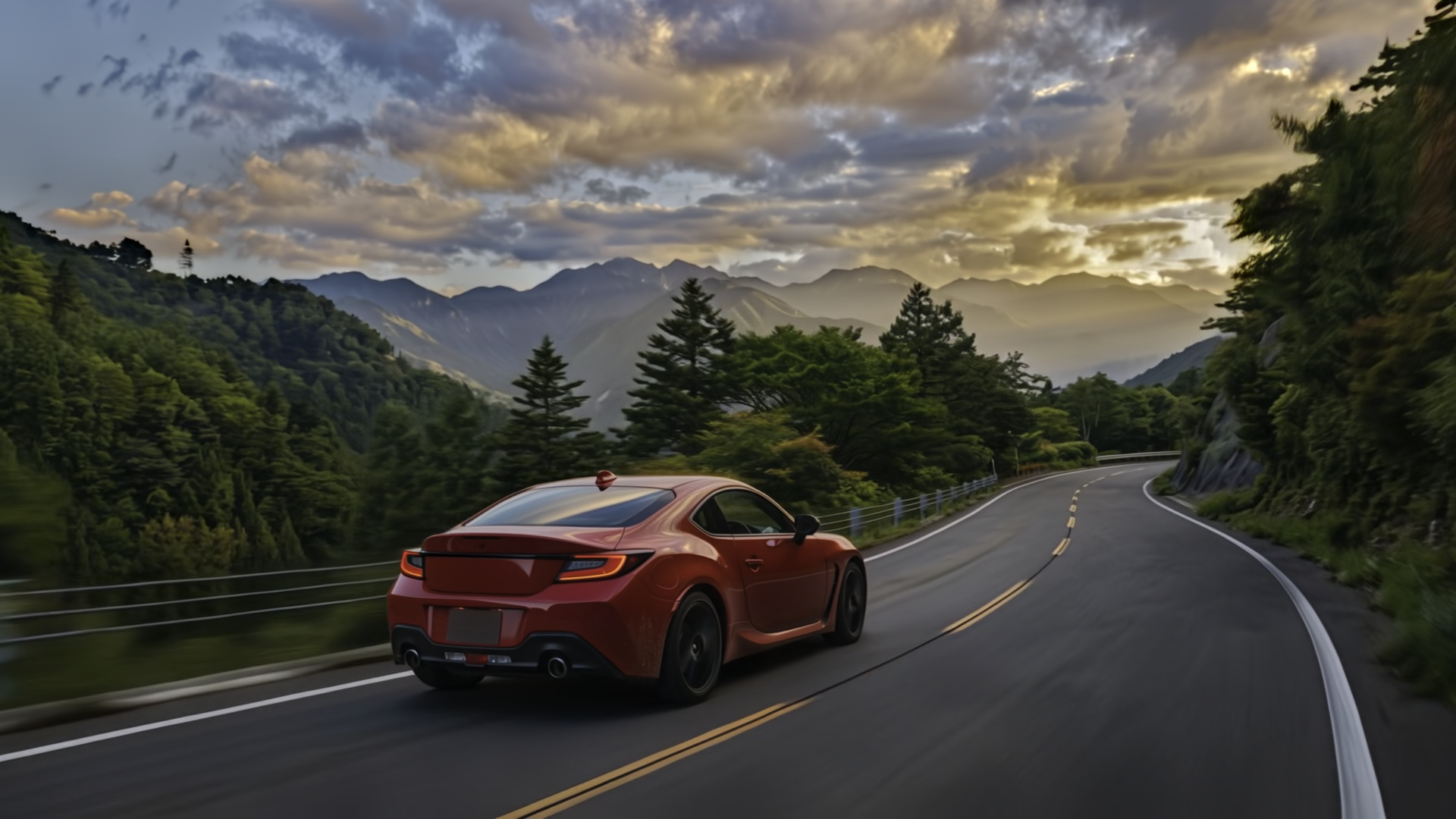} &
    \includegraphics[width=0.24\textwidth]{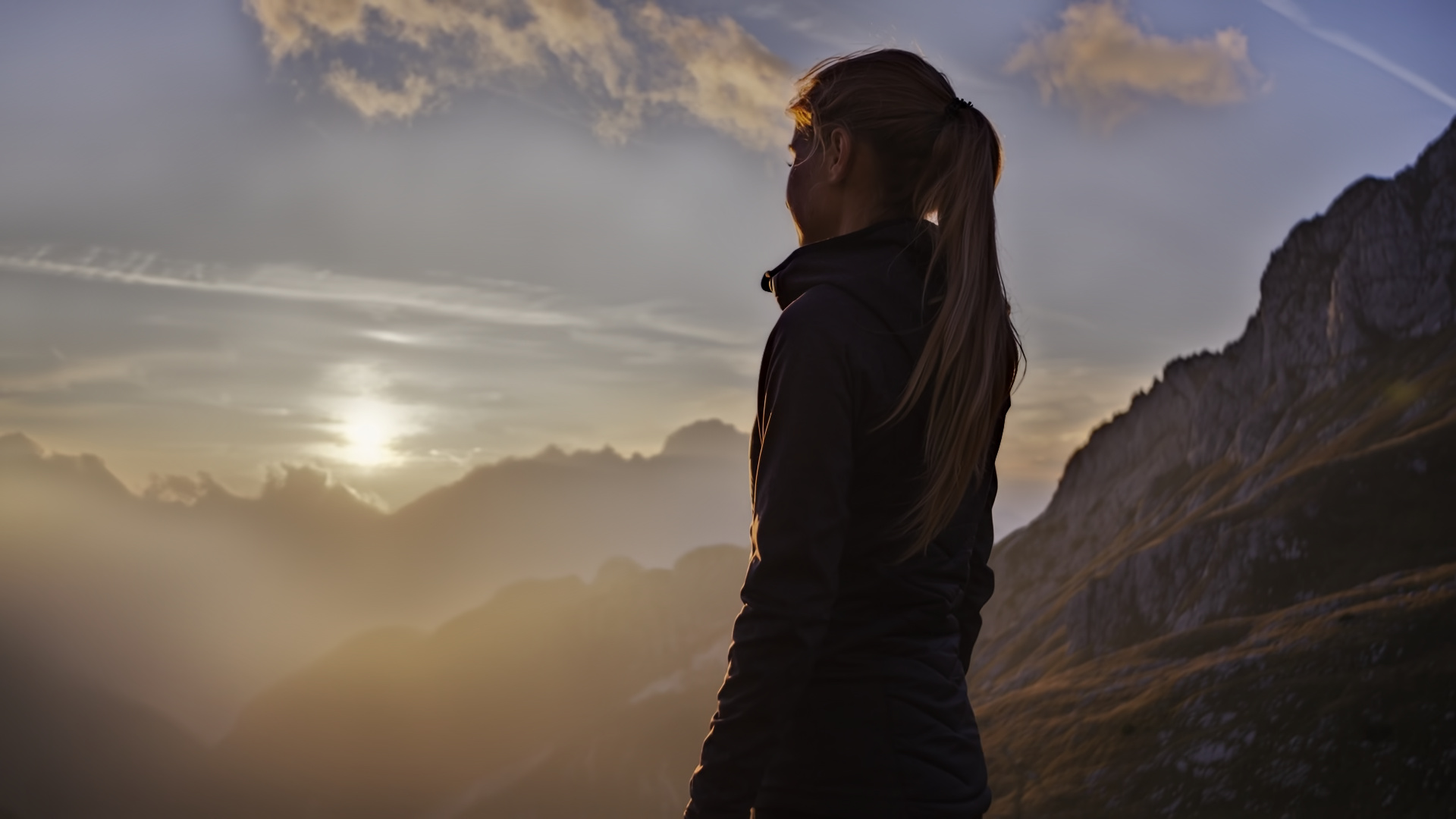} &&
    \includegraphics[width=0.24\textwidth]{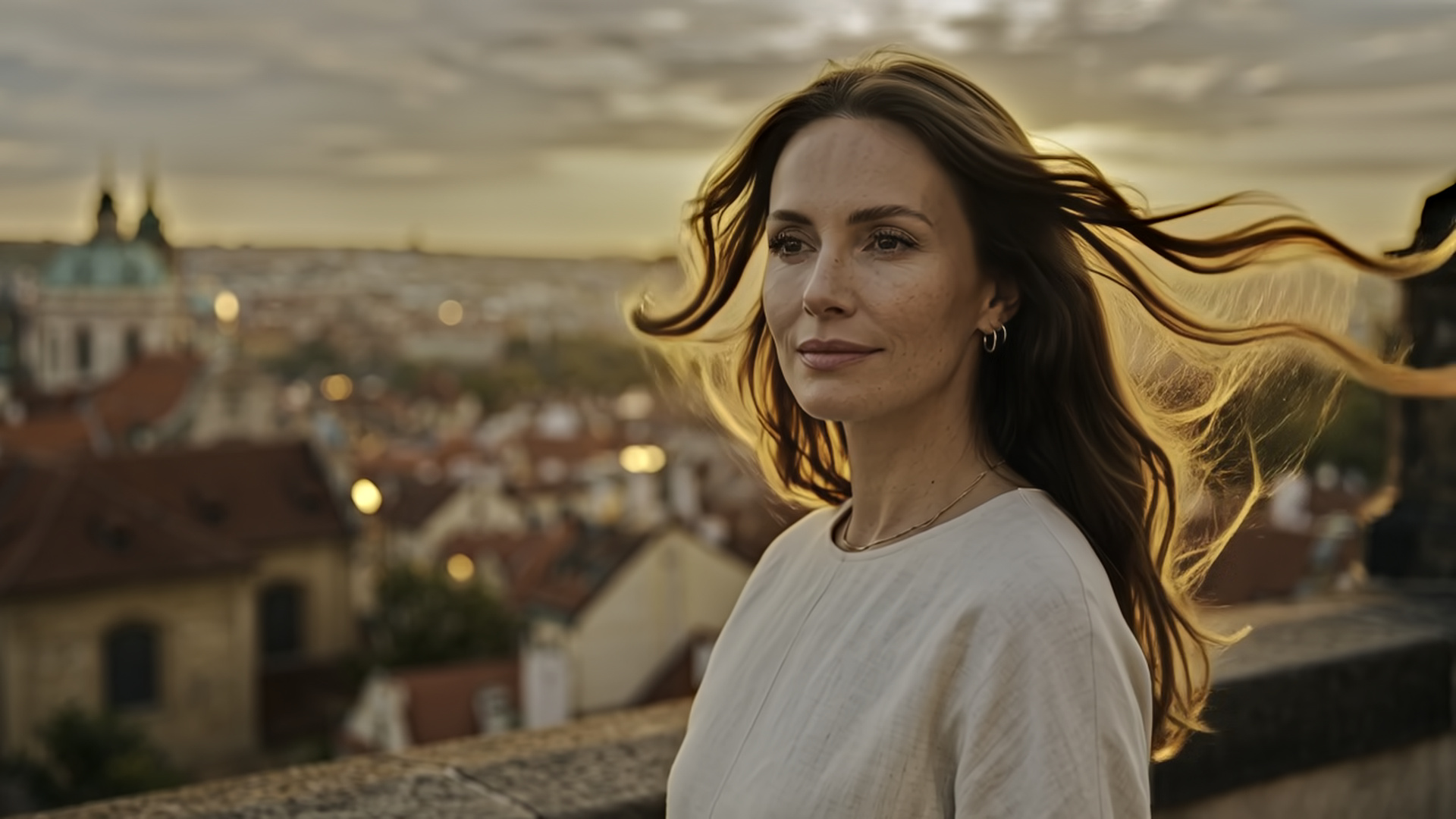} &
    \includegraphics[width=0.24\textwidth]{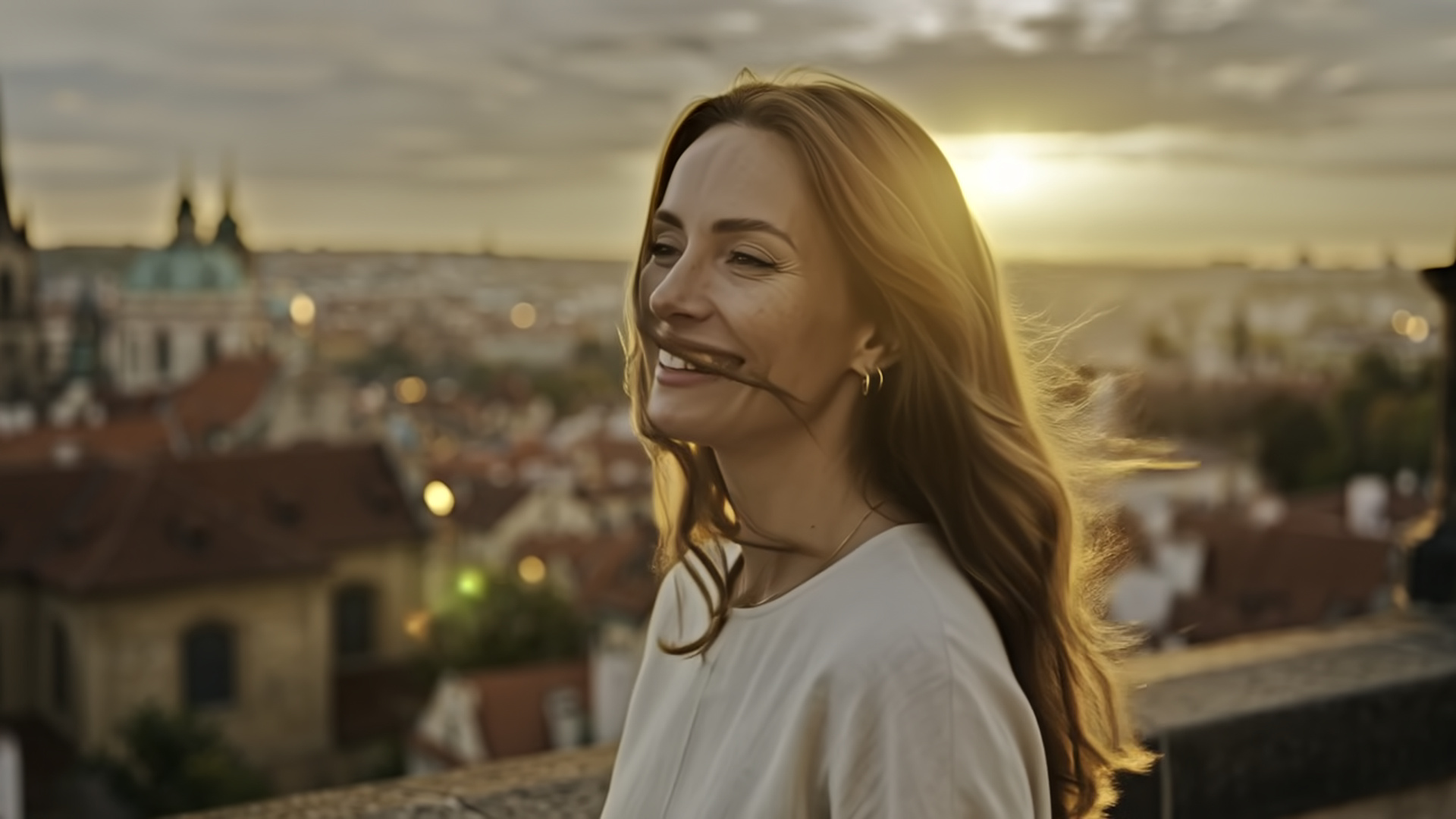} \\
    \end{tabular}
    \captionof{figure}{\textbf{\method{}: HDR video generation from SDR input.}
Given an input SDR image or video (top), our method produces HDR (bottom).
Both rows use Mertens fusion of the same frame at multiple EVs.
The image panel shows a mountain-road still at EV $-2$ and a sunset still at EV $0$.
The video panel shows two times from a rooftop clip at EV $-1$ and EV $-3$.}
    \label{fig:teaser}
\end{minipage}
\vspace{0.6em}
}
\maketitle

\begin{abstract}
High dynamic range (HDR) video captures continuous scene radiance essential for professional color grading and post-production workflows, yet generative video diffusion models remain bound to 8-bit standard dynamic range (SDR) formats. Existing adaptation methods either alter the VAE latent space or rely on auxiliary inputs like luminance masks, text prompts, or external reference images. However, relying on auxiliary inputs introduces manual overhead, makes the exact prompt hard to guess, and yields less stable results that often fail to remain consistent with the original footage. We present LumiVid, a framework that generates HDR video directly from a single SDR input, without requiring masks, text prompts, or external reference images. Key to our method is mapping the HDR video into a logarithmic color space representation, aligning high-magnitude radiance with the existing latent space of a frozen video VAE without losing luminance details. Fine-tuning a dual-canvas LoRA adapter on just ~300 synthetic paired clips enables data-efficient SDR-to-HDR translation while preserving temporal consistency. To prevent shortcut pixel copying during training and force the model to infer missing radiance and details from its learned priors, we introduce camera-mimicking condition degradations on the SDR reference input. Extensive evaluations show that LumiVid outperforms state-of-the-art baselines by +5.5 dB PU21-PSNR on cinema-grade ARRI footage, producing high-fidelity half-float EXR video with exceptional re-exposure latitude across extreme exposure pulls.
\end{abstract}

\section{Introduction}

High dynamic range (HDR) imagery captures scene radiance over a wide range of intensities,
enabling faithful representation of highlights, shadows, and subtle illumination effects
that are lost in standard imagery.
This capability is essential for realistic rendering, post-production, professional color grading,
and emerging display technologies.

Professional film and visual effects workflows rely on \textit{scene-referred data}: a format where pixel values represent the actual physical light captured by the camera sensor. In this system, diffuse white (the light bouncing off a flat, non-shiny surface like a white sheet of paper) is set to a baseline value of 1.0. Brighter light isn't capped at 1.0. Specular glare, car headlights, and direct sunlight extend unbounded into values of 10, 100, or even thousands.
Holding onto real-world light values makes post-production reversible. Colorists adjust exposure in stops (the standard measurement of camera exposure, where each stop doubles or halves the light energy ($\text{Pixel}_{\text{new}} = \text{Pixel} \times 2^{\text{stops}}$). Since the video file retains data far above 1.0, dropping the exposure by a few stops reveals crisp details inside a glowing window instead of leaving a blown-out white blob. Shadow recovery also avoids color banding because dark tones were saved as continuous light levels. This linear math is also essential for realistically blending 3D CGI with live-action footage.
Only at the end of the workflow is the footage converted into \textit{display-referred data}, a set of image instructions tailored specifically to the physical limits of a monitor. For standard SDR screens, this meant squashing all light into a rigid 8-bit scale from 0 to 255, where 255 is the absolute brightest white the screen can produce. Modern HDR screens can display 10-bit or 12-bit color depths, but even they require a final display transform to map unbounded light values onto what the screen hardware can physically output.

This compression explains why video AI models struggle to generate native HDR. Video AI models are trained mostly on web videos stored in 8-bit SDR formats, where a piece of white paper and the sun are both squashed into the exact same ceiling of 255. These models learn exclusively from flattened, monitor-bound data, they lack an internal knowledge of how real physical light behaves beyond standard display limits.

A natural strategy for extending diffusion models to HDR is to leverage the visual structure,
semantics, and temporal dynamics already learned by pretrained video models~\cite{hacohen2026ltx}.
One common approach is to train a dedicated HDR encoder or fine-tune VAE decoders to emit linear radiance~\cite{wang2025lediff}, creating an HDR-specific latent space. However, altering or replacing the encoder shifts the underlying latent manifold, forcing the downstream diffusion model to adapt to an unfamiliar space. This dependency demands massive HDR datasets and makes training complex and computationally expensive. 

Drastically altering the latent space breaks compatibility with the model's pretrained features. Keeping the HDR representation aligned with the model's native latent space allows for data-efficient adaptation while preserving existing capabilities, such as video editing, inpainting, and controlled generation, directly in the HDR domain without needing to retrain those tools.

DiffHDR ~\cite{yu2026diffhdr} has begun to explore aligning HDR with the model's native latent space, but relies on external prompts and masks to synthesize missing highlights. In contrast, our method performs direct SDR-to-HDR translation conditioned only on the input video itself.
We map scene-linear HDR into ACEScct, a standardized cinematic logarithmic color space (see Appendix~\ref{app:color} for color space fundamentals).
Encoding this log video with the frozen SDR VAE yields latents already aligned with the
pretrained manifold, allowing the generator to emit log-encoded HDR with lightweight fine-tuning
and without a new encoder.

We train an in-context LoRA on a dual-canvas architecture,
where SDR tokens condition the denoising of ACEScct HDR target tokens.
 300 paired clips are sufficient:
the LoRA learns a targeted dynamic-range conversion. To encourage the model to learn true HDR generation from an SDR source, we make the task even more challenging. Since SDR video already lacks detail in extreme luminance regions, we apply camera-mimicking degradations, such as aggressive clipping, heavy compression, and selective blur, to strip away even more information from the highlights and shadows. Combined with varied exposure shifts during training, this forces the network to rely on its internal priors to actively synthesize continuous, high-magnitude radiance rather than copying flat SDR pixels, eliminating the need for text prompts or luminance masks.

We evaluate on professional ARRI cinema footage 
and on 66 YouTube in-the-wild SDR videos. Both evaluation benchmarks comprise real cinematic footage featuring complex actor motion, dynamic lighting transitions, and non-rigid moving objects.
Our findings show that HDR capabilities can be enabled in existing video models with high data efficiency,
without architectural redesign or full retraining.

Our key contributions are:
\begin{enumerate}[leftmargin=*]
\item SDR-to-HDR generation: We generate HDR video directly from a pretrained video model using only the SDR video as an input, requiring no luminance masks, text prompts, or external reference images.

\item Logarithmic color space alignment and data-efficient adaptation: Mapping HDR video into a logarithmic color space representation aligns high-magnitude radiance latents with the existing latent space of a frozen SDR VAE without losing intricate luminance details. Framing the process as a dynamic-range translation enables fine-tuning a dual-canvas LoRA adapter on just $\sim$300 synthetic paired clips while preserving temporal consistency. Despite training exclusively on synthetic camera orbits, our model generalizes seamlessly to real-world footage containing dynamic actor motion and complex scene transitions at inference.

\item Camera-mimicking degradations and exposure augmentations: We combine contrast clipping, compression, highlight/shadow blur, and joint exposure shifts, alongside an asymmetric clamp ($[0, 0.55]$ on the SDR reference input vs. $[0, 1.0]$ on the HDR target), to prevent shortcut pixel copying during training, forcing the model to infer extreme high-magnitude highlights and deep shadow details from its learned priors. 

Video results, exposure pulls, and an overview are at \url{https://hdr-lumivid.github.io/}. 

\end{enumerate}

\section{Related Work}
\label{sec:related_work}

\subsection{Classical SDR-to-HDR Reconstruction}
\label{sec:rw_sdr_to_hdr}

The challenge of capturing and manipulating high dynamic range content has long been a focal point of computational photography. Early foundations were built on multi-exposure HDR recovery~\cite{debevec1997recovering} and photographic tone mapping for display~\cite{reinhard2002photographic}. A landmark in forward tone-mapping (HDR-to-SDR) was the work of Fattal et al.~\cite{fattal2002gradient}, who pioneered gradient-domain dynamic range compression. By selectively attenuating large luminance gradients (global contrast) while preserving small gradients (local detail), they demonstrated that integrating the modified field via a Poisson equation could compress dynamic range without introducing edge artifacts. This was further refined by Farbman et al.~\cite{farbman2008edge}, who introduced edge-preserving multiscale decompositions based on Weighted Least Squares (WLS) optimization, providing a robust framework for detail manipulation that effectively eliminated the halo artifacts common in non-linear filtering.

While these classical operators focused on compressing scene-linear HDR for bounded displays, the complementary task of Inverse Tone Mapping (ITM) aims to expand SDR content back into unbounded radiance. ITM evolved from early heuristic expansion operators~\cite{banterle2007framework,rempel2007ldr2hdr} to modern deep CNN-based reconstruction architectures that reverse the camera pipeline or estimate linear radiance from single exposures~\cite{eilertsen2017hdrcnn,marnerides2018expandnet,liu2020singlehdr}. A significant step toward modern generative approaches was taken by Marnerides et al.~\cite{marnerides2021deep}, whose work on Deep HDR Estimation moved beyond deterministic pixel mapping by utilizing generative detail hallucination for saturated regions. More recent architectures improve efficiency via frequency-domain processing~\cite{xu2024dft} or leverage specialized sensing modalities such as event cameras~\cite{yang2023hdrev}.

A parallel line converts SDRTV to display-referred HDRTV (PQ, BT.2020) rather than scene-linear radiance. HDRTVNet~\cite{chen2021hdrtvnet} introduced a practical SDR-to-HDRTV pipeline and the HDRTV1K benchmark; HDRTVNet++~\cite{chen2309towards} extends it with adaptive global color mapping, local enhancement, and highlight refinement. HDRTVDM~\cite{guo2023hdrtvdm} adds degradation models and HDRTV4K. We use HDRTVNet++ as a per-frame baseline, but it targets 10-bit HDR display delivery rather than 16-bit scene-linear EXR files.

Extending to video, prior work focuses on temporal alignment and consistency using optical flow~\cite{kalantari2019hdr}, coarse-to-fine alignment~\cite{chen2021deephdr}, or specialized sensing modalities~\cite{yang2023hdrev}. Recent methods further address robustness to illumination changes~\cite{ye2024vitm} and real-time performance~\cite{xu2024hdrflow}, supported by large-scale HDR video datasets~\cite{shu2024realhdrv}. 

These approaches treat SDR-to-HDR as a reconstruction problem, learning dedicated mappings under restricted imaging assumptions. In contrast, we treat HDR generation through the lens of distribution alignment: pretrained video diffusion models already capture rich priors over illumination, highlights, and shadows. The key challenge is to expose and unlock this knowledge by aligning HDR input/output distributions with the latent representations expected by the frozen model.

\subsection{Generative \& Diffusion-Based Image HDR Synthesis}
\label{sec:rw_hdr_diffusion}

Recent research has shifted toward leveraging generative models to synthesize missing radiance. Early works explored unsupervised HDR synthesis via GlowGAN, a StyleGAN variant~\cite{wang2023glowgan} or applied conditional diffusion to multi-exposure image deghosting~\cite{yan2023diffhdr} and multi-exposure reconstruction~\cite{guan2024hdrvdiff}. Frame-level frameworks like BracketDiffusion~\cite{bemana2025bracket} employ exposure-aware modeling to ensure consistency across simulated brackets, while other approaches adapt pretrained diffusion backbones through gain-map decomposition~\cite{guan2025gmdiffusion,liao2025gmnet} or zero-shot transfer~\cite{zhu2024zeroshot}. LEDiff~\cite{wang2025lediff} enables text-to-HDR and LDR-to-HDR via exposure-aware latent fusion.

X2HDR ~\cite{x2hdr2025} is concurrent image work. It adapts pretrained image diffusion models to HDR by operating in PU21 encoded perceptual space with a frozen VAE and LoRA fine-tuning—reflecting the view that HDR can be handled by mapping it into a compatible latent space. However, these methods rely primarily on display-oriented encodings and are restricted to single images. When applied frame-by-frame to video sequences, single-image diffusion models exhibit temporal flickering and lack cross-frame motion coherence.

\subsection{SDR-to-HDR Video Diffusion Models}
\label{sec:rw_video_diffusion}

Video diffusion models have rapidly advanced from temporal extensions of image diffusion~\cite{ho2022videodiffusion} to large-scale generative foundation models such as Stable Video Diffusion~\cite{blattmann2023svd}, CogVideoX~\cite{yang2024cogvideox}, and Wan2.1~\cite{wan2025}. Modern architectures, including LTX-Video~\cite{ltxvideo2024} and LTX-2~\cite{hacohen2026ltx}, combine highly compressed spatial-temporal video VAEs with scalable Diffusion Transformers (DiT)~\cite{peebles2023dit} and rotary position embeddings~\cite{su2021rope}. Low-Rank Adaptation (LoRA)~\cite{hu2022lora} has become the standard mechanism for parameter-efficient adaptation of these backbones, as demonstrated by concurrent lighting estimation frameworks like LuxDiT~\cite{liang2025luxdit}.

Two recent works explore video diffusion backbones for SDR-to-HDR conversion. DiffHDR (Yu et al., 2026)~\cite{yu2026diffhdr} shares our use of a frozen VAE, log-space encoding, and LoRA adaptation on a video DiT. However, it formulates SDR-to-HDR as a masked radiance inpainting task where luminance masks mark clipped regions, and external text prompts or reference images specify what content to synthesize.
The other one is Tedla et al. (2026)~\cite{tedla2026generating} Fully fine-tunes Wan2.2 to generate a linearized three-exposure SDR bracket ($-4/0/+4$~EV) via temporal latent concatenation, subsequently fusing the bracket with a learned per-frame merger network. All VAE operations remain in native SDR space, forming HDR only post-decoding via classical bracket merging.

Our framework, \method{}, differs from both approaches. Unlike DiffHDR, \method{} treats conversion as SDR to HDR process. The input SDR video is the only input signal, requiring no masks, text prompts, or reference images. Unlike Tedla et al., who keep generative priors in SDR and rely on multi-exposure bracket generation, we map scene-linear HDR directly into ACEScct log space. This allows a frozen VAE to represent high-magnitude radiance directly and enables a lightweight LoRA to emit 16-bit scene-linear HDR video in a single pass, without requiring an exposure ladder or secondary fusion network.

\section{Method}
\label{sec:method}

\begin{figure*}[t]
\centering
\includegraphics[width=\linewidth]{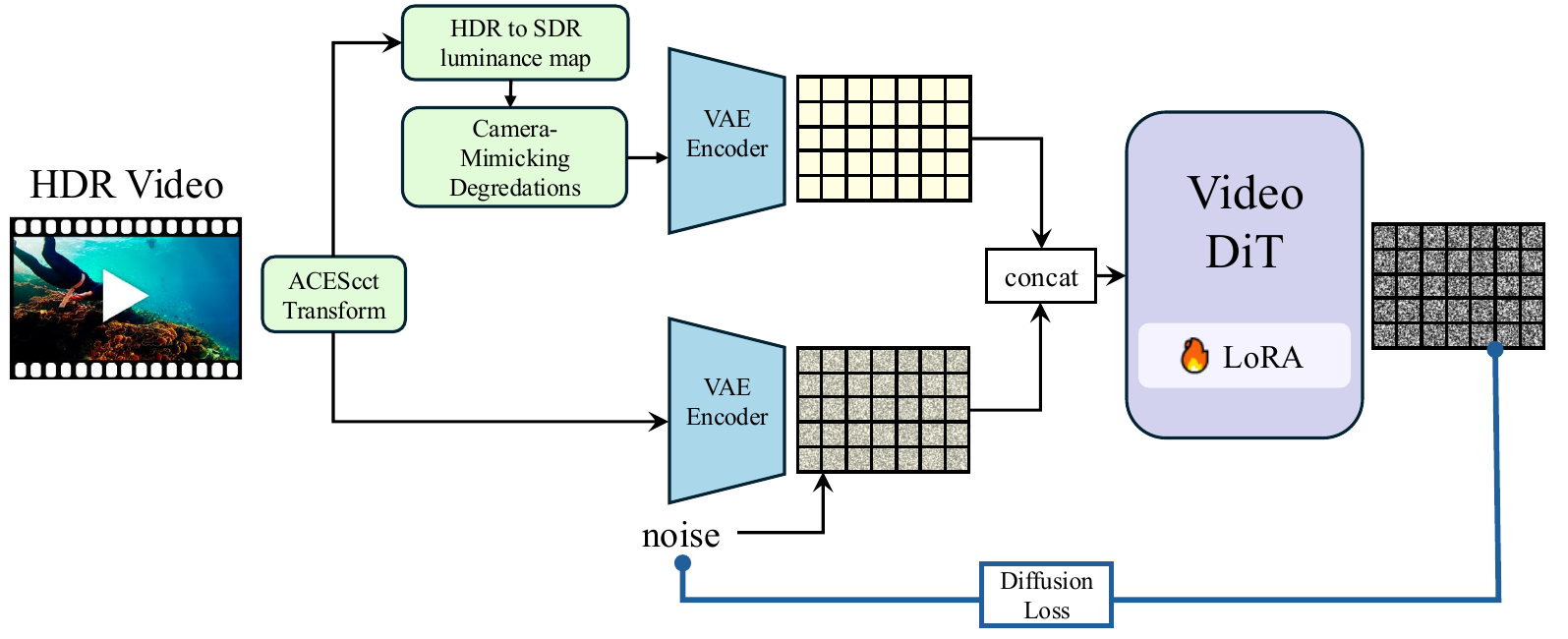}
\caption{\textbf{\method{} training overview:}
Scene-linear HDR is encoded with ACEScct and split.
The HDR branch is passed through the frozen VAE; noise is added in latent space to form the diffusion target.
The SDR condition is formed by the HDR-to-SDR luminance map and camera-mimicking degradations, then encoded by the same VAE.
The two latents are concatenated as a dual canvas, and fed to the video DiT; only the LoRA adapters are trained, with a diffusion loss on the HDR latents.}
\label{fig:training_pipeline}
\end{figure*}

Standard video diffusion models are bound to standard dynamic range (SDR) formats. We address SDR-to-HDR video generation by fine-tuning a Dual Canvas Lora adapter ~\cite{ben2026avcontrol} on top of a frozen video diffusion model. The pretrained video that we are using is video DiT LTX-2.5 ~\cite{hacohen2026ltx}. Given an SDR source video, the network generates a spatially and temporally aligned HDR target. During the training, the model gets clean SDR latents and noisy HDR target to denoise as a dual canvas . At inference, it receives the input SDR latents alongside Gaussian noise, iteratively denoising the noise into the final HDR video.

Our training recipe consists of three core components (Figure~\ref{fig:training_pipeline}):
\begin{enumerate}
    \item ACEScct Log Working Space: Mapping unbounded scene-linear radiance into a cinematic log encoding aligned with the frozen VAE's expected input range.
    \item Condition Degradations \& Exposure Shifts: Applying camera-mimicking degradations and exposure shifts to the SDR canvas to erase shortcut pixels, forcing the model to infer extreme highlights and shadows.
    \item Dual-Canvas LoRA Adaptation: Fine-tuning a lightweight adapter on $\sim 300$ paired clips to learn targeted dynamic-range translation.
\end{enumerate}

\subsection{ACEScct Log Working Space and Latent Alignment}
\label{sec:acescct}

To adapt a pretrained video diffusion model to HDR with lightweight fine-tuning, the model must operate on a latent distribution that closely matches what it was originally trained on. Pretrained video backbones contain learned priors about real-world physics, lighting, and motion acquired from web-scale SDR datasets. Since HDR video data is scarce, retraining the backbone or learning a new latent space would destroy these pretrained representations. Keeping the VAE encoder and decoder frozen ensures that we preserve the original model's knowledge, enabling data-efficient adaptation.

However, directly encoding scene-linear HDR data with the frozen VAE fails. Unbounded, heavy-tailed RGB values deviate drastically from the VAE's expected SDR input range, causing severe  reconstruction artifacts. As shown in Table~\ref{tab:vae_roundtrip}, measuring the direct VAE roundtrip ($\mathbf{x} \to \text{Encoder} \to \text{Decoder} \to \hat{\mathbf{x}}$) reconstruction error on linear RGB frames yields poor quality, resulting in a low PU21-PSNR of 13.99~dB and a high $L_2$ error of 2.81. This low performance is expected since the VAE was trained on SDR data where the color range was 0-1, and not unbounded. A high roundtrip reconstruction error indicates that the VAE latents fail to faithfully capture the HDR video; because radiance details are lost during encoding, the decoder cannot accurately reconstruct the original scene. These degraded latents cannot serve as a viable target for diffusion training.

To bridge this representation gap while maintaining our frozen latent space, we map scene-linear HDR into ACEScct, a standardized cinematic logarithmic color space. ACEScct compresses unbounded radiance into a quasi-logarithmic scale aligned with the input statistics expected by the frozen VAE. Evaluating the VAE roundtrip reconstruction through the full ACEScct HDR target space (clamped at $1.0$) yields an improvement, boosting the PU21-PSNR to 29.97~dB and reducing the $L_2$ error to 0.69. By contrast, clamping the SDR reference condition at $0.55$ (diffuse white) yields 23.02~dB PU21-PSNR, explicitly stripping highlights above white so the reference cannot leak high dynamic range codes into the condition canvas.
We also show that clipping the ACEScct at $1.0$ and then converting  back to ACEScg results in almost no loss in the data with PU21-PSNR > $99.0$

This confirms that the frozen SDR VAE already possesses sufficient capacity to encode and decode HDR content once the color distribution is properly aligned. By maintaining this alignment, we can teach the model to generate full HDR video while retaining the rich generative priors of the original model.

\begin{figure*}[t]
\centering
\includegraphics[width=\linewidth]{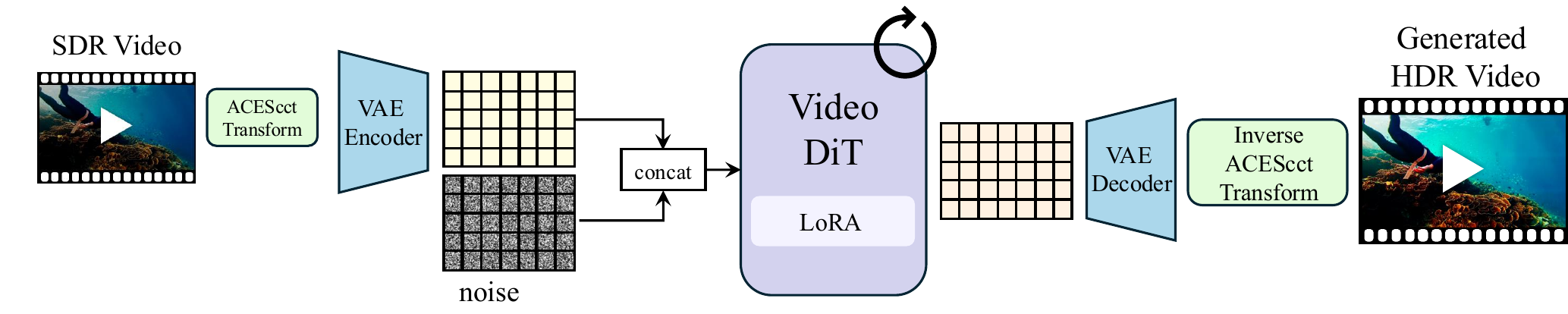}
\caption{\textbf{\method{} Inference Overview.} An SDR video is mapped with ACEScct and encoded by the frozen VAE to produce reference latents $\mathbf{z}_\text{ref}$.
These reference latents are concatenated as a dual canvas with noise and iteratively denoised by the Video DiT with LoRA adapters.
The output latents are VAE-decoded and inverted through ACEScct to recover ACEScg scene-linear HDR video.
The VAE and DiT remain frozen throughout; only the lightweight LoRA adapters were trained via flow matching.}
\label{fig:pipeline}
\end{figure*}

\begin{table}[t]
\centering
\caption{Frozen LTX-2.5 VAE roundtrip.
Poly Haven is 50 orbits, 17 frames; ARRI is 50 clips, 17 frames.
PU21-PSNR and RMSE ($L_2$) are in scene-linear Rec.709
The last row clamps ACEScct at $1.0$ and inverts, with no VAE.}
\label{tab:vae_roundtrip}
\scriptsize
\setlength{\tabcolsep}{3.2pt}
\begin{tabular}{@{}lcccc@{}}
\toprule
Input & \multicolumn{2}{c}{Poly Haven} & \multicolumn{2}{c}{ARRI} \\
\cmidrule(lr){2-3} \cmidrule(lr){4-5}
& PU21-PSNR $\uparrow$ & $L_2$ $\downarrow$ & PU21-PSNR $\uparrow$ & $L_2$ $\downarrow$ \\
\midrule
Linear unclipped & 13.99 & 2.81 & 14.77 & 0.88 \\
ACEScct $0.55$ & 23.02 & 2.80 & 24.15 & 0.80 \\
ACEScct $1.0$ & 29.97 & 0.69 & 30.17 & 0.16 \\
ACEScct $1.0$, no VAE & $>99.00$ & $\sim 0$ & $>99.00$ & $\sim 0$ \\
\bottomrule
\end{tabular}
\end{table}

\textbf{Logarithmic Formulation} Let $x$ denote scene-linear radiance in ACEScg. We map $x$ into the normalized ACEScct logarithmic space via the piecewise transformation $\mathcal{T}(x)$:
\begin{equation}
\mathcal{T}(x) = \begin{cases} 
10.540 \cdot x + 0.073, & x < 2^{-7}, \\
\frac{\log_2(x) + 9.72}{17.52}, & x \ge 2^{-7}.
\end{cases}
\label{eq:acescct_forward}
\end{equation}
The linear region below $x = 2^{-7}$ avoids logarithmic encoding and stabilizes numerical representations in deep shadows. Following iterative denoising and VAE decoding, scene-linear radiance is reconstructed via the inverse transformation $\mathcal{T}^{-1}(y)$ for an encoded log value $y \in [0, 1]$:
\begin{equation}
\mathcal{T}^{-1}(y) = \begin{cases} 
\frac{y - 0.073}{10.540}, & y < 0.155, \\
2^{17.52 \cdot y - 9.72}, & y \ge 0.155.
\end{cases}
\label{eq:acescct_inverse}
\end{equation}

\noindent\textit{(Note: Constants above are rounded for readability; exact double-precision coefficients follow the standard ACEScct specification. ~\cite{ampas2016acescct})}

\subsection{Dual Canvas LoRA}
We adapt a frozen LTX-2.5 video Diffusion Transformer (DiT) \cite{hacohen2026ltx} using a rank-128 dual-canvas LoRA \cite{ben2026avcontrol} trained with flow matching, keeping the VAE frozen. Following the dual-canvas formulation, the input sequence is constructed by concatenating two latent streams along the sequence dimension: a reference canvas containing clean (un-noised) SDR latents, and a target canvas containing noisy HDR latents. During denoising, the target HDR tokens directly attend to the clean SDR tokens through the DiT’s self-attention mechanism. To maintain precise spatial-temporal correspondence, the 3D Rotary Position Embeddings (RoPE) of the SDR reference tokens are aligned directly with those of the target HDR tokens. The flow-matching diffusion loss is computed only on the target HDR tokens.

This in-context conditioning allows the LoRA adapter to learn a localized dynamic-range mapping from the ACEScct SDR reference to the ACEScct HDR target. The frozen backbone already retains priors over video motion and physics, thus, the model does not need to learn video synthesis from scratch, enabling rapid convergence on just a few hundred paired clips ($\sim 300$). Text-to-HDR generation drops the SDR reference canvas and is detailed in Appendix~\ref{app:t2hdr}.

\subsection{Synthetic Dataset and SDR Conditioning}
High-quality HDR video data is scarce in the wild, thus, we construct a synthetic paired SDR-HDR video dataset by rendering synthetic camera orbits through high-resolution PolyHaven 360° HDRi environment maps \cite{polyhaven}. We produced approximately 300 49-frame scene-linear HDR video clips. Each clip is converted to ACEScg and mapped into ACEScct log space via $\mathcal{T}(x)$. 

While our training dataset uses camera orbits through static HDRi environments, all evaluation benchmarks (ARRI cinema footage and youtube in-the-wild clips) are conducted strictly on real-world video footage featuring dynamic real motion, including actor movements, non-rigid object dynamics, complex occlusions, and cinematic camera work, rather than synthetic camera orbits. Generalization from synthetic training data to real-world footage succeeds because motion and temporal priors remain in the frozen DiT backbone. The backbone handles scene dynamics and physics, while the LoRA adapter focuses exclusively on translating ACEScct SDR codes to ACEScct HDR codes. Training the LoRA on PolyHaven HDRI camera orbits provides rich photometric highlight and shadow variations while relying on the frozen backbone for temporal and motion dynamics.

Given the HDR rendered videos we can derive the corresponding SDR conditioning sequences used during training.
In ACEScct log space, diffuse white ($x = 1.0$) maps to $\mathcal{T}(1.0) \approx 0.55$, while the upper boundary $y = 1.0$ caps extreme scene radiance at $\mathcal{T}^{-1}(1.0) \approx 223 \times$ diffuse white. To force the model to learn true dynamic-range translation from the SDR reference to the HDR target we use an asymmetric clamping strategy during training.
\begin{itemize}
    \item SDR reference: Projected to ACEScct and clamped to $[0, 0.55]$. This strips highlight radiance above diffuse white from the reference input, ensuring the SDR condition carries no latent HDR codes.
    \item HDR Target: Clamped to $[0, 1.0]$, exposing target latents to the full dynamic range up to $\approx 223 \times$ white.
\end{itemize}
This $0.55$ vs.\ $1.0$ split forces the network to infer extreme high-magnitude highlights ($y \in (0.55, 1.0]$) from learned diffusion priors rather than leaking pixels from the conditioning canvas.

\subsection{Camera-Mimicking Degradations and Exposure Augmentations}
\label{sec:degradations}

To encourage the model to synthesize true HDR and make the SDR-to-HDR training task more challenging, we apply three camera-mimicking degradation operators to the SDR reference canvas prior to VAE encoding:

\begin{enumerate}
    \item Luminance-Targeted Spatial Blur: Rather than applying uniform spatial blur, we selectively degrade extreme luminance regions in log space using softened selection masks. For clipped highlights near the SDR ceiling ($y \approx 0.55$), Gaussian blur simulates optical dispersion and lens bloom. To prevent the model from overfitting to a fixed blur pattern, the Gaussian standard deviation $\sigma \in [5, 12]$ and kernel size are dynamically randomized across clips. For crushed shadows near the log toe ($y \approx 0.073$), an area-downsampling and cubic-upsampling roundtrip strips fine spatial texture.
    \item Anchored Contrast Scaling: We scale contrast in ACEScct space via $y' = 0.5 + (y - 0.5) \cdot f$. Anchoring around a fixed log midpoint ($0.5$) rather than a per-channel frame mean prevents white-balance and color-neutral drift while driving highlight and shadow values into clipping boundaries.
    \item H.264 Video Recompression: To model web-video compression artifacts, reference sequences undergo H.264 recompression with variable Constant Rate Factors ($\text{CRF} \in [23, 35]$), introducing macroblocking and quantization noise directly into the SDR input.
\end{enumerate}
Full hyperparameter distributions and implementation specifics are provided in Appendix~\ref{app:degradations}.

In addition to structural degradations, we apply joint exposure shifts across both the SDR condition and HDR target canvases. Simultaneously shifting both streams by shared random exposure offsets (spanning $-2,0,2$ $\text{EV}$ stops) exposes the network to diverse luminance scales. This forces the model to recover both extreme highlights and deep shadow details across varied exposure levels while preserving spatial alignment between condition and target.

Together, these operations erase structural details in clamped or blurred regions, forcing the network to synthesize continuous physical radiance from its learned diffusion priors rather than reproducing input pixels. Consequently, the model generates smooth, unclipped specular gradients when pulling exposure down ($\text{EV}{-4}/{-3}$) and reveals coherent spatial texture when pushing shadows up ($\text{EV}{+3}/{+4}$).

\subsection{Inference Pipeline}
During inference, an input SDR video is converted from Rec.709 primaries to linear ACEScg, mapped via $\mathcal{T}(x)$ into the clamped ACEScct $[0, 0.55]$ and encoded by the frozen VAE. The Video Diffusion Transformer (DiT) with dual-canvas LoRA adapters performs iterative denoising in ACEScct latent space. Finally, decoded latents in ACEScct $[0, 1]$ are mapped back to scene-linear ACEScg via $\mathcal{T}^{-1}(y)$ and saved as 16-bit half-float EXR files.

\section{Experiments}
\label{sec:experiments}

\begin{figure*}
      \setlength{\tabcolsep}{1pt}
      \centering
      \small
      \newcommand{\carlab}[1]{\rotatebox{90}{\makebox[1.5cm][c]{\scriptsize #1}}}
      \begin{tabular}{@{}c@{\hspace{6pt}}c@{}}
      \begin{tabular}{cccc}
      \carlab{Input (SDR)} &
      \includegraphics[width=0.152\linewidth]{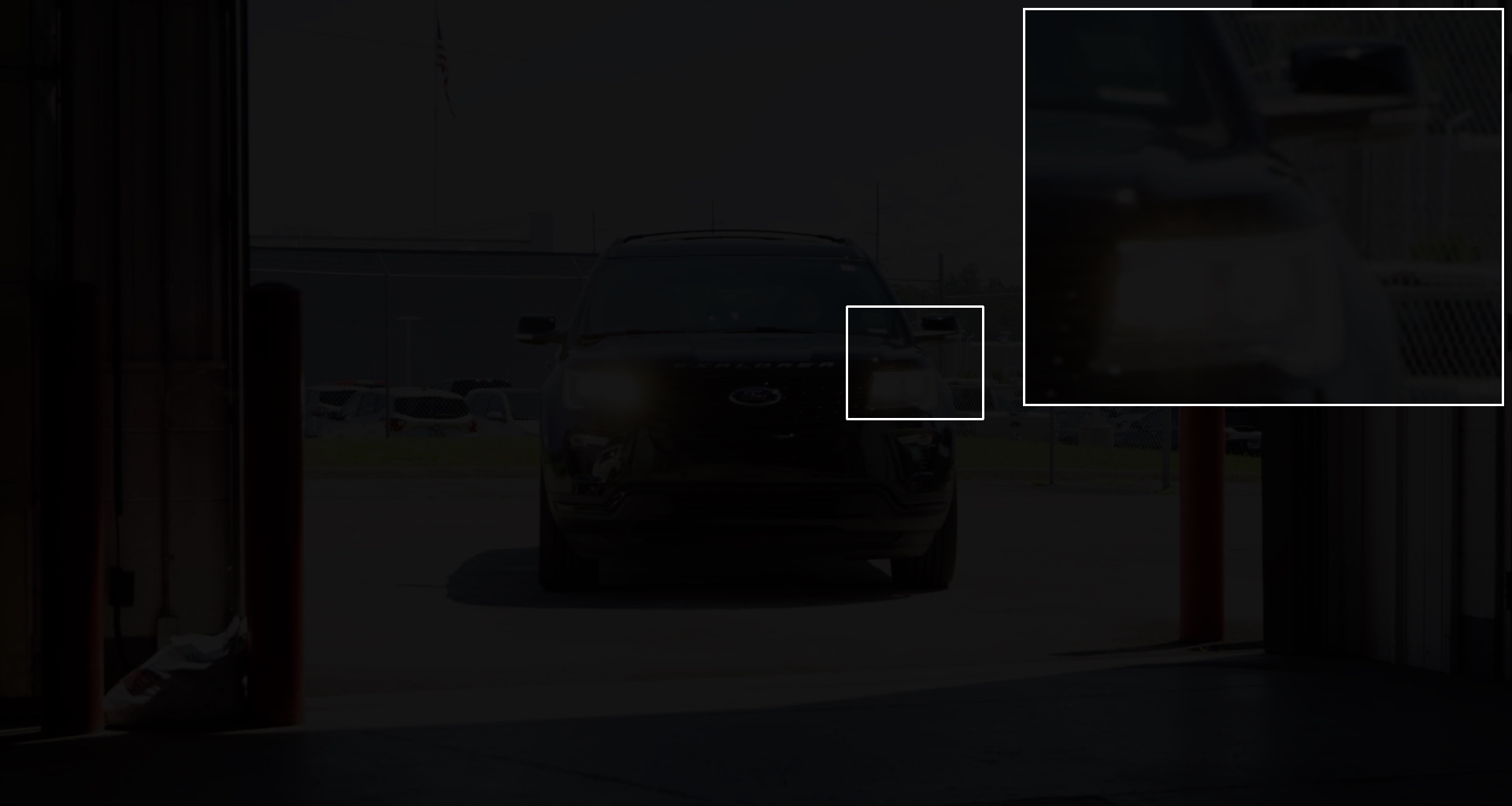} &
      \includegraphics[width=0.152\linewidth]{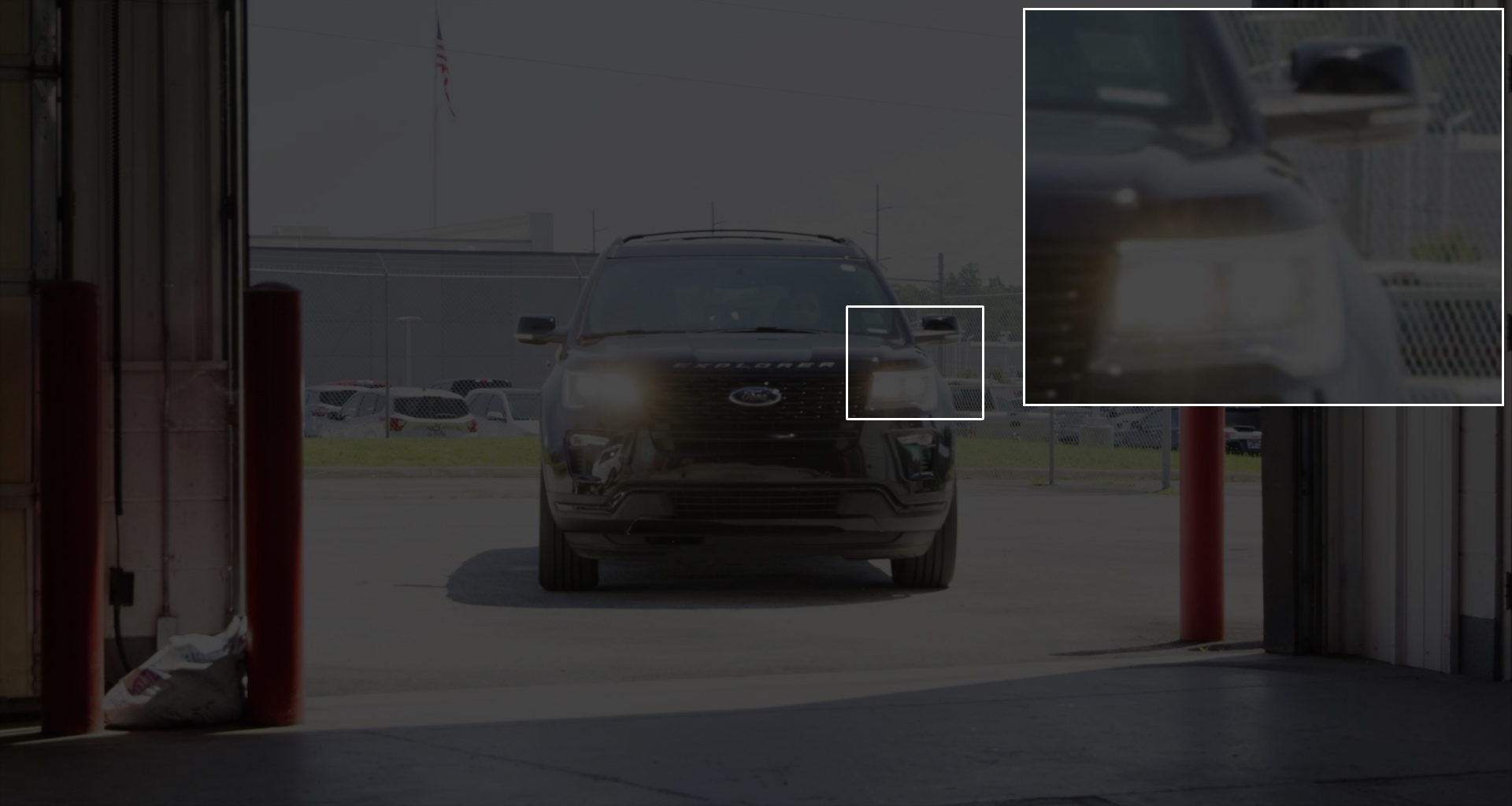} &
      \includegraphics[width=0.152\linewidth]{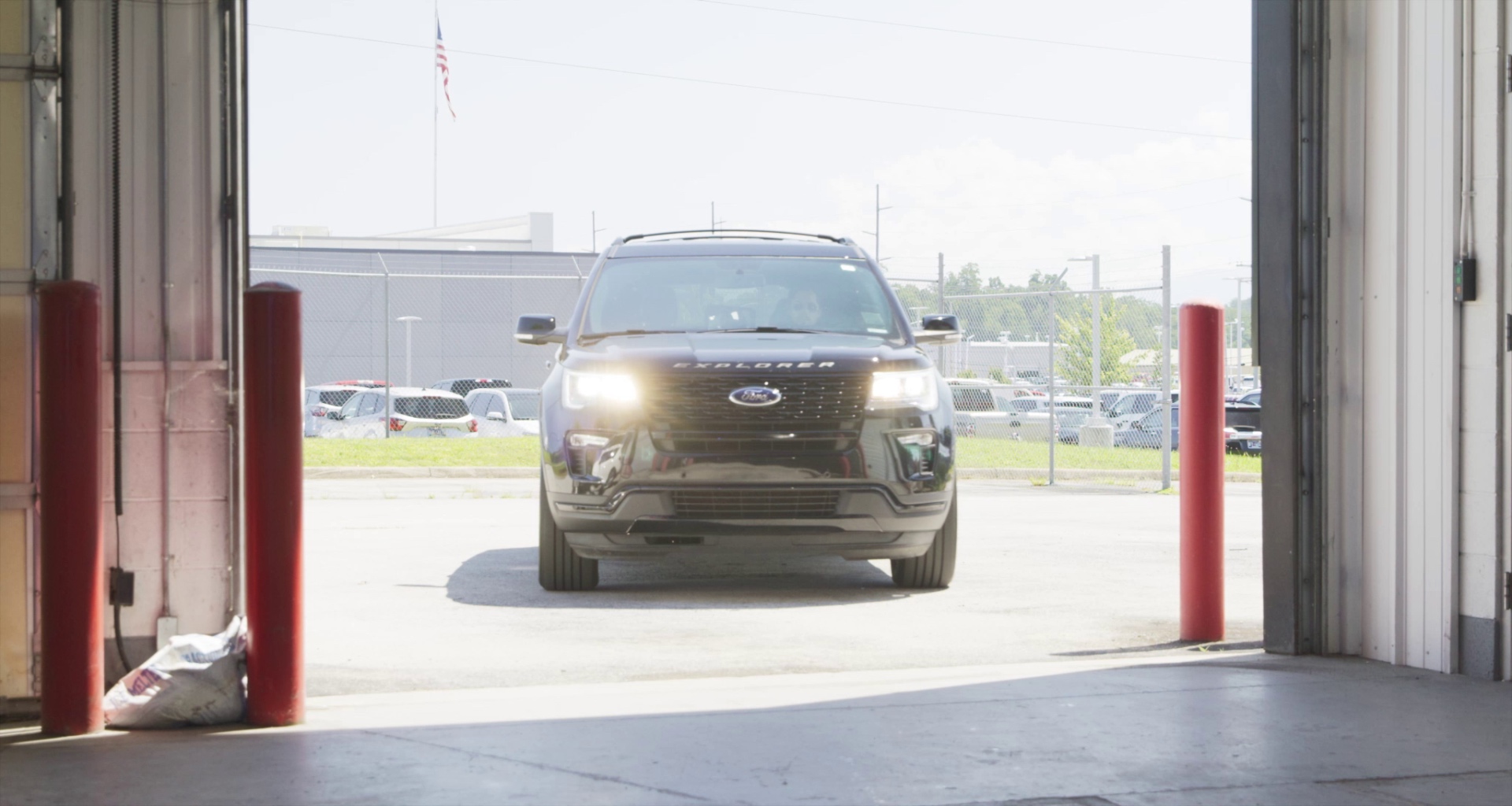} \\
      \carlab{HDRTVNet++} &
      \includegraphics[width=0.152\linewidth]{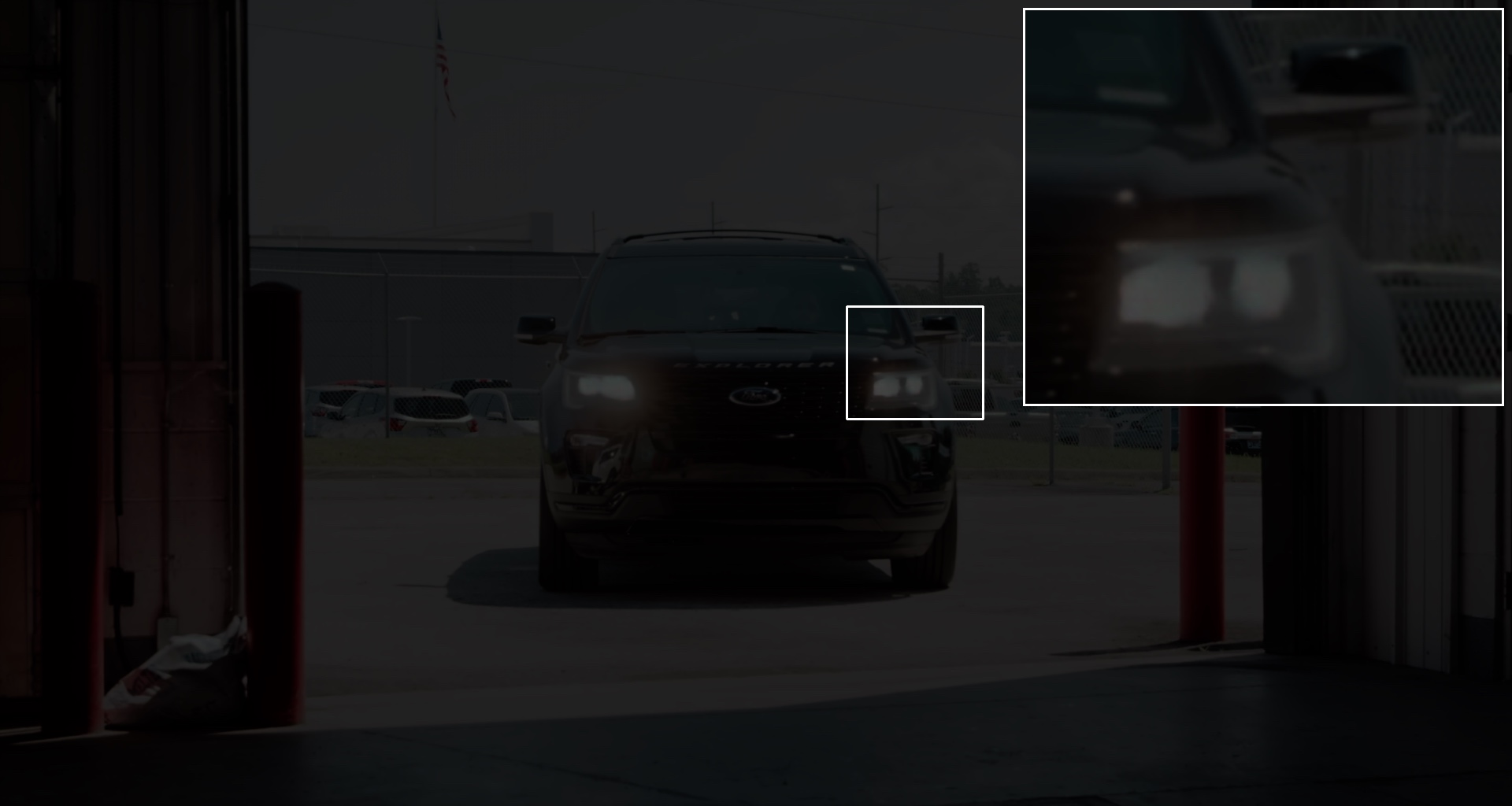} &
      \includegraphics[width=0.152\linewidth]{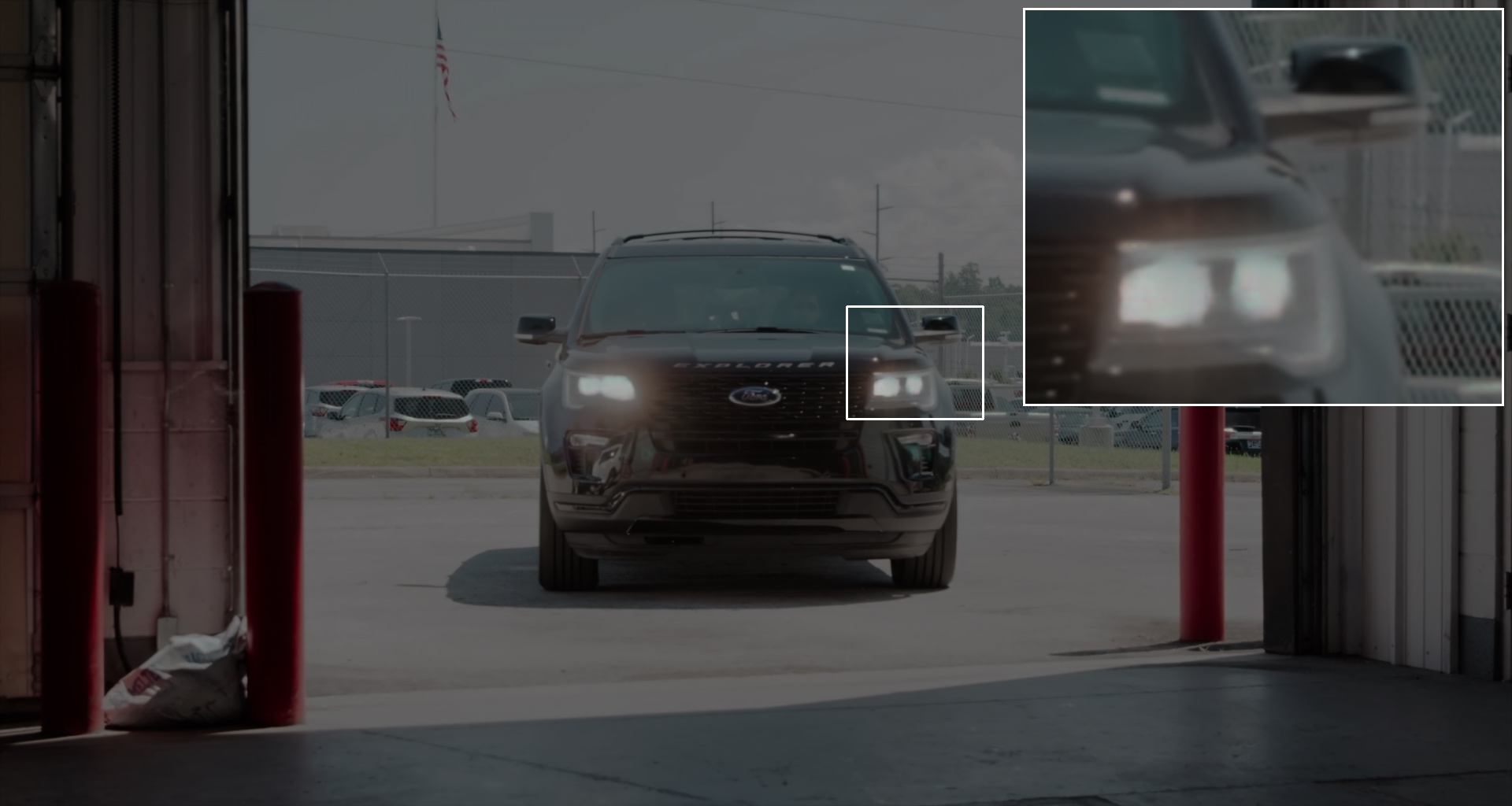} &
      \includegraphics[width=0.152\linewidth]{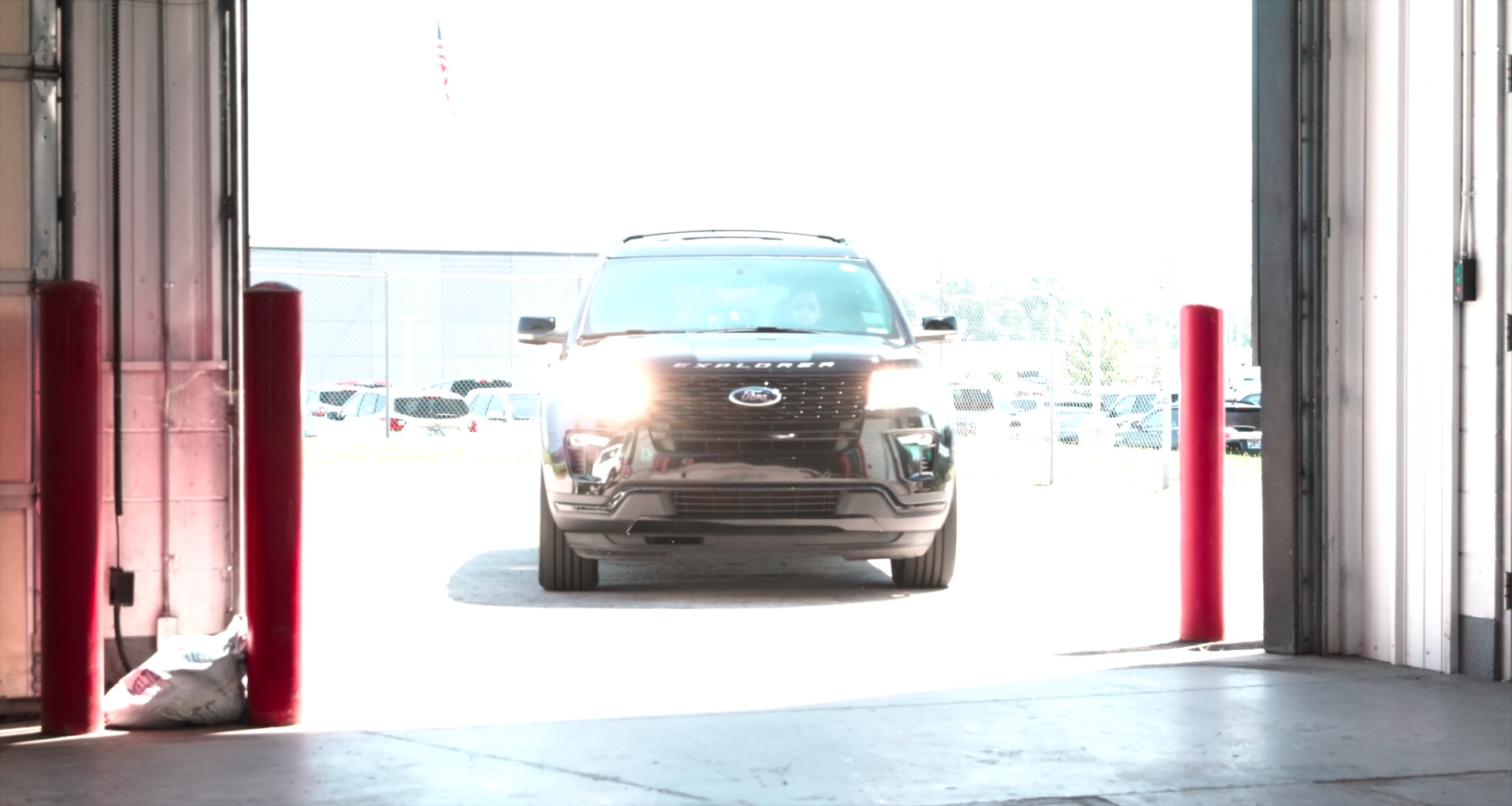} \\
      \carlab{X2HDR} &
      \includegraphics[width=0.152\linewidth]{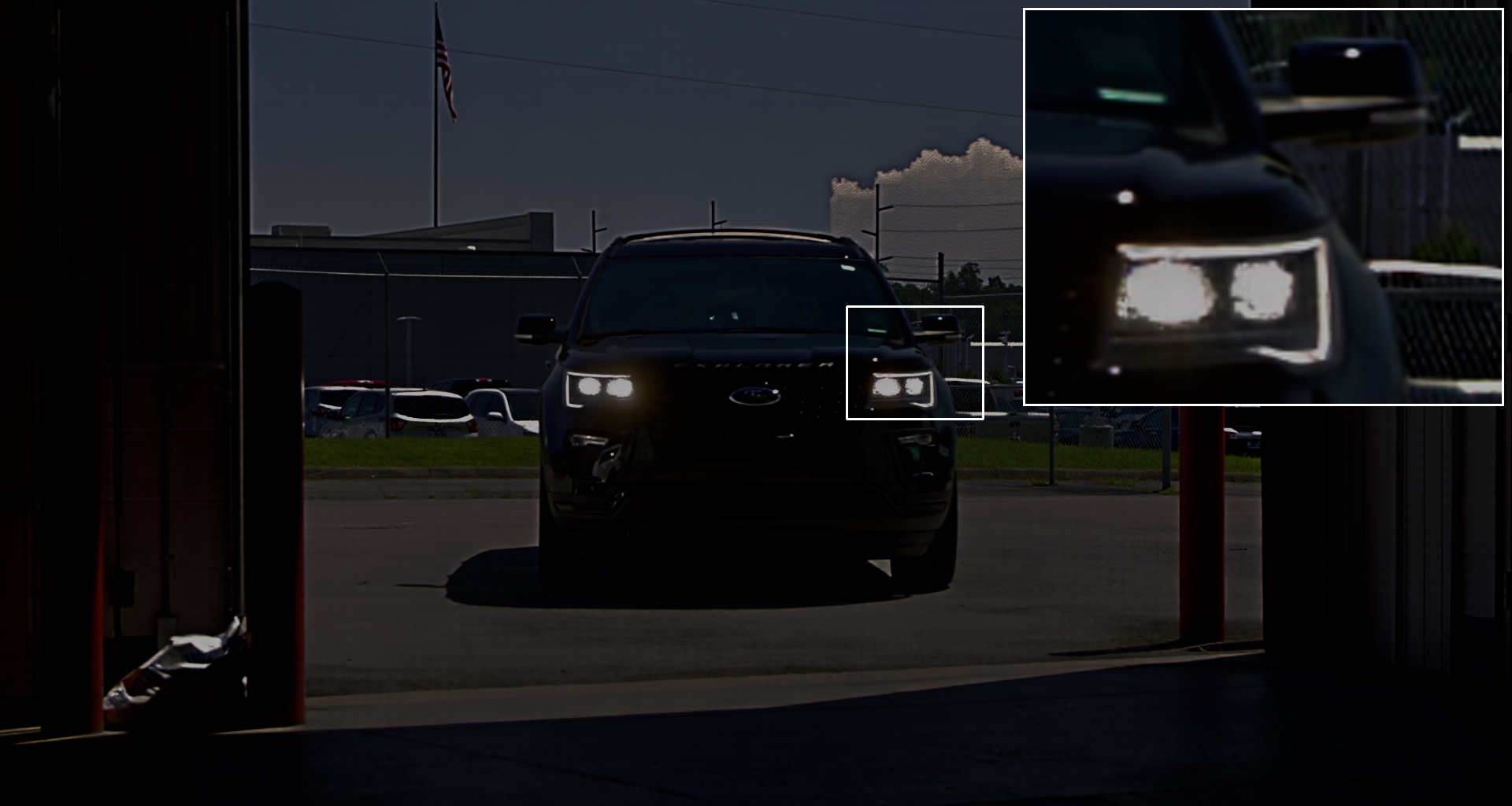} &
      \includegraphics[width=0.152\linewidth]{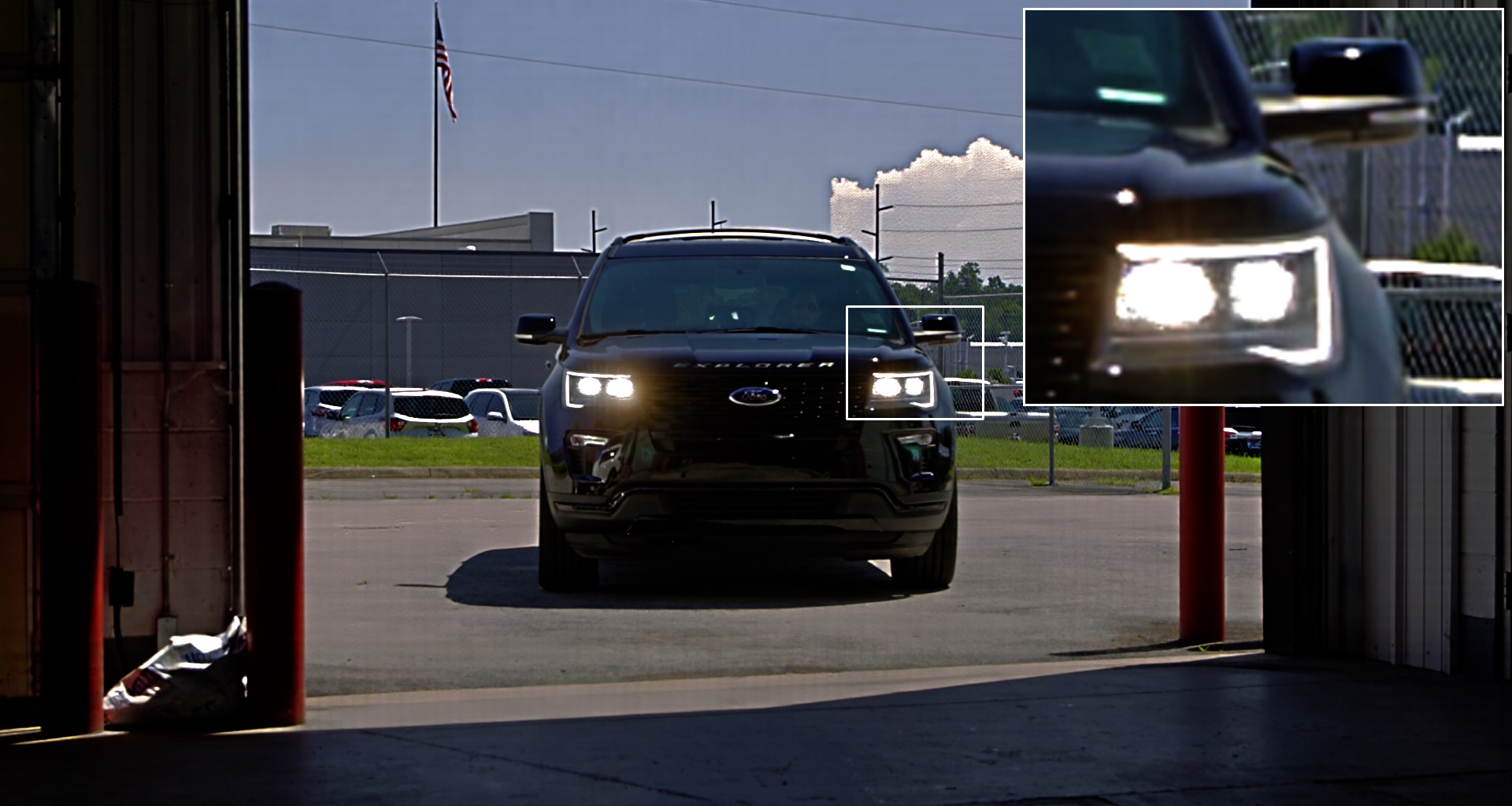} &
      \includegraphics[width=0.152\linewidth]{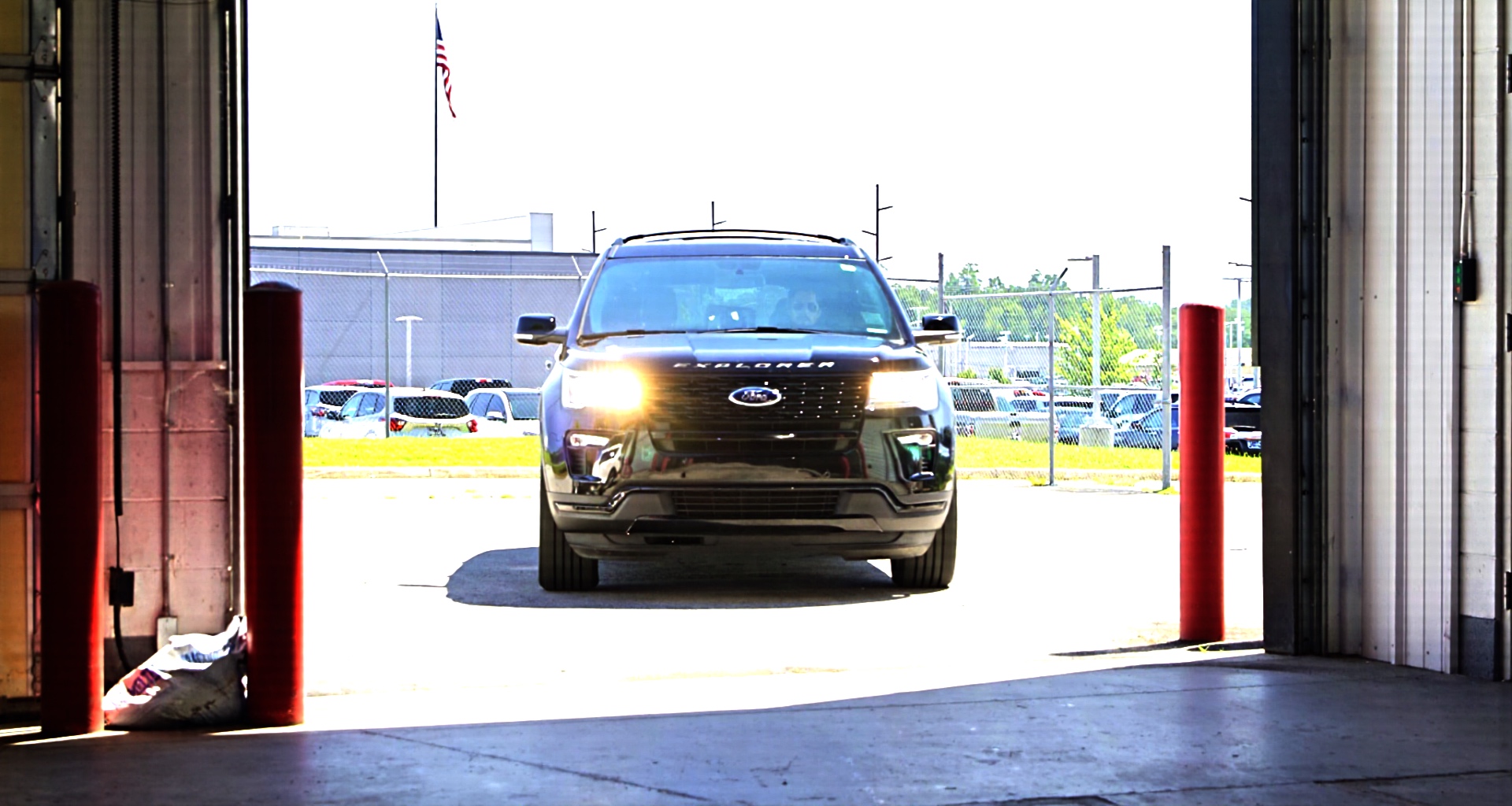} \\
      & EV $-7$ & EV $-3$ & EV $0$
      \end{tabular}
      &
      \begin{tabular}{cccc}
      \carlab{DiffHDR} &
      \includegraphics[width=0.152\linewidth]{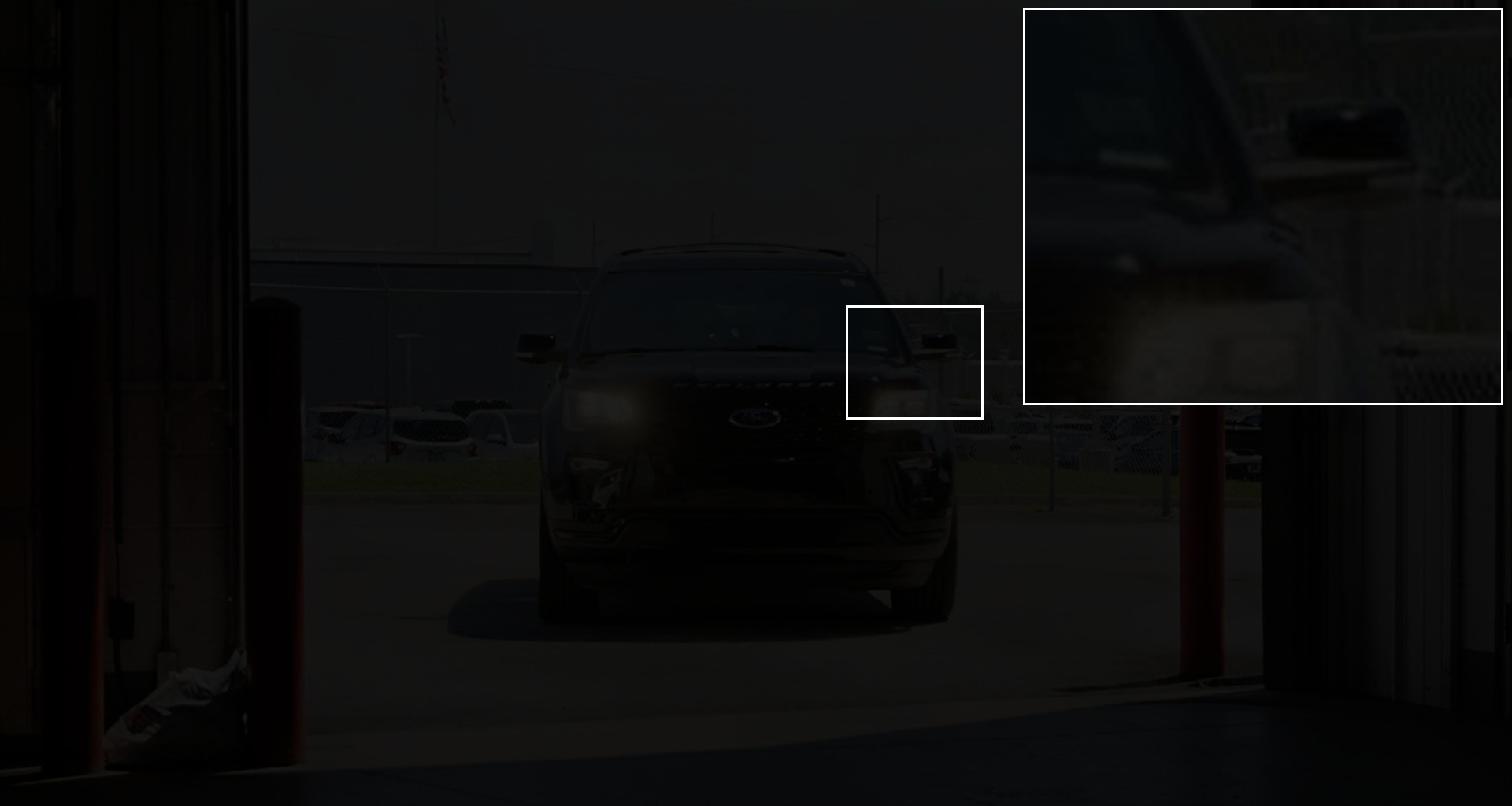} &
      \includegraphics[width=0.152\linewidth]{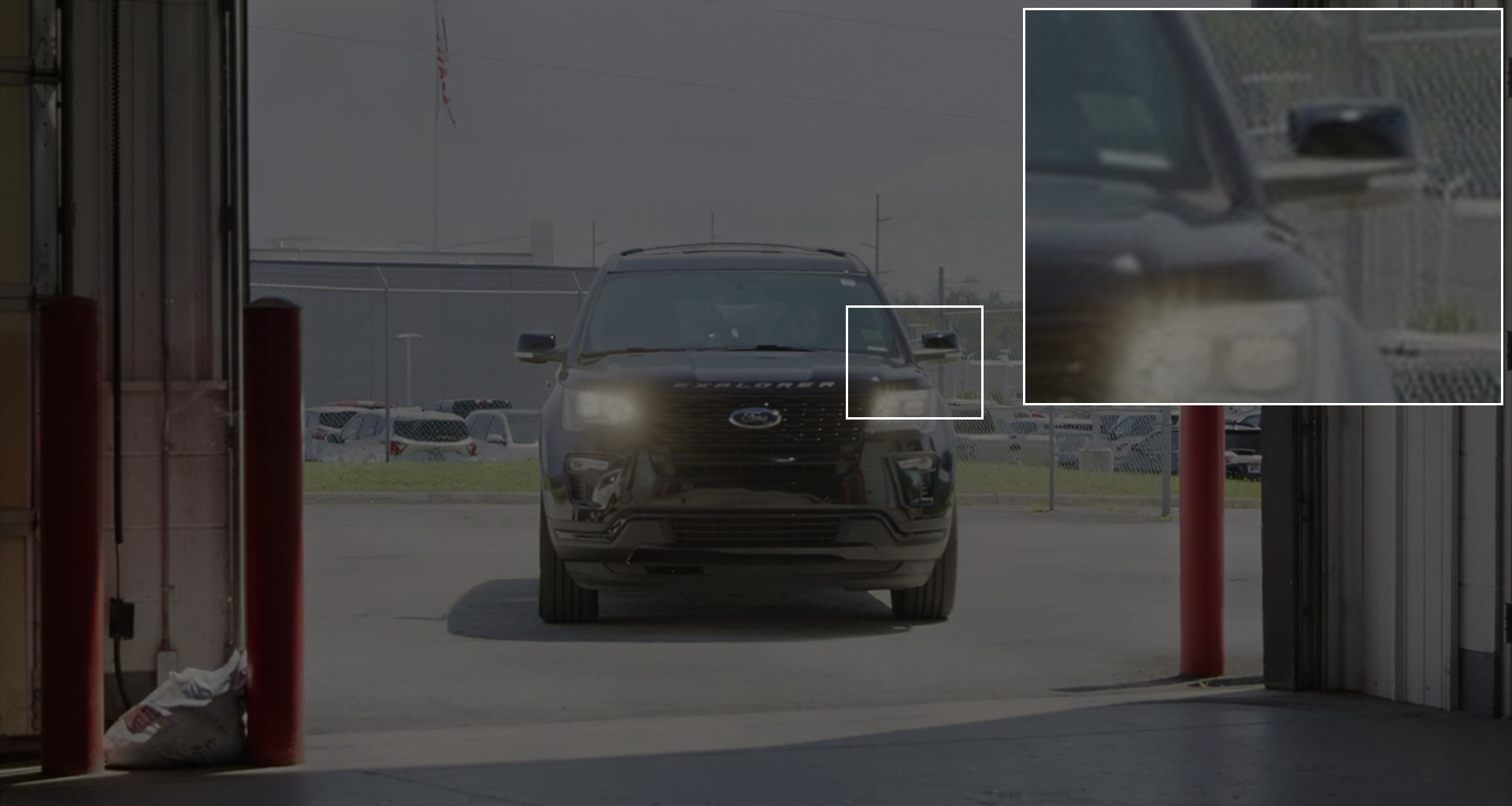} &
      \includegraphics[width=0.152\linewidth]{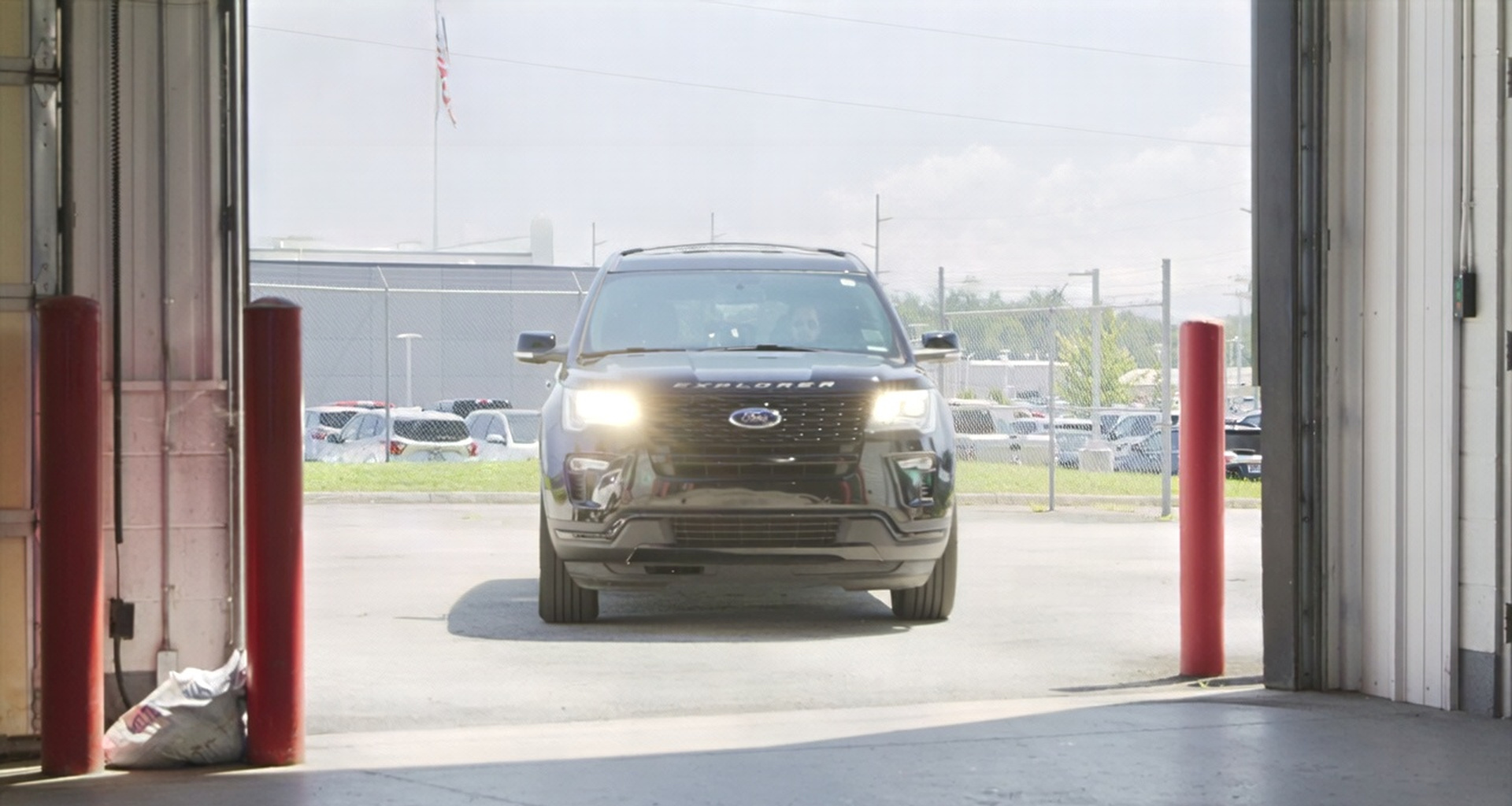} \\
      \carlab{Tedla et al.} &
      \includegraphics[width=0.152\linewidth]{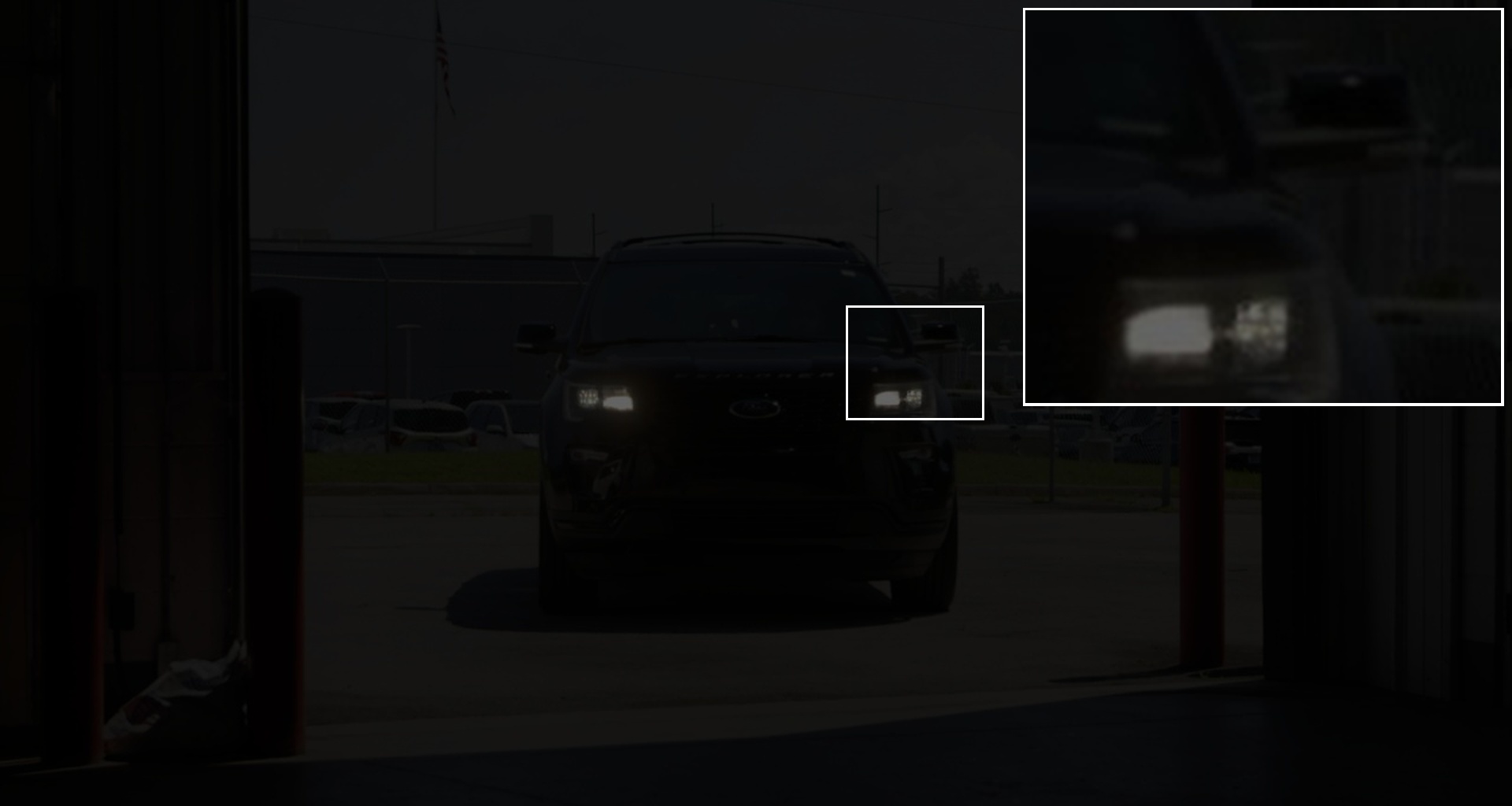} &
      \includegraphics[width=0.152\linewidth]{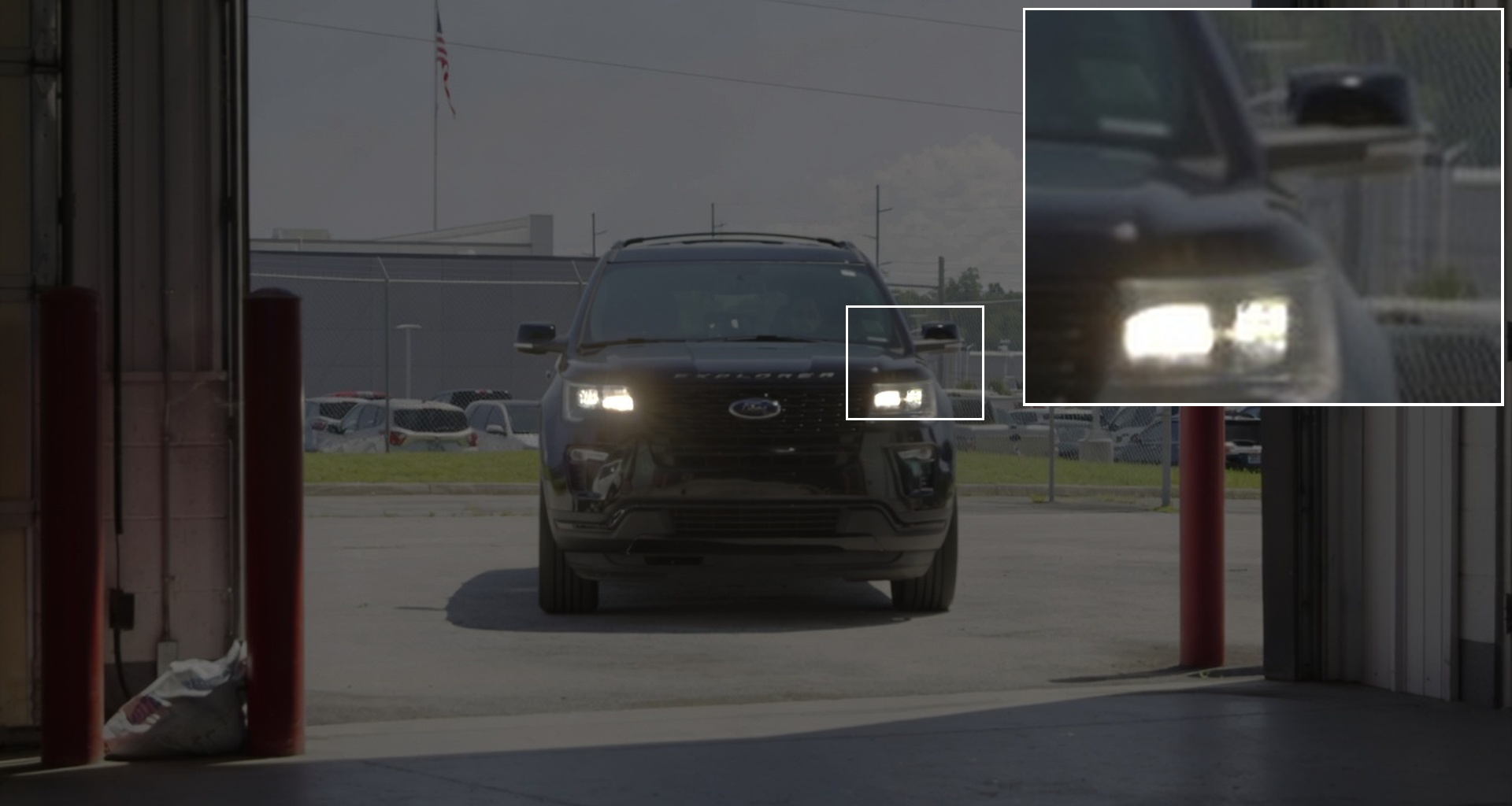} &
      \includegraphics[width=0.152\linewidth]{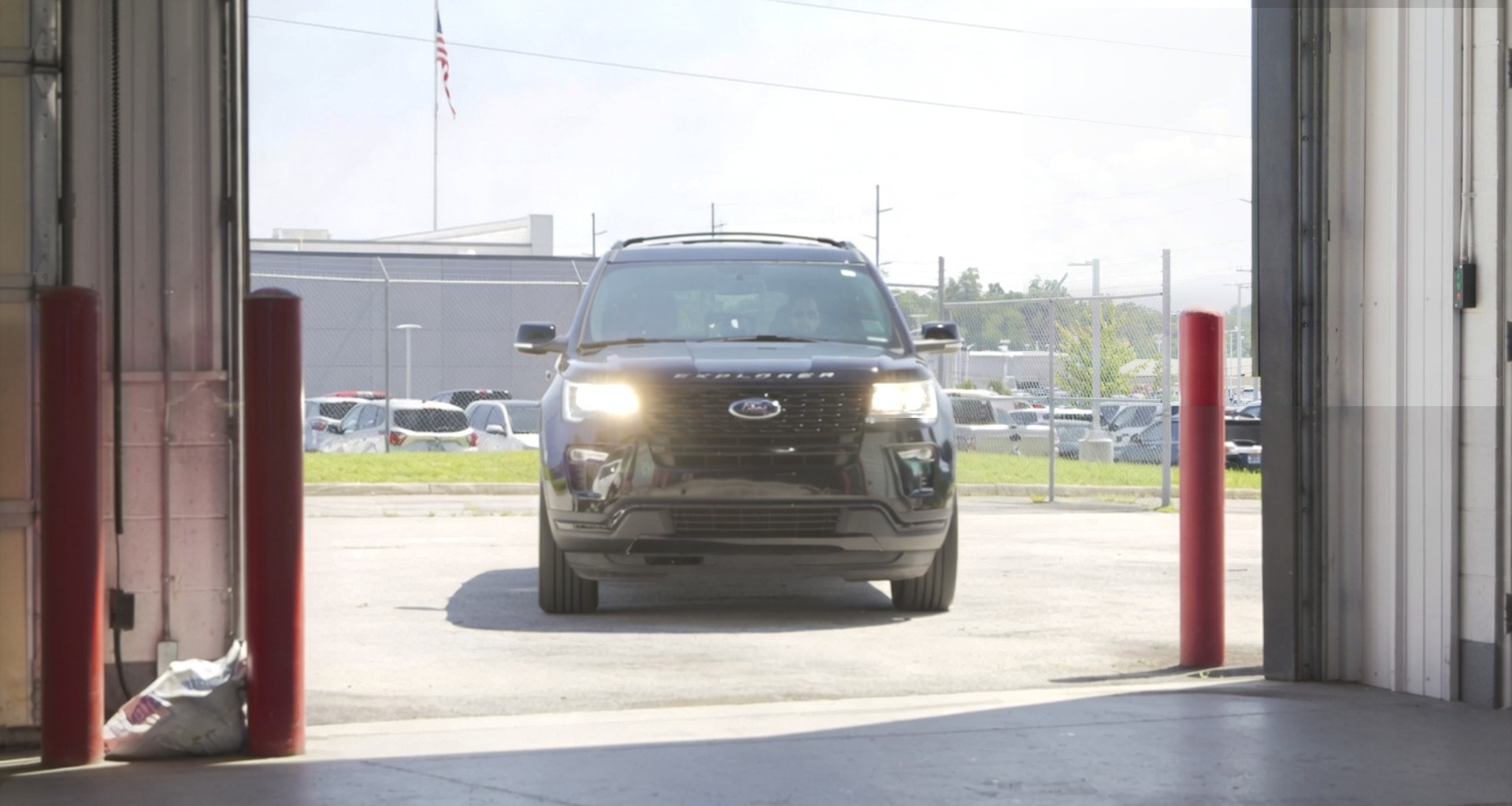} \\
      \carlab{\method{}} &
      \includegraphics[width=0.152\linewidth]{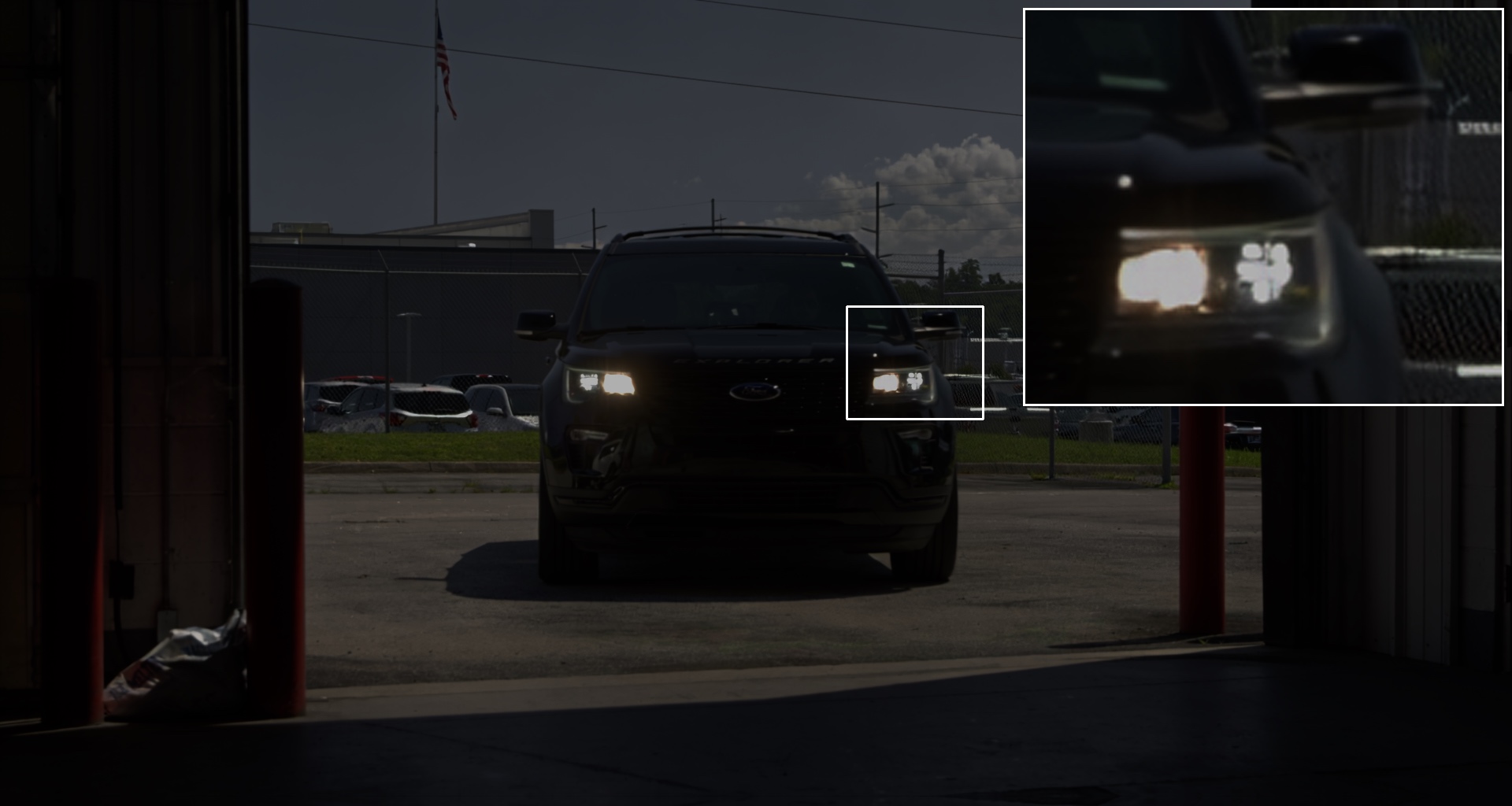} &
      \includegraphics[width=0.152\linewidth]{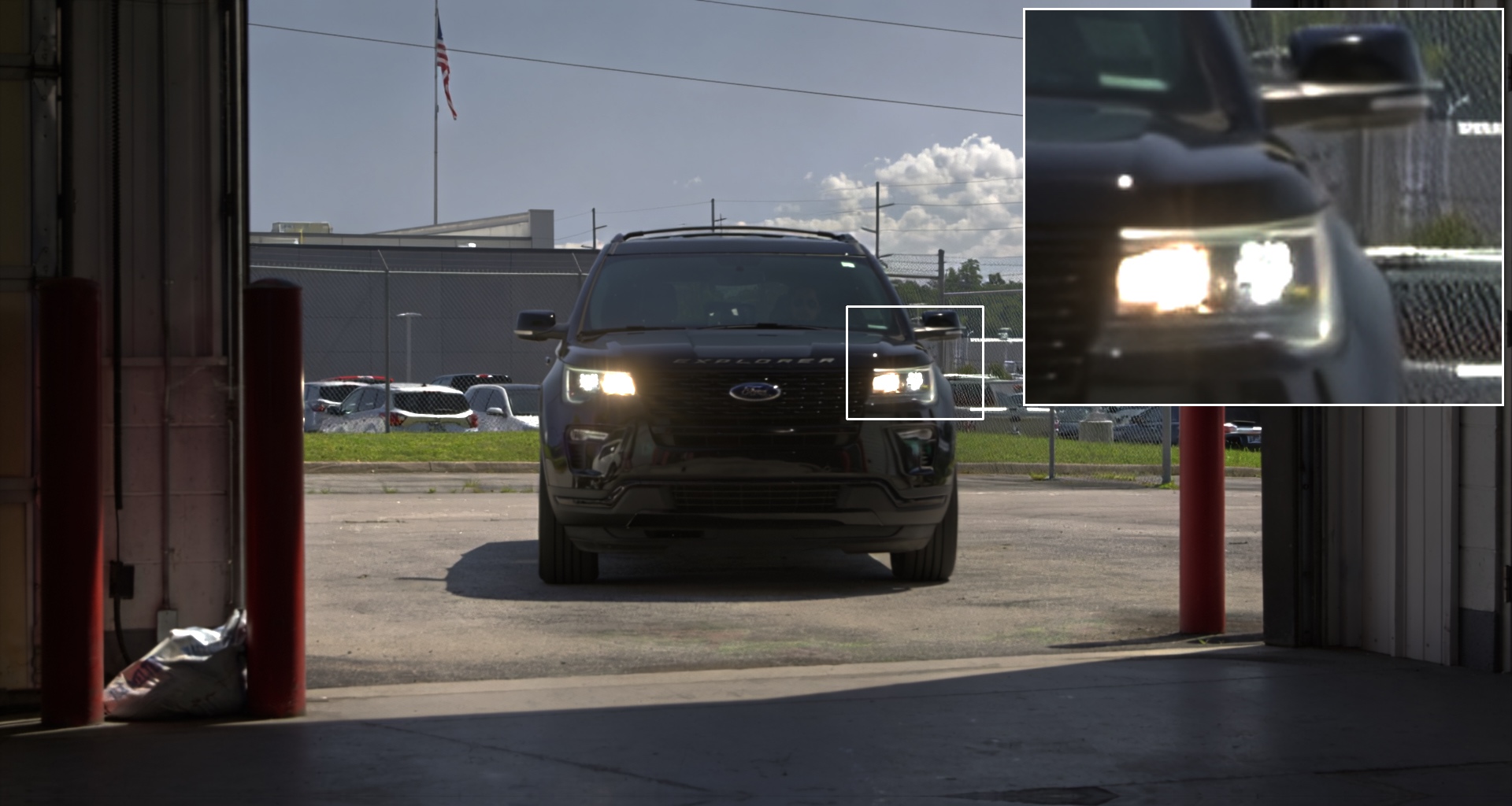} &
      \includegraphics[width=0.152\linewidth]{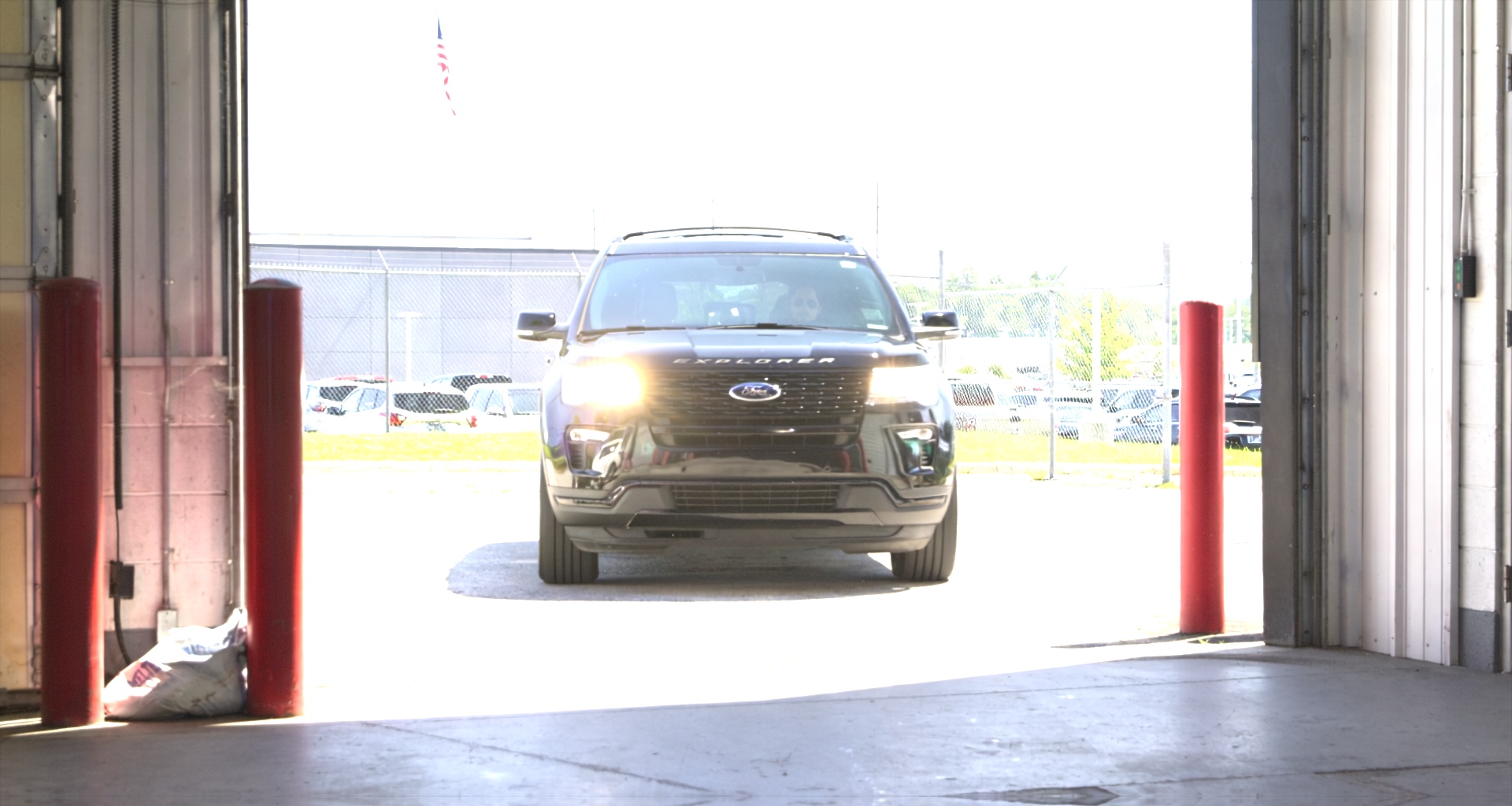} \\
      & EV $-7$ & EV $-3$ & EV $0$
      \end{tabular}
      \end{tabular}
      \caption{HDR generation on an ActionVFX~\cite{actionvfx_license} garage plate,
      tone-mapped at EV $-7$, $-3$, and $0$.
      Insets show the right headlight.
      At EV~$0$ the opening is blown out in every row; pulling exposure down
      shows what each method stored in the highlights.
      \method{} recovers cloud structure through the door and a structured
      headlight.
      HDRTVNet++ barely expands the range.
      DiffHDR stays close to the SDR input (Table~\ref{tab:wild}: $Y_{99.9}{=}0.84$).
      Tedla et al.~\cite{tedla2026generating} (SDR2HDR) leaves little structure in the
      lamp at EV $-7$.
      X2HDR recovers some sky but clips the lamps.}
      \label{fig:car_comparison}
\end{figure*}

\begin{figure*}
      \setlength{\tabcolsep}{1pt}
      \centering
      \small
      \begin{tabular}{ccccc}
      \raisebox{4pt}{\rotatebox{90}{Input (SDR)}} &
      \includegraphics[width=0.24\linewidth]{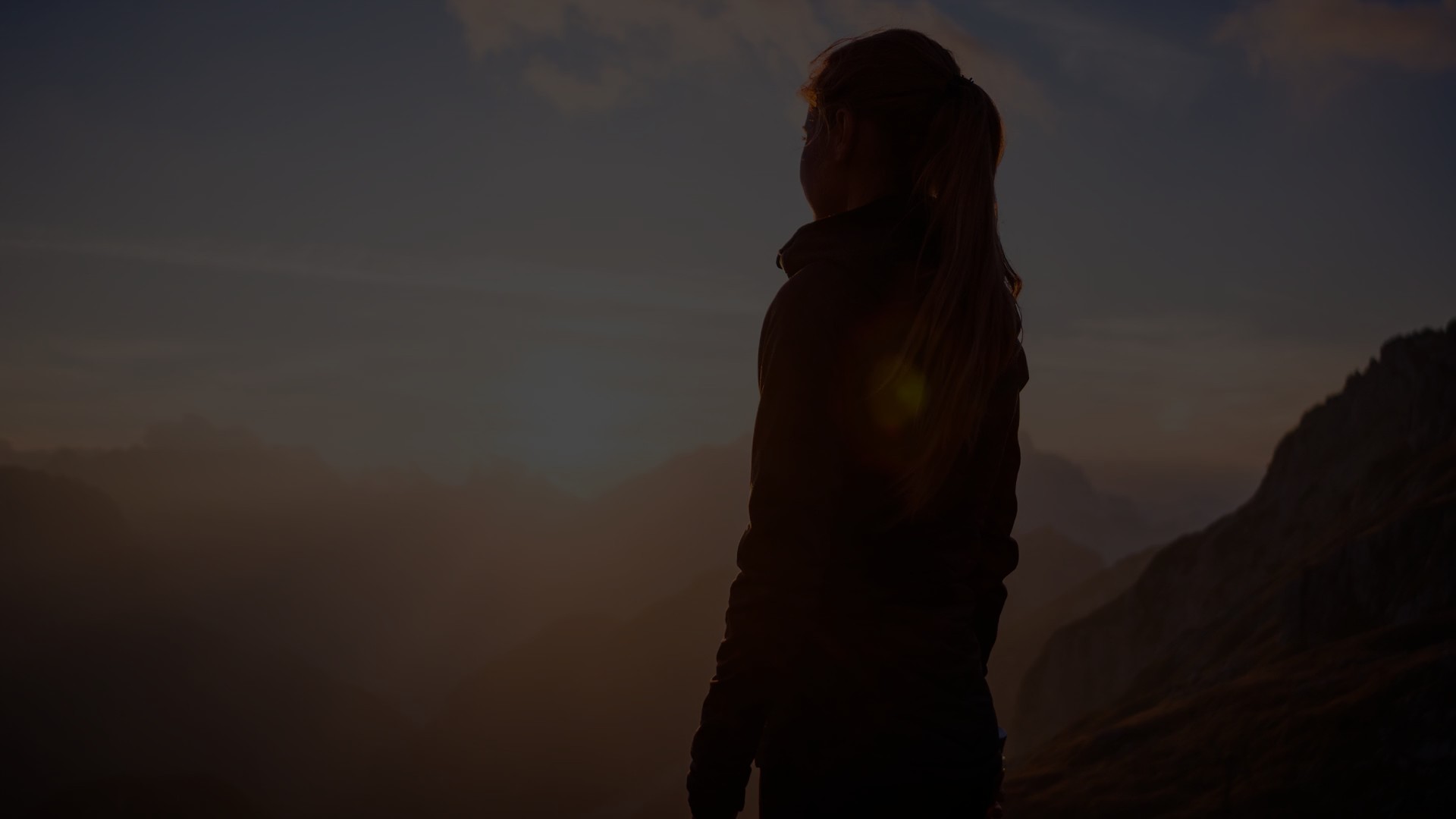} &
      \includegraphics[width=0.24\linewidth]{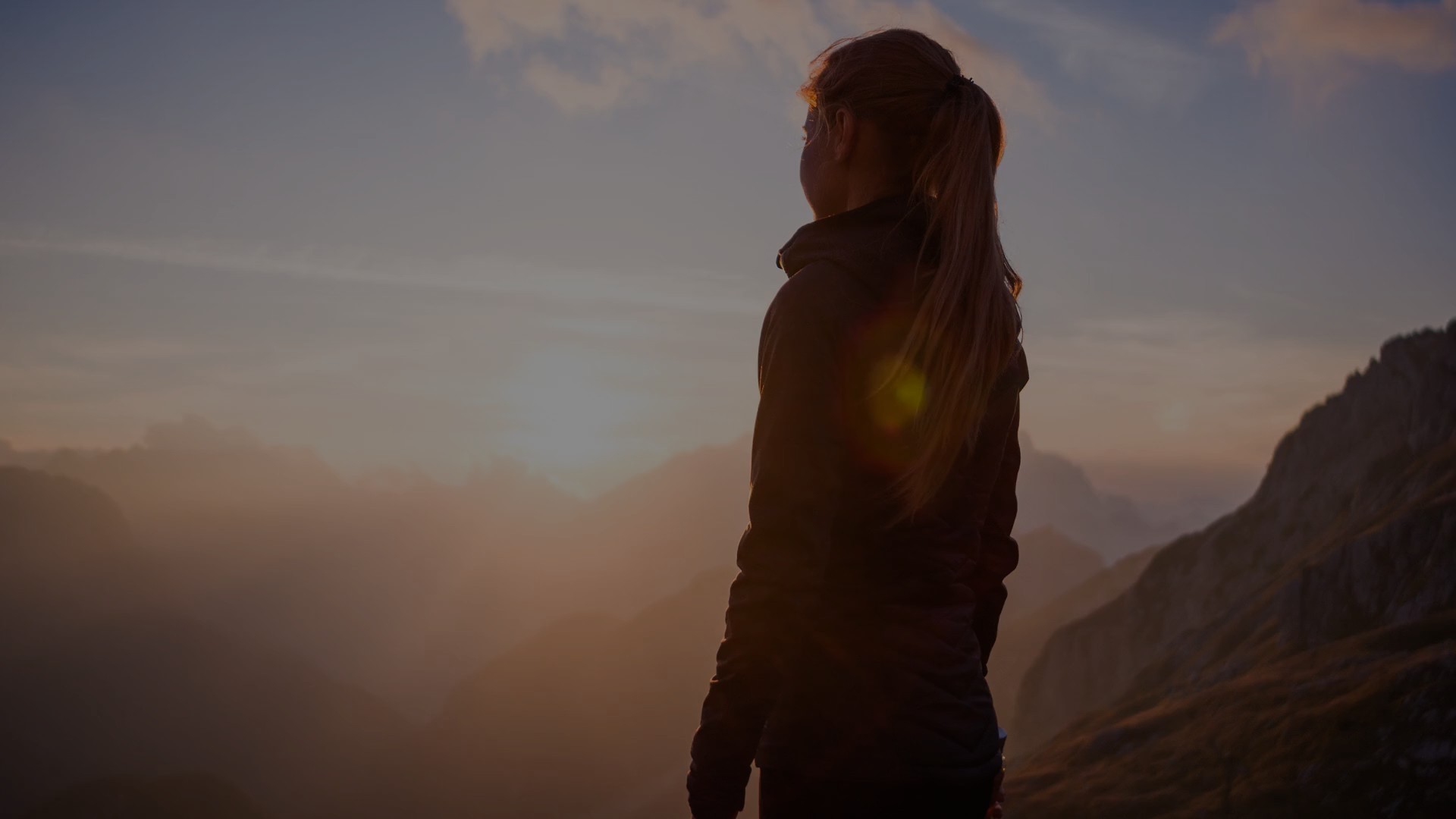} &
      \includegraphics[width=0.24\linewidth]{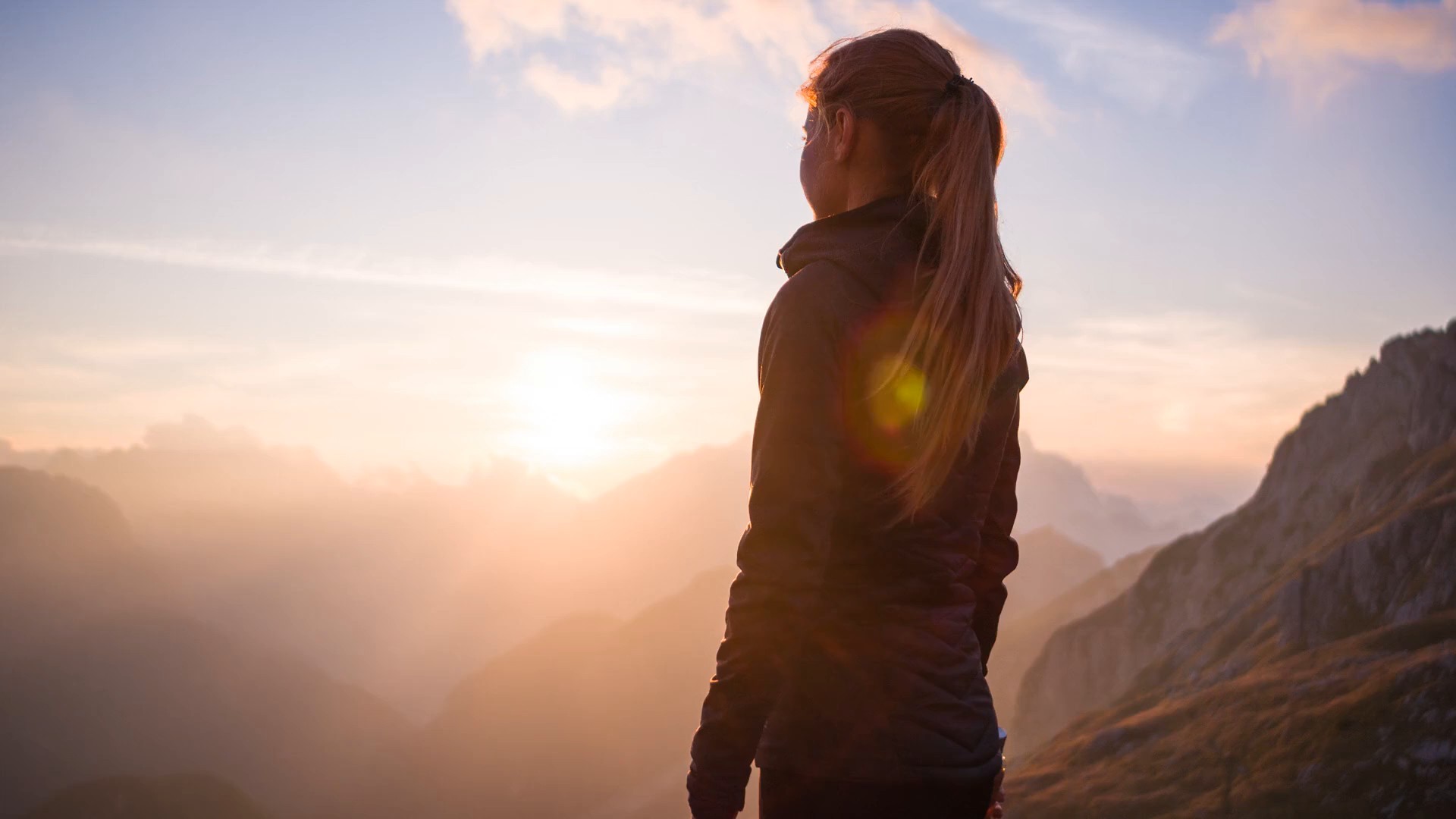} &
      \includegraphics[width=0.24\linewidth]{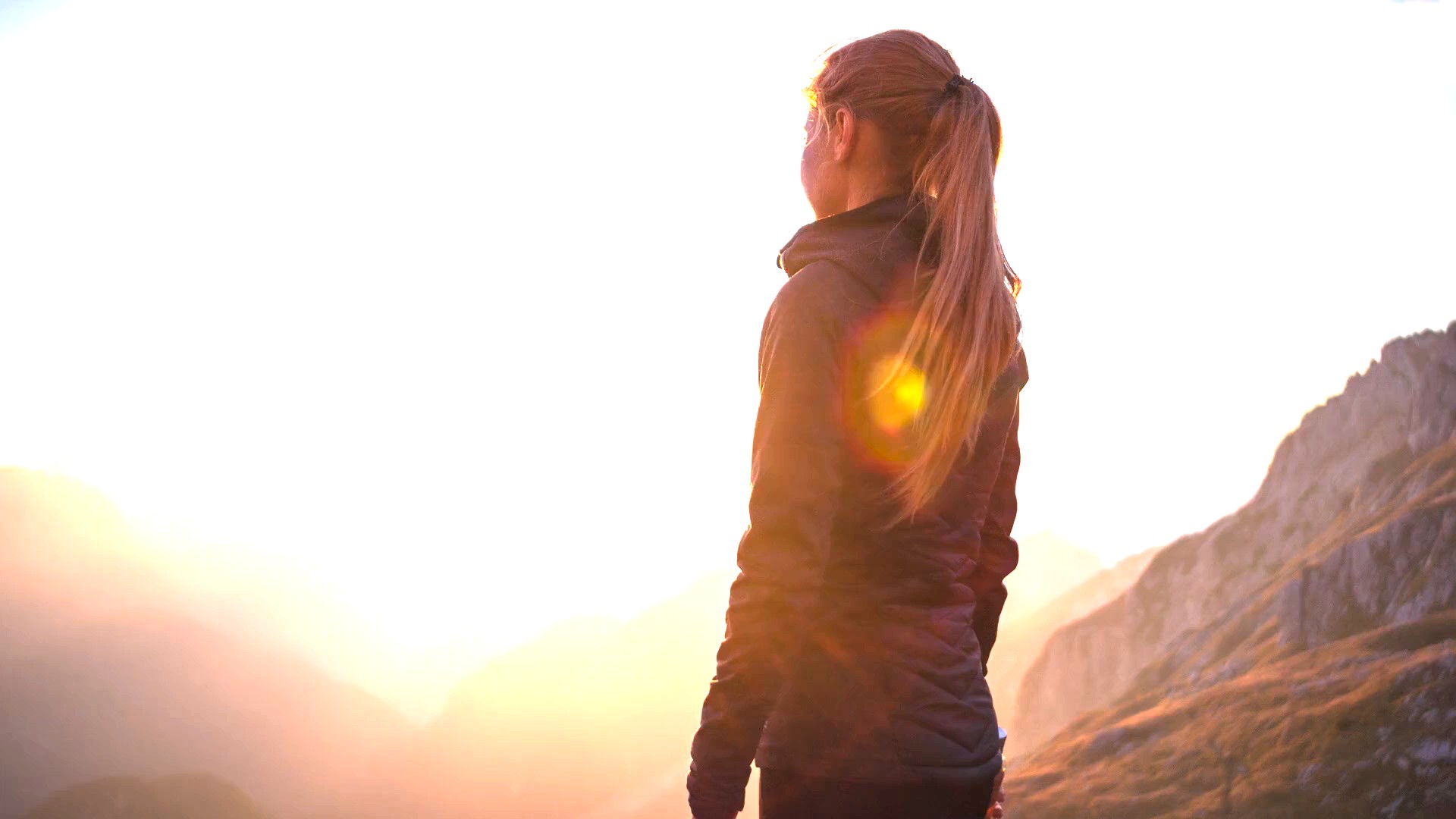} \\
      \raisebox{12pt}{\rotatebox{90}{DiffHDR}} &
      \includegraphics[width=0.24\linewidth]{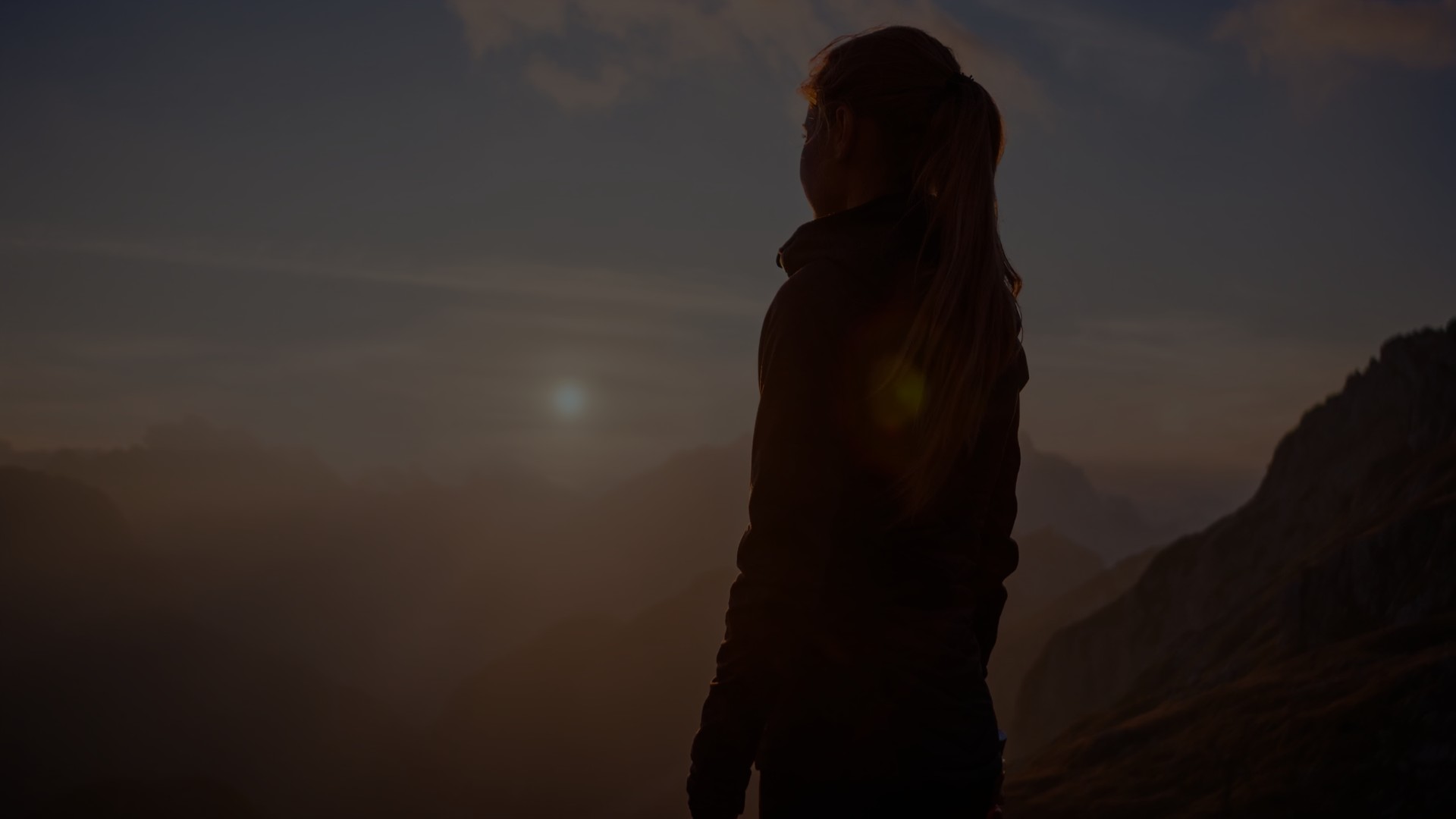} &
      \includegraphics[width=0.24\linewidth]{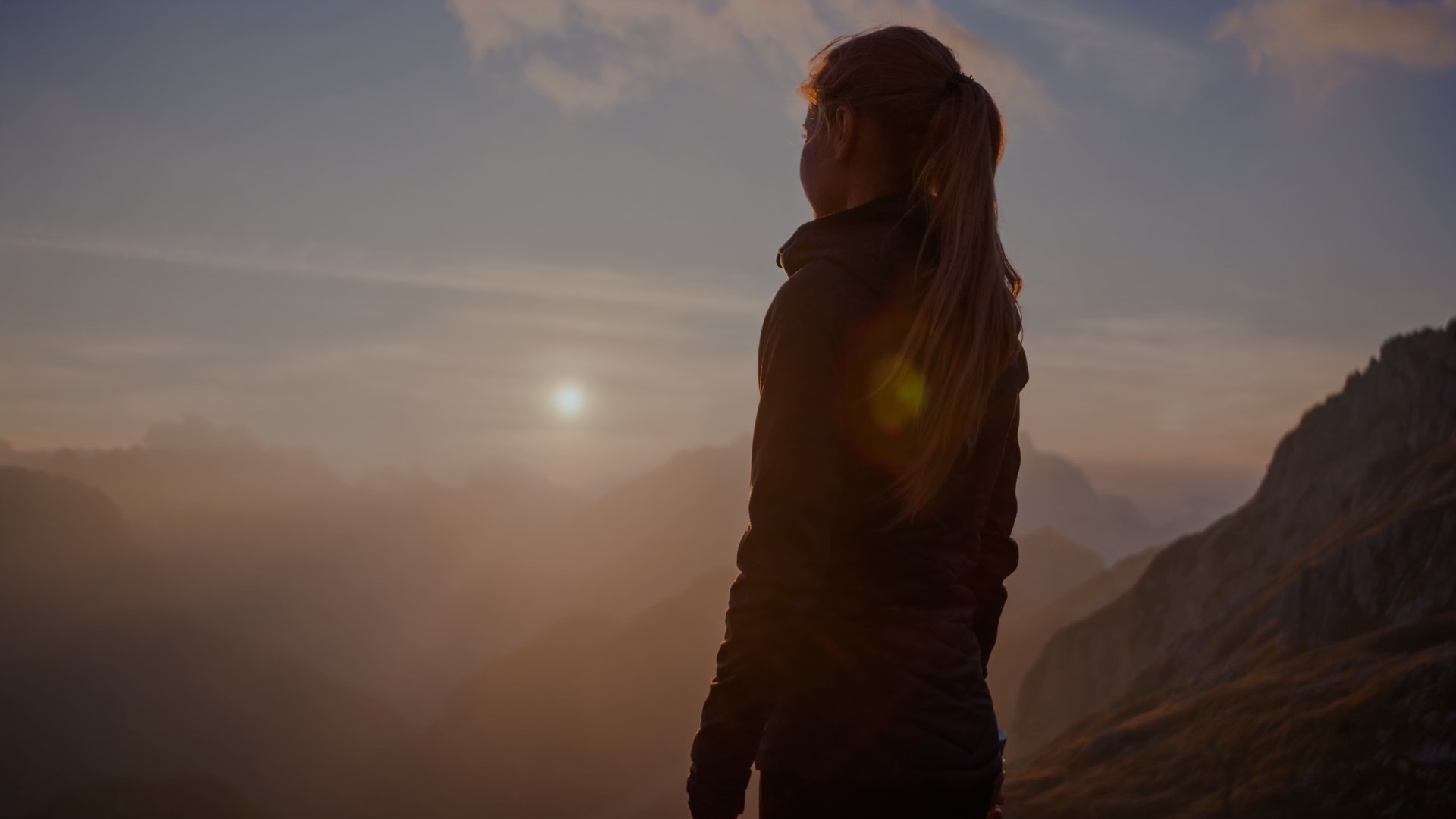} &
      \includegraphics[width=0.24\linewidth]{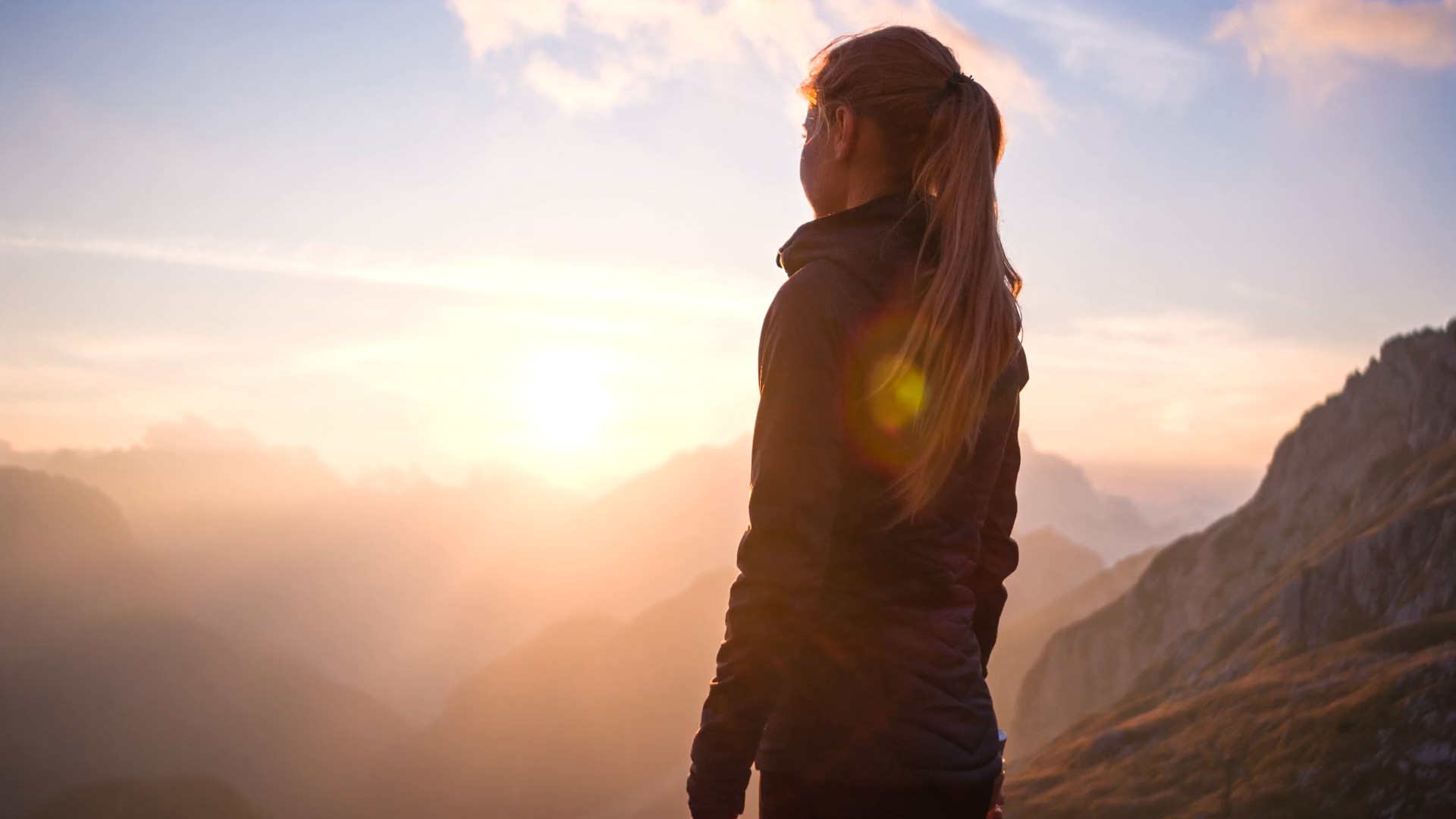} &
      \includegraphics[width=0.24\linewidth]{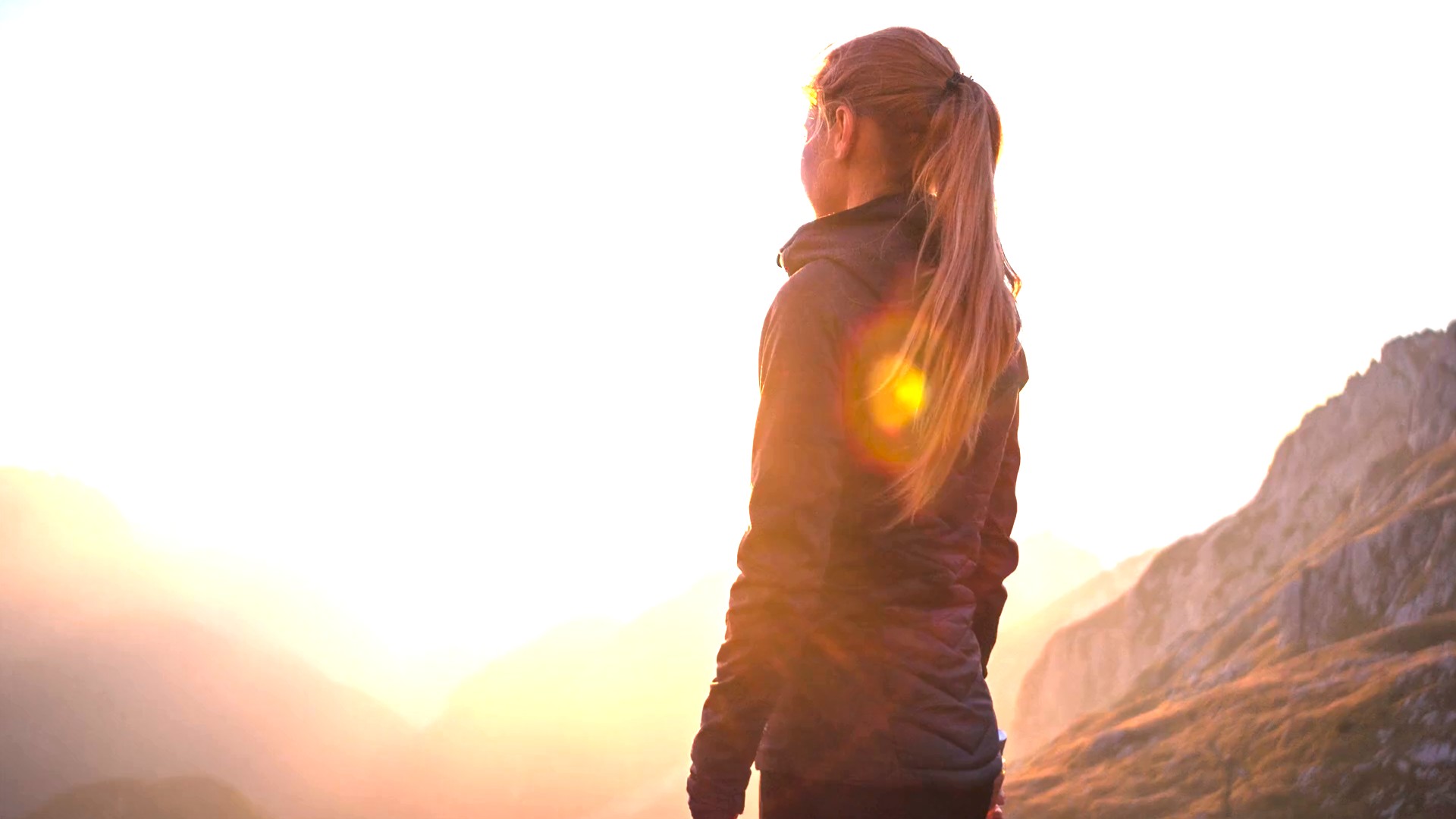} \\
      \raisebox{8pt}{\rotatebox{90}{Tedla et al.}} &
      \includegraphics[width=0.24\linewidth]{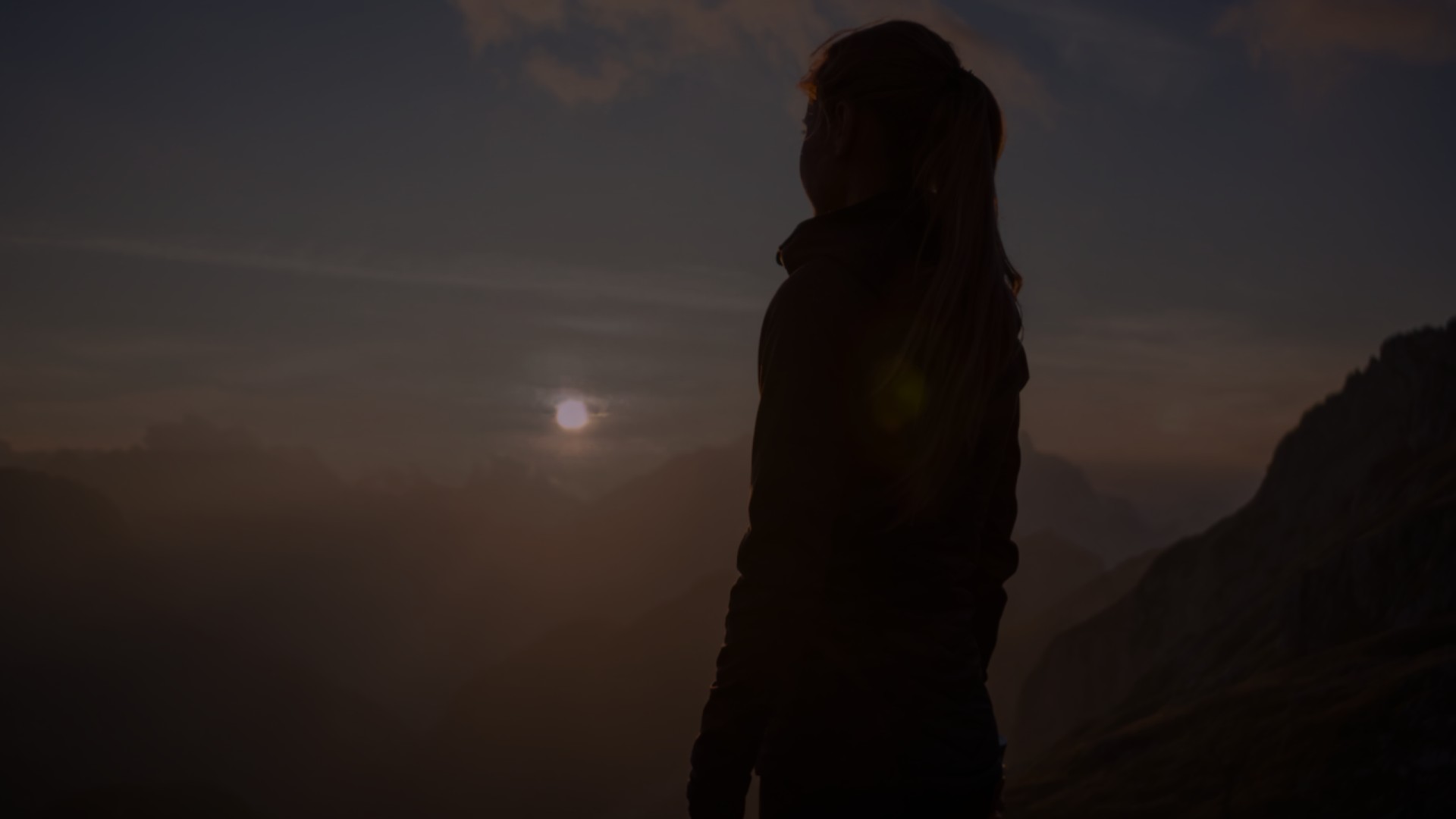} &
      \includegraphics[width=0.24\linewidth]{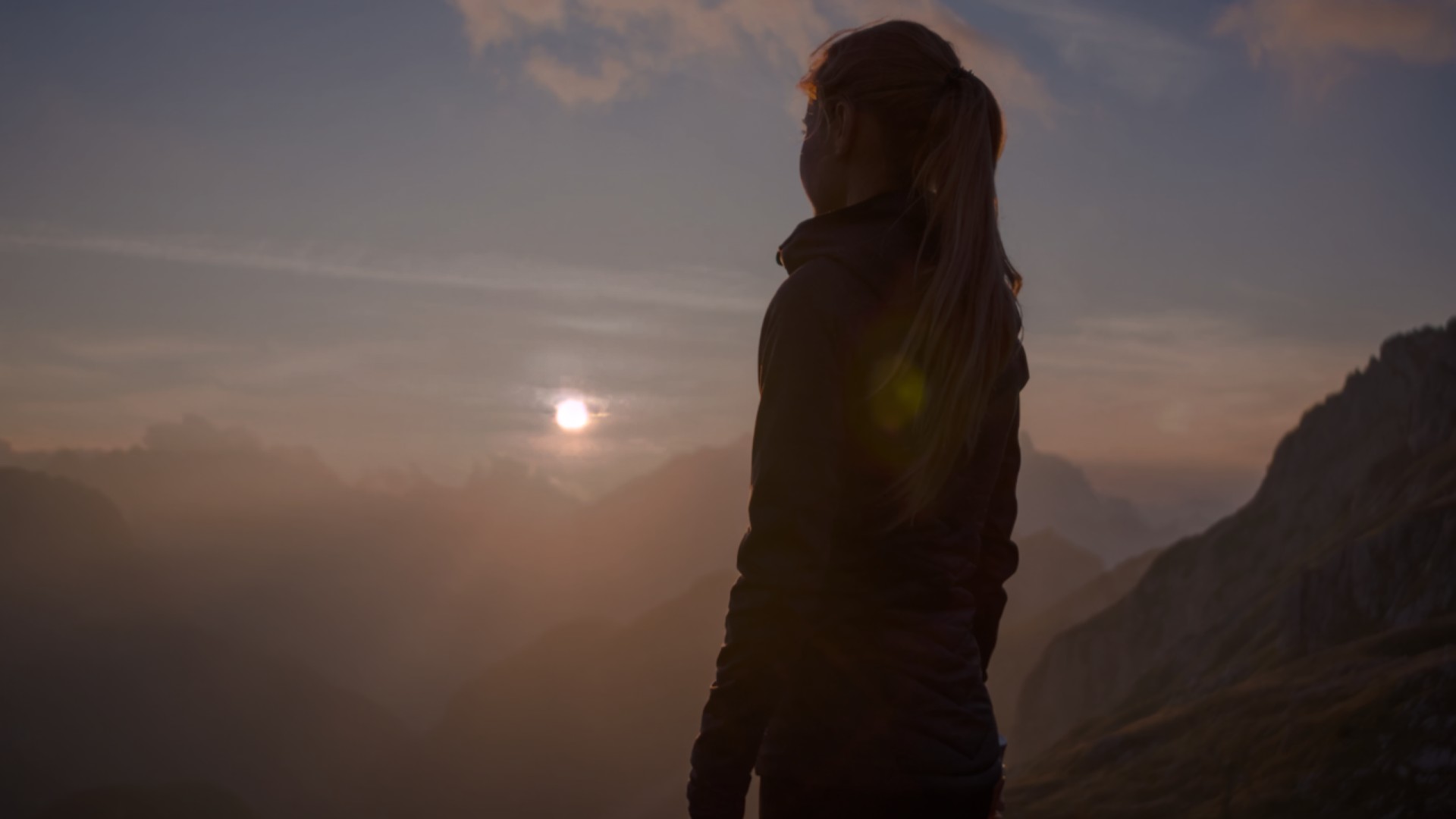} &
      \includegraphics[width=0.24\linewidth]{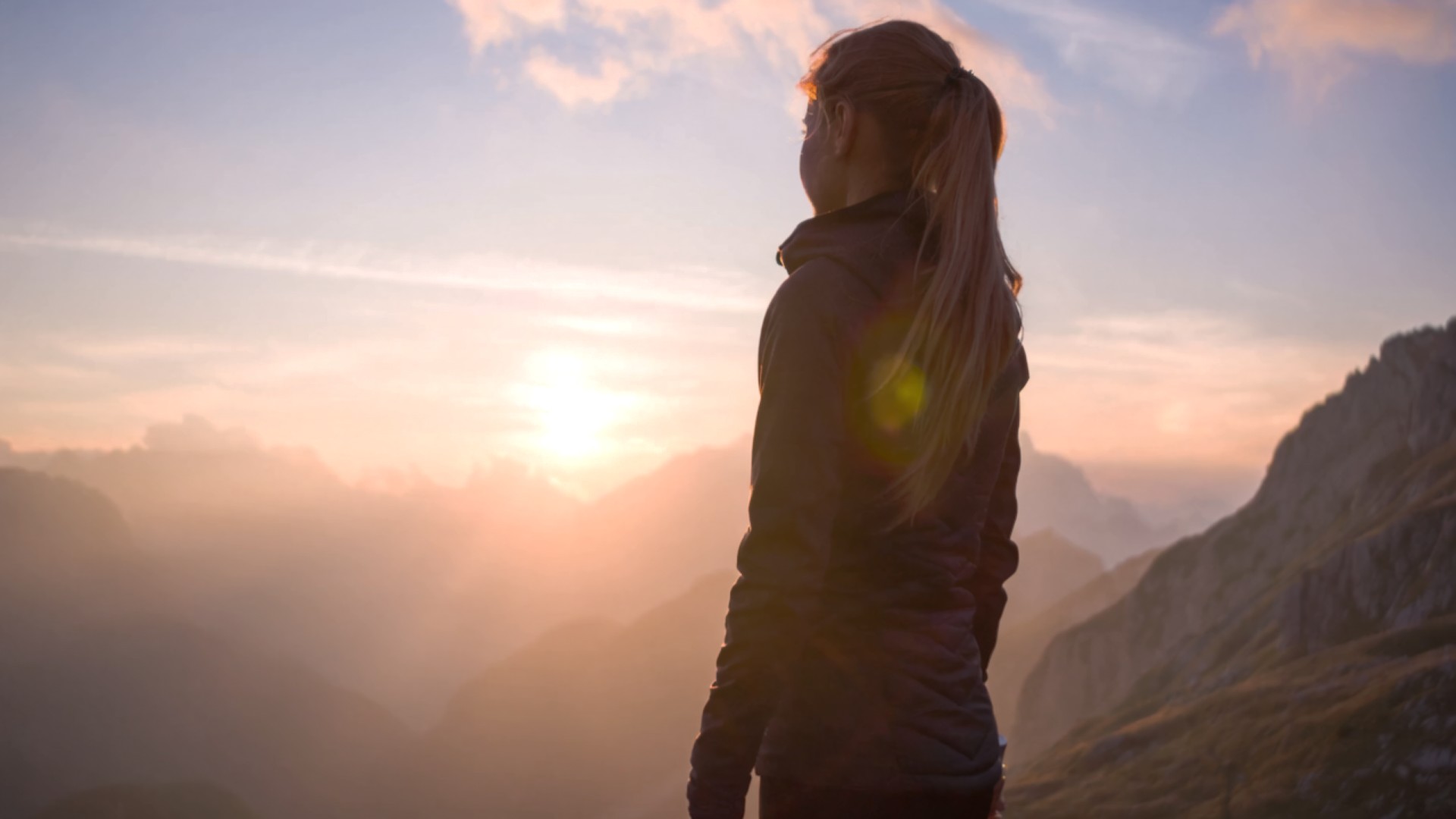} &
      \includegraphics[width=0.24\linewidth]{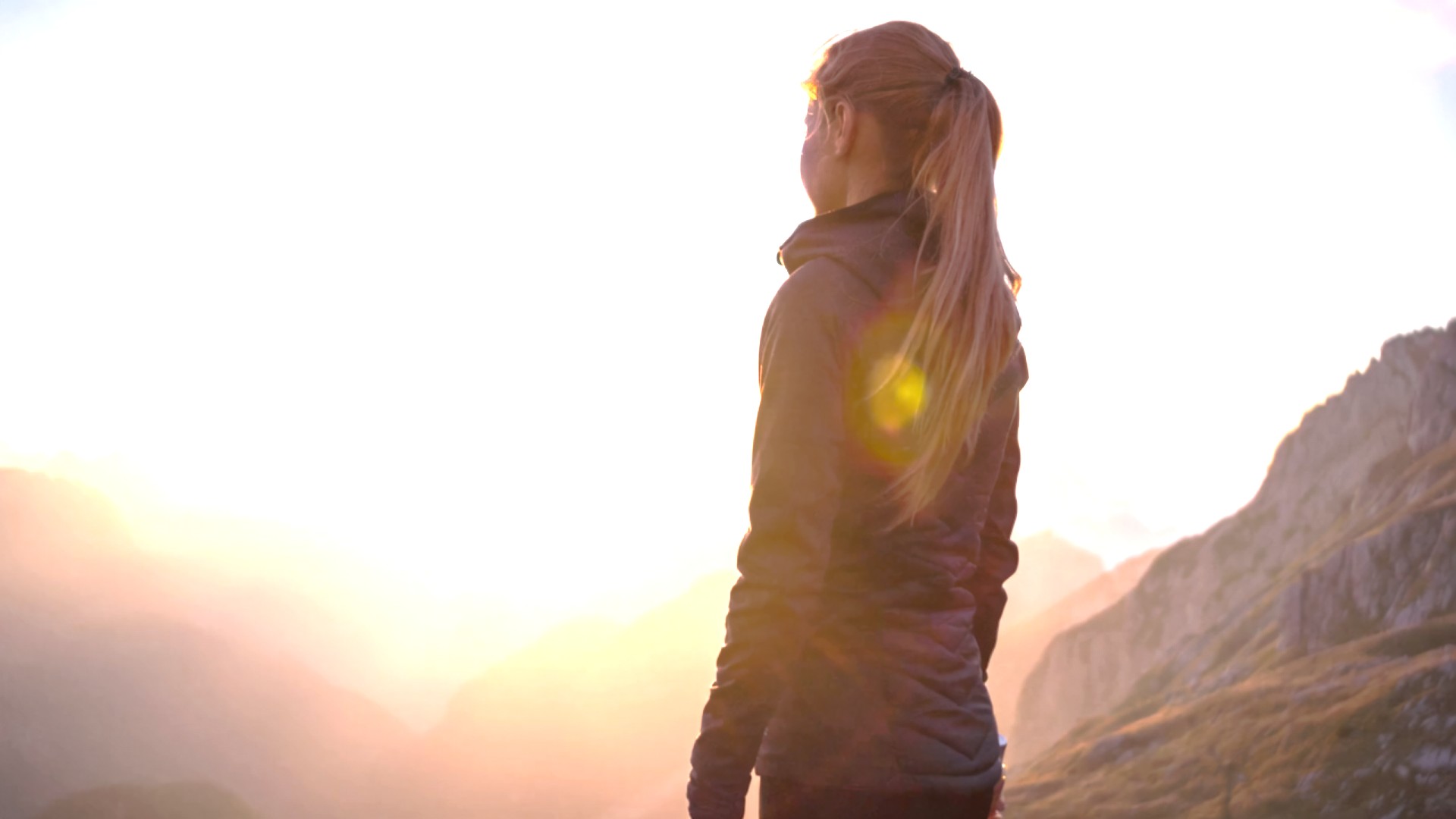} \\
      \raisebox{15pt}{\rotatebox{90}{\method{}}} &
      \includegraphics[width=0.24\linewidth]{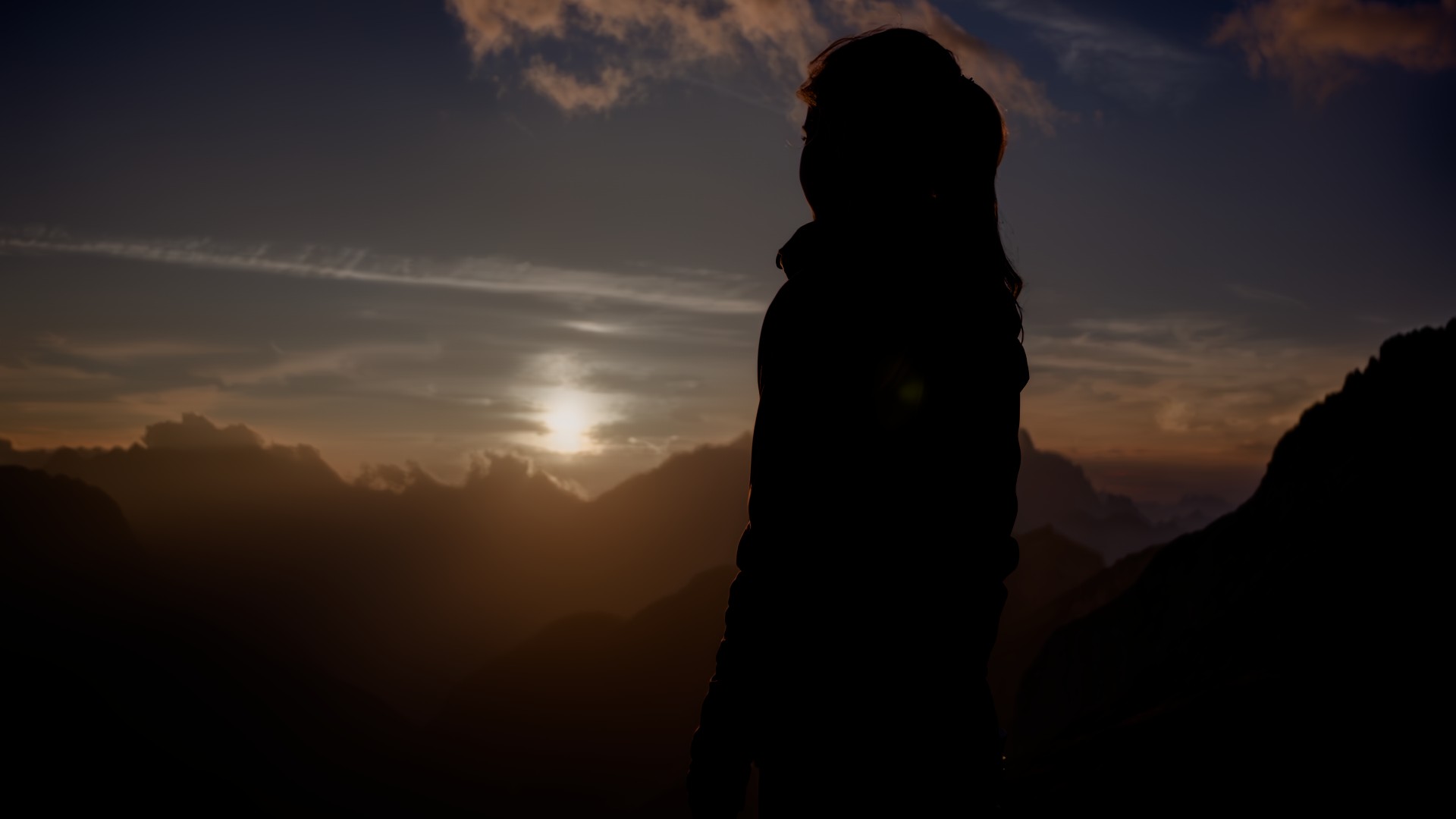} &
      \includegraphics[width=0.24\linewidth]{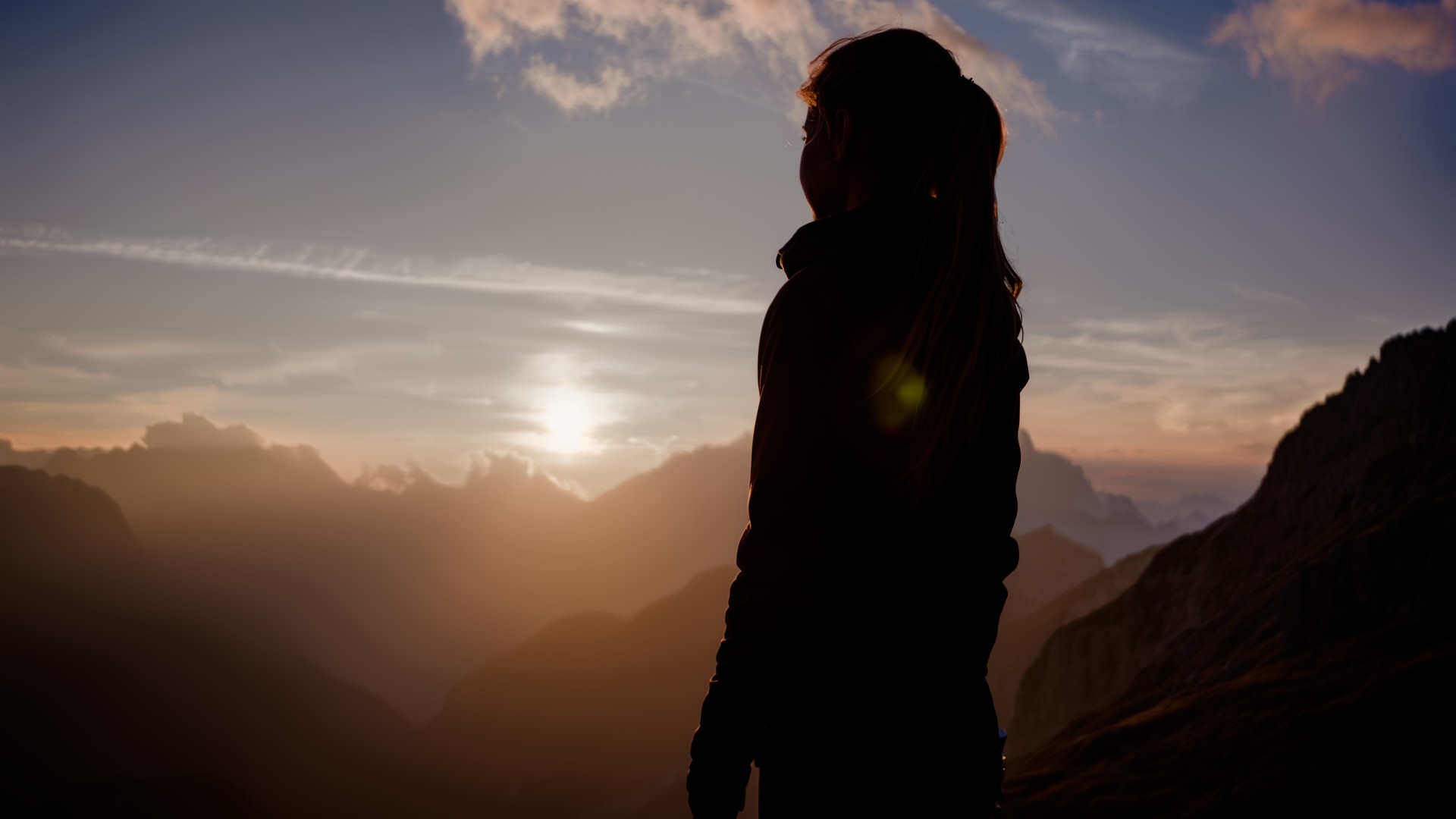} &
      \includegraphics[width=0.24\linewidth]{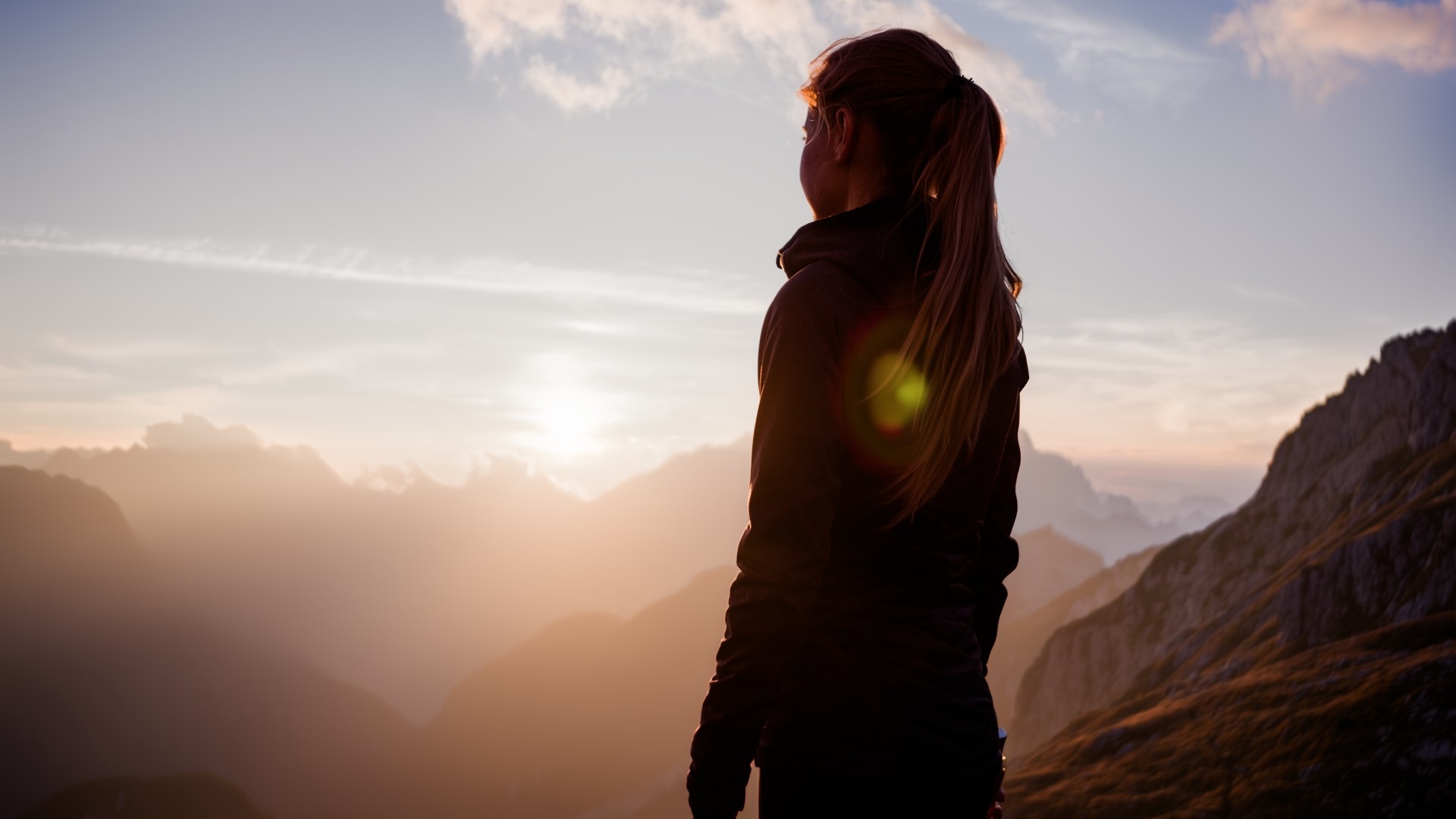} &
      \includegraphics[width=0.24\linewidth]{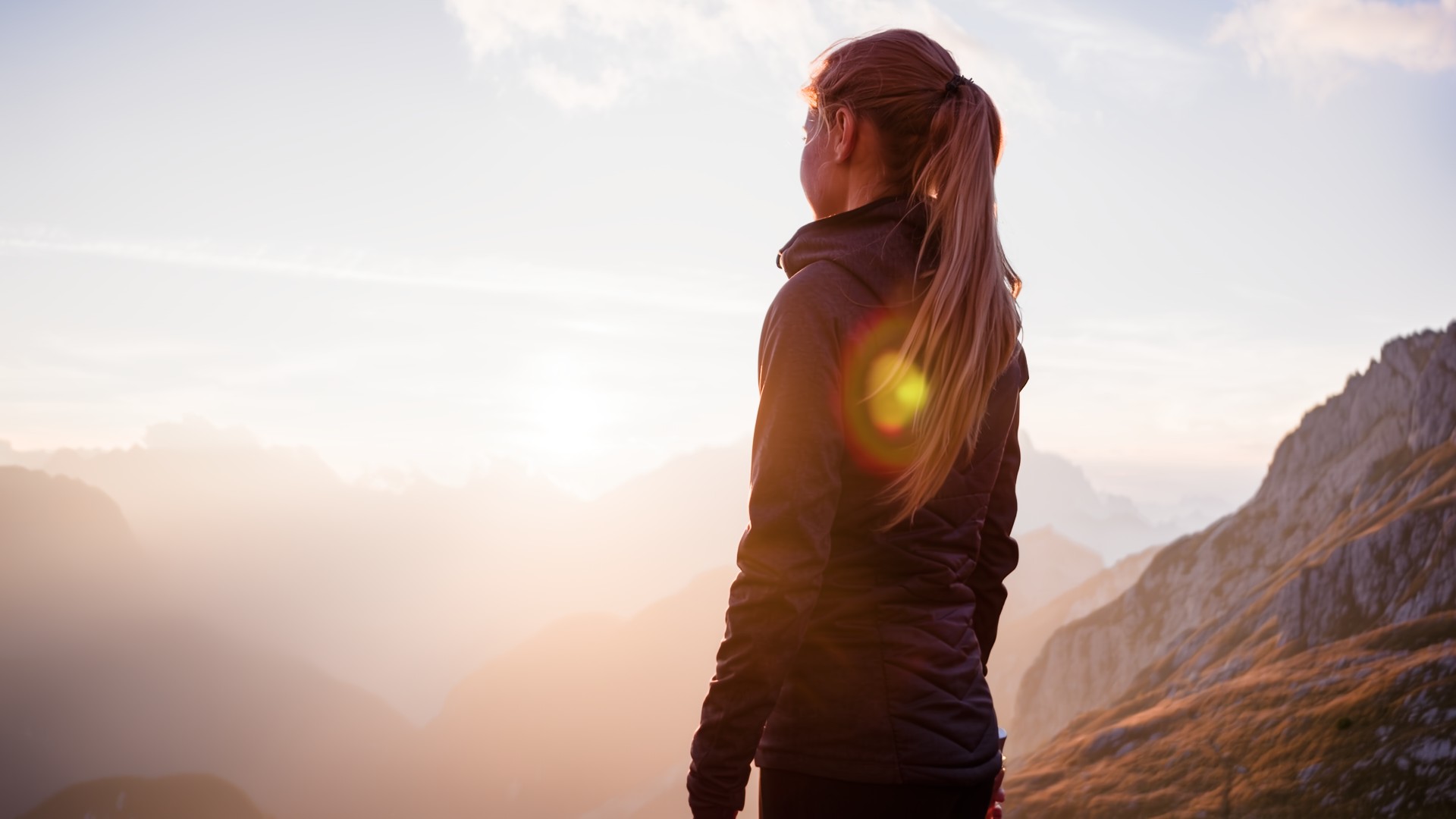} \\
      & EV $-4$ & EV $-2$ & EV $0$ & EV $+2$
      \end{tabular}
      \caption{HDR generation on the project-page sunset clip, tone-mapped at
      EV $-4$, $-2$, $0$, and $+2$.
      EV~$0$ matches the SDR look; negative EVs show recovered highlights and
      EV~$+2$ lifts the shadows.
      \method{} keeps a compact solar disk and the cloud structure around it.
      DiffHDR stays close to the SDR input (Table~\ref{tab:wild}: $Y_{99.9}{=}0.84$).
      Tedla et al. leaves a small pale disk
      at EV $-4$.}
      \label{fig:sunset_comparison}
\end{figure*}

\begin{figure*}
    \setlength{\tabcolsep}{0pt}
    \scriptsize
    \centering

    \begin{tabular}{ccc c ccc}

    EV -3 & EV 0 & EV +3 && EV -3 & EV 0 & EV +3 \\
    \includegraphics[width=0.165\linewidth]{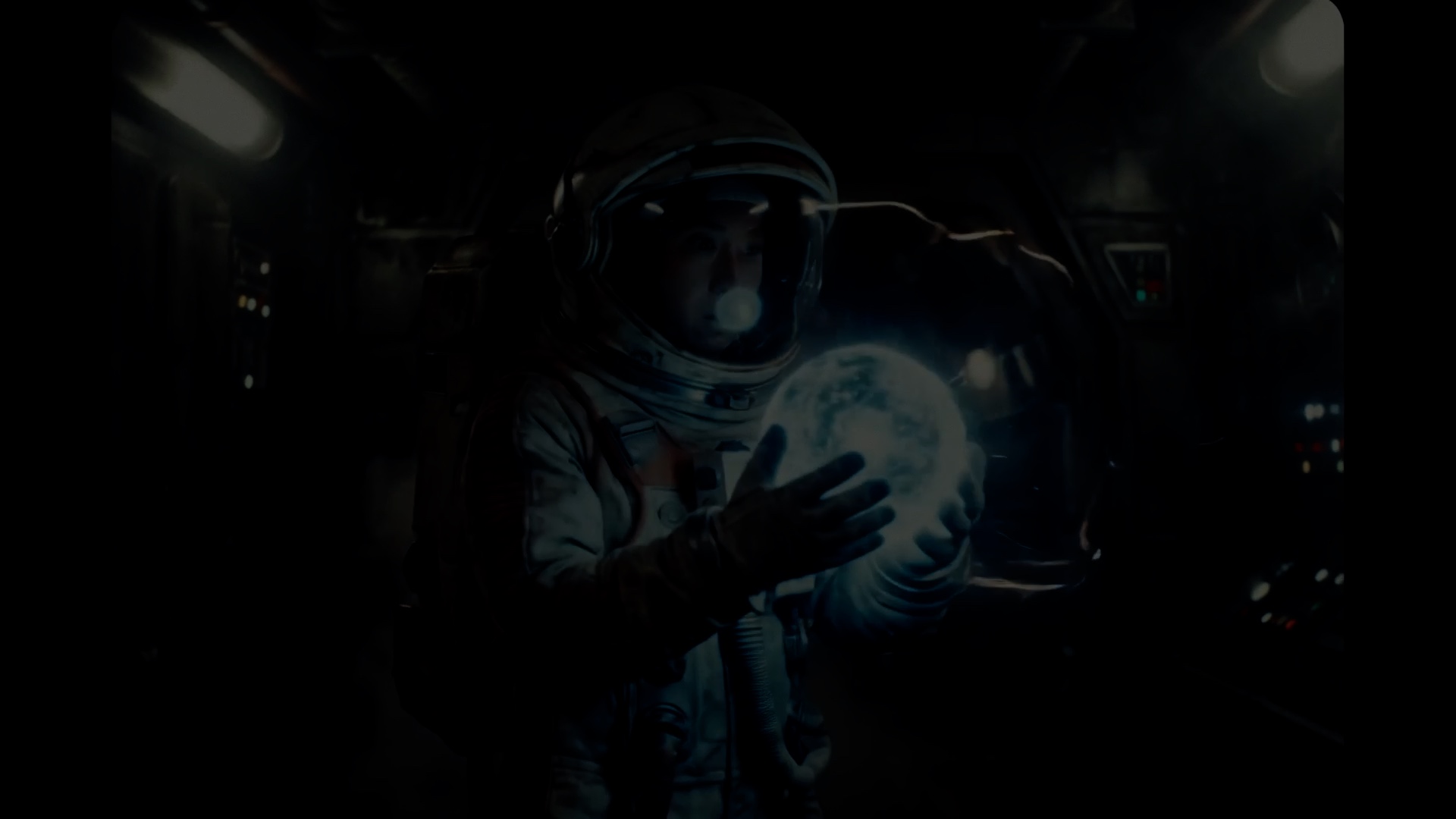} &
    \includegraphics[width=0.165\linewidth]{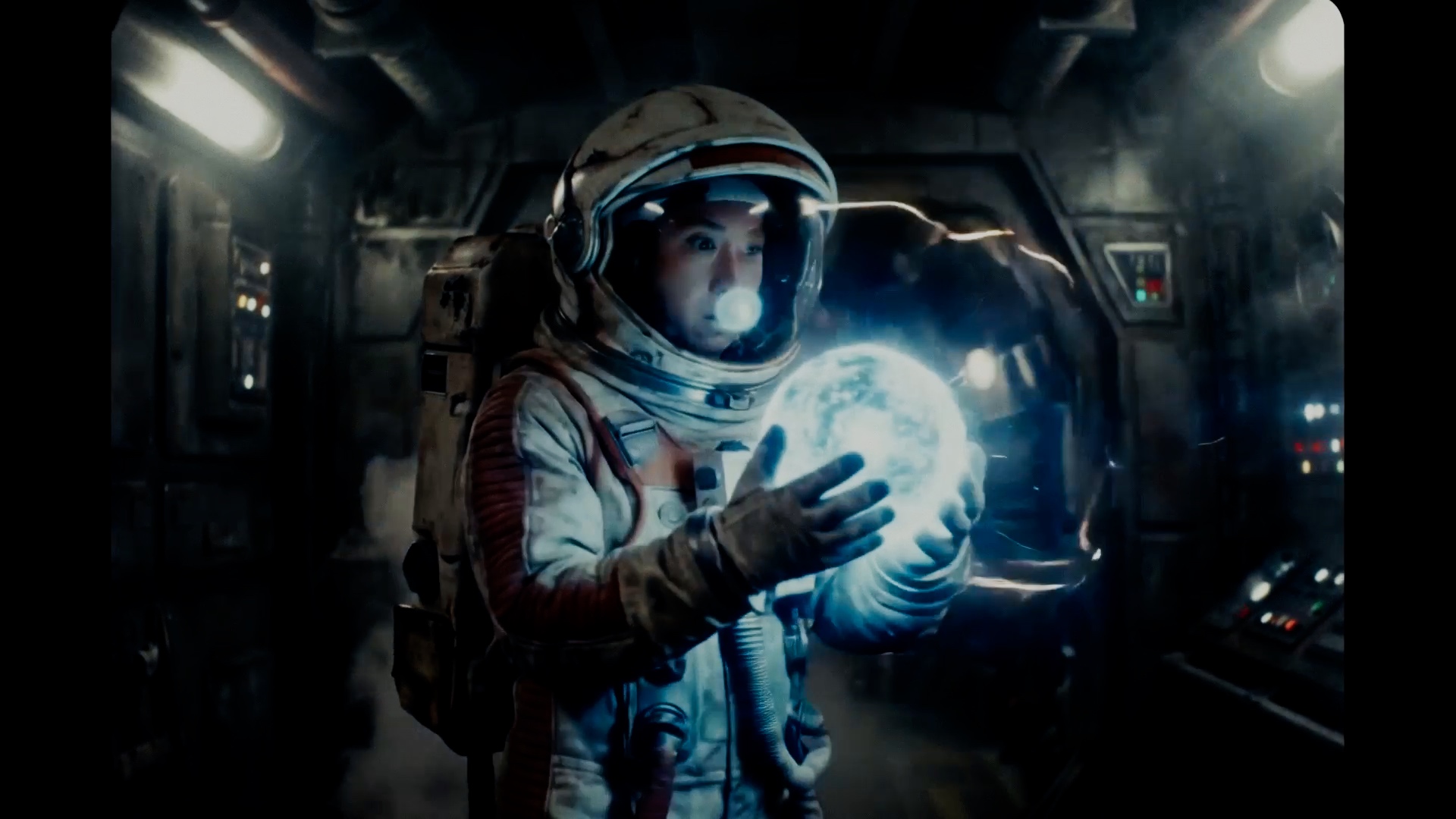} &
    \includegraphics[width=0.165\linewidth]{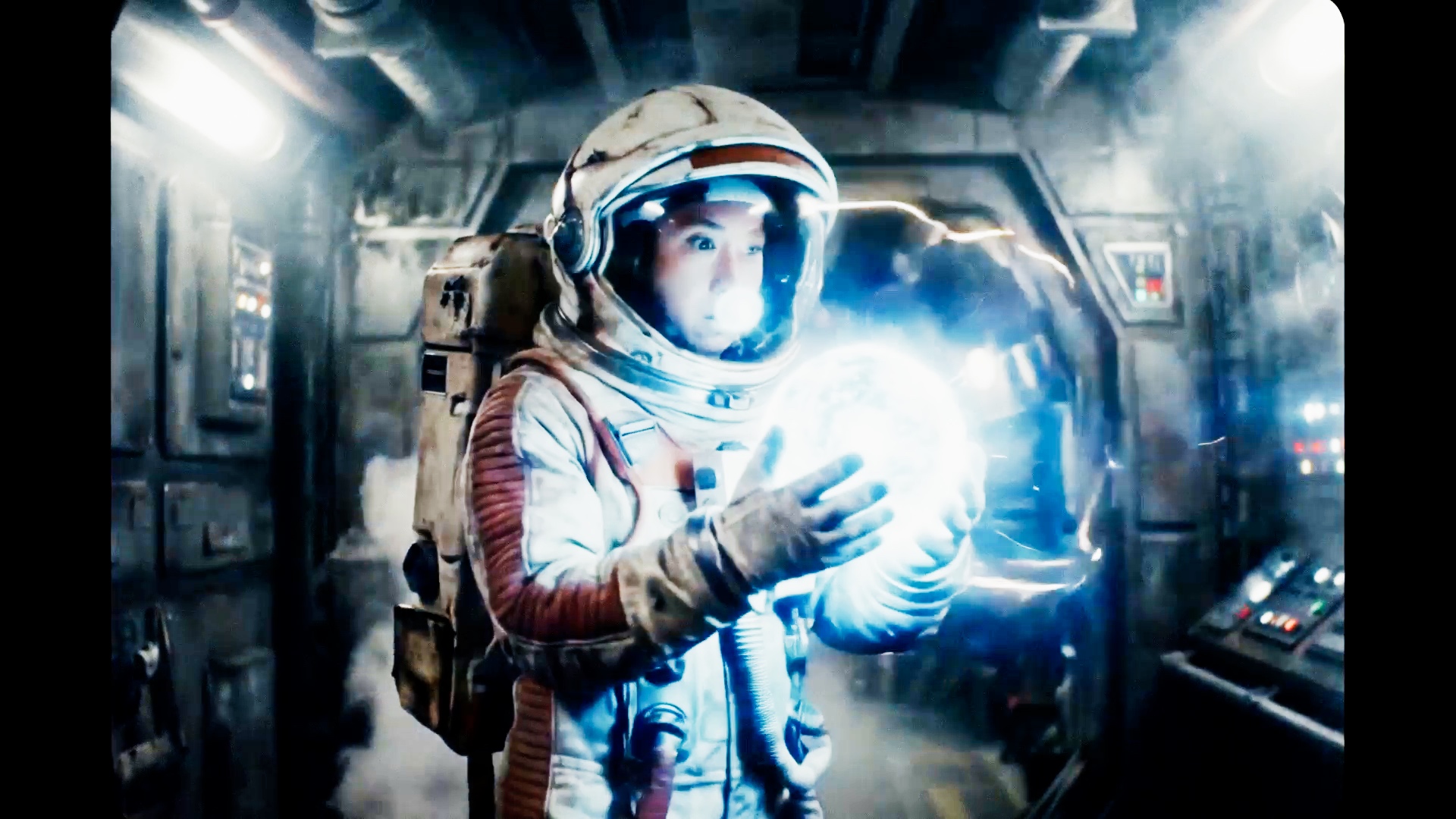} &
    { } &
    \includegraphics[width=0.165\linewidth]{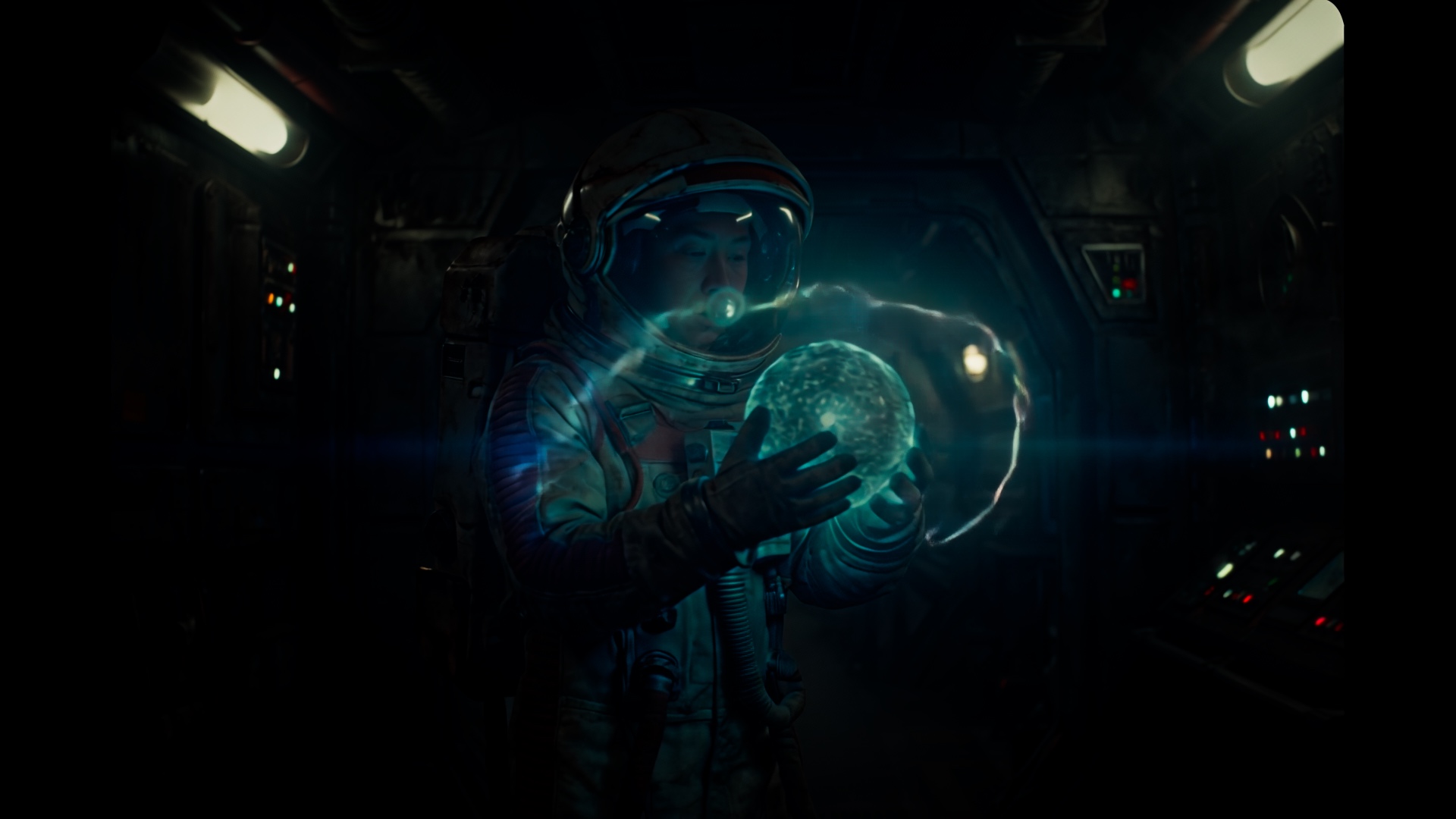} &
    \includegraphics[width=0.165\linewidth]{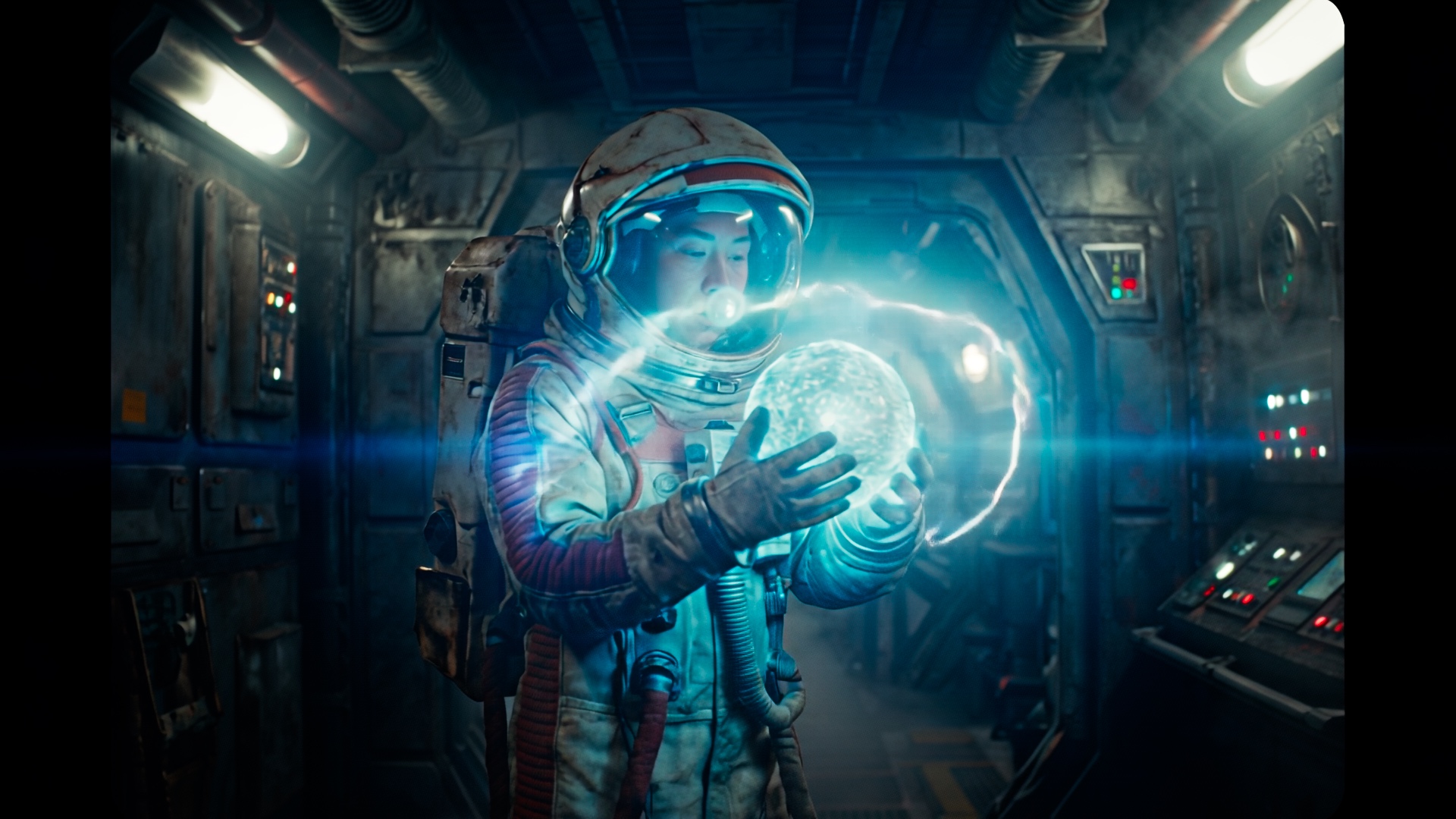} &
    \includegraphics[width=0.165\linewidth]{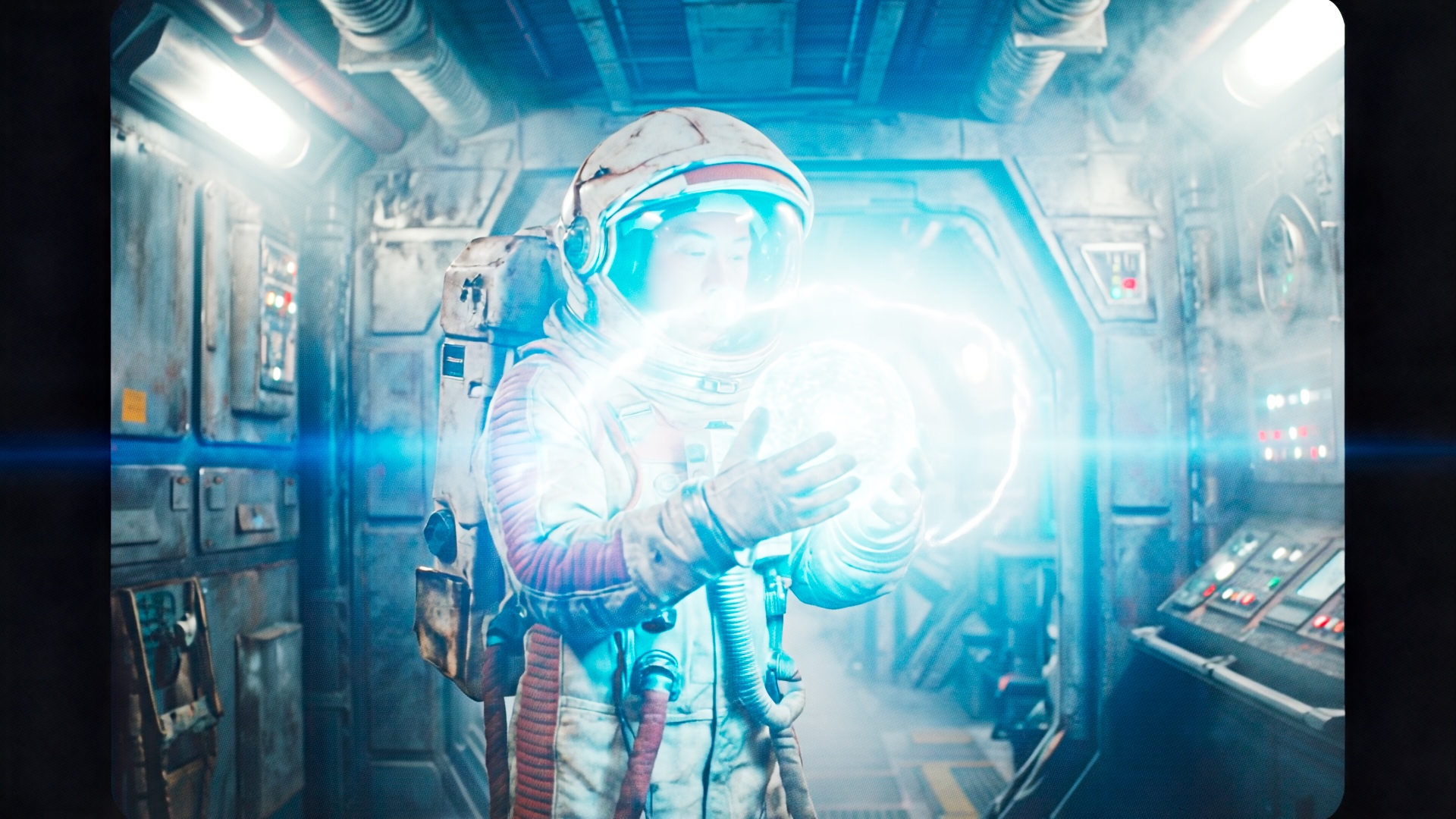} \\  
    EV -5 & EV -3 & EV 0 && EV -5 & EV -3 & EV 0 \\
    \includegraphics[width=0.165\linewidth]{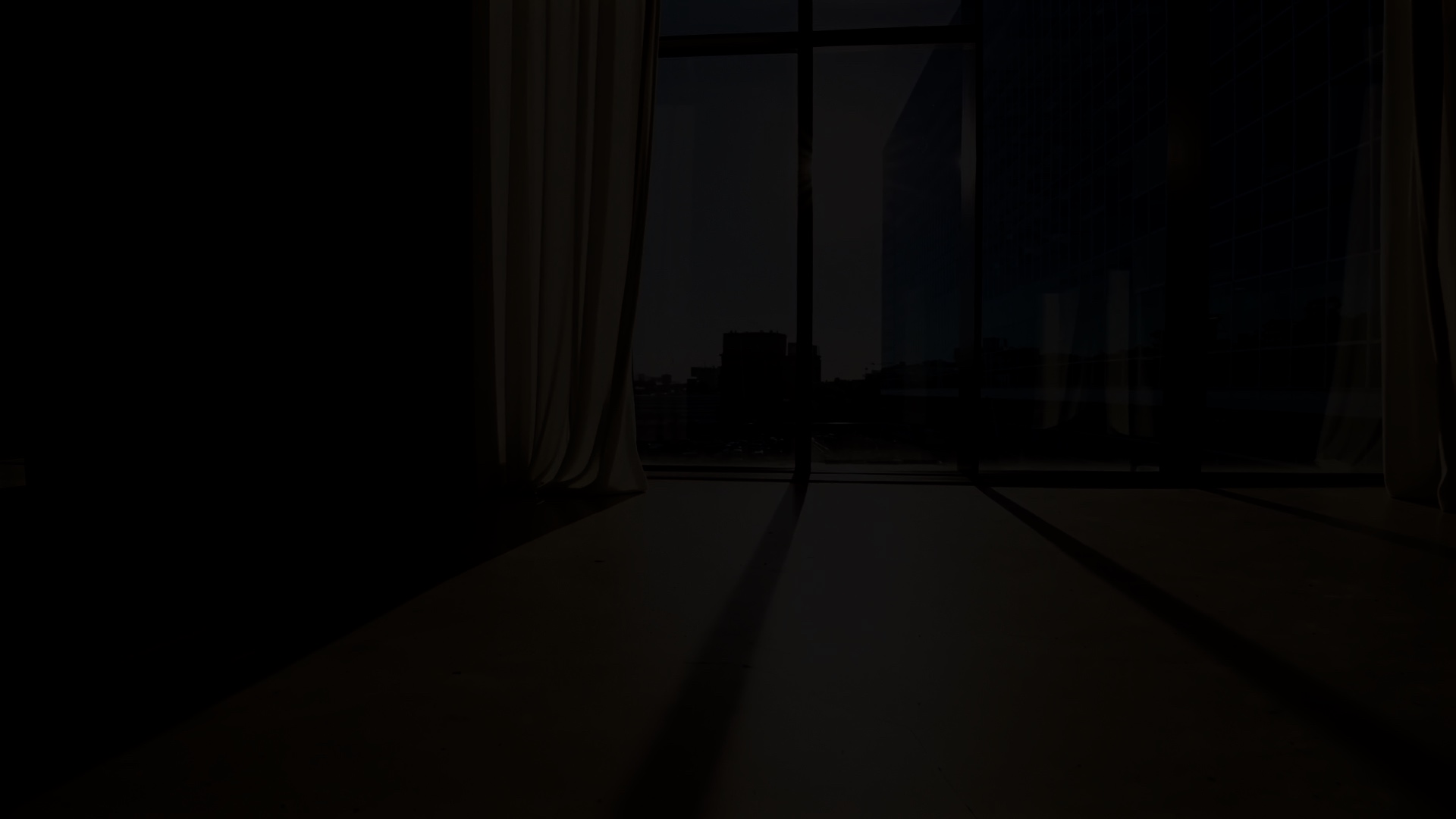} &
    \includegraphics[width=0.165\linewidth]{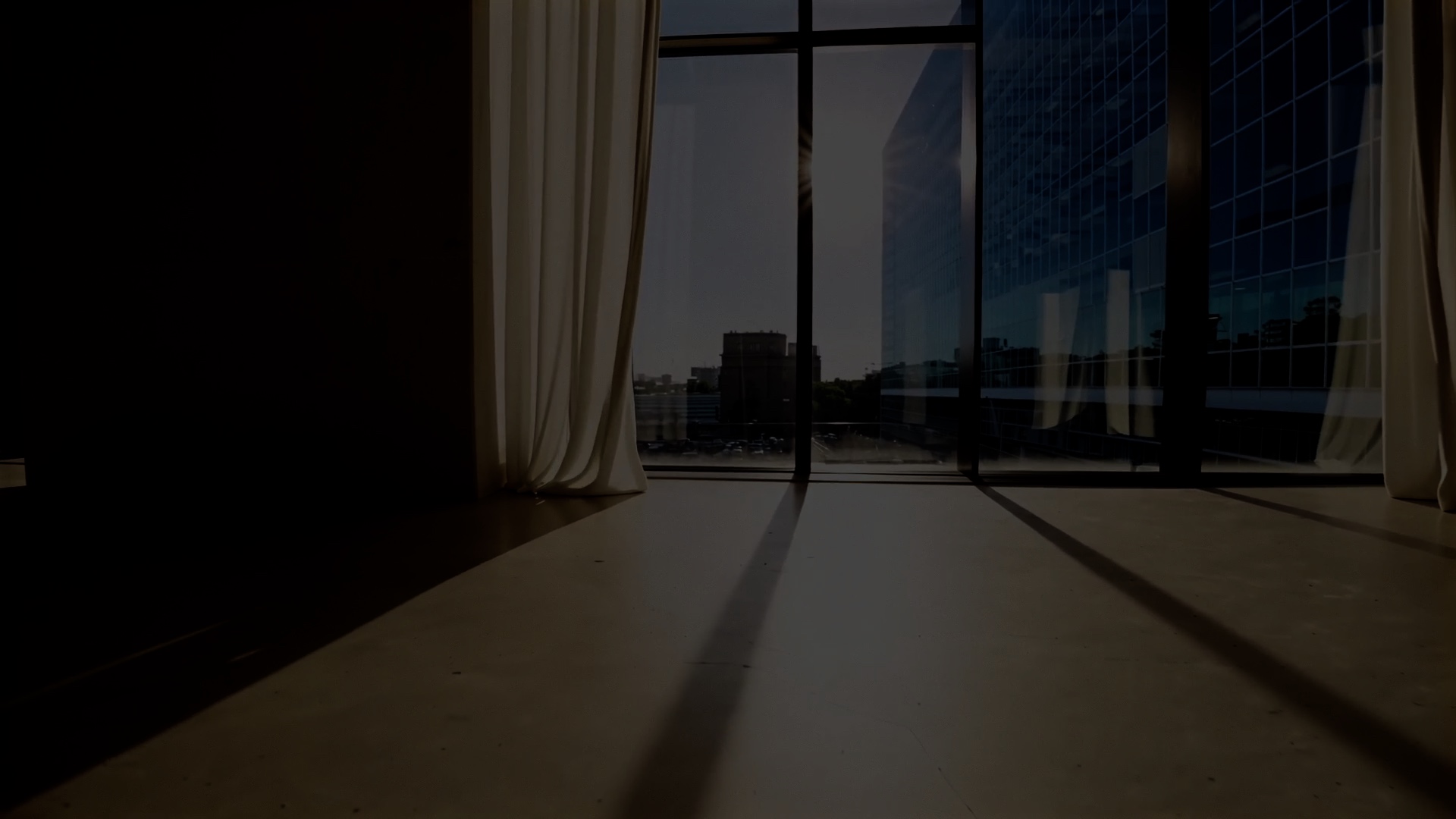} &
    \includegraphics[width=0.165\linewidth]{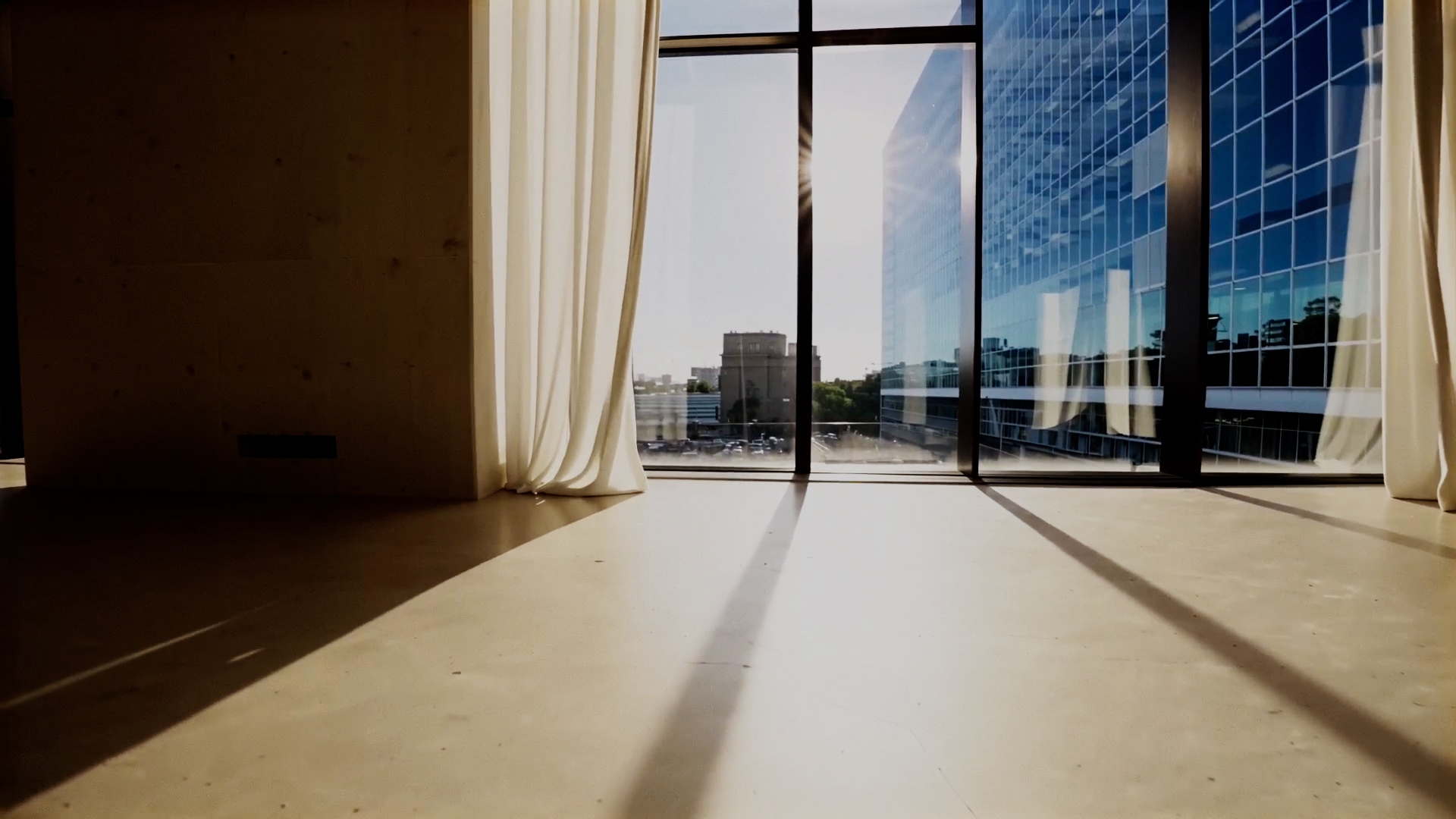} &
    { } &
    \includegraphics[width=0.165\linewidth]{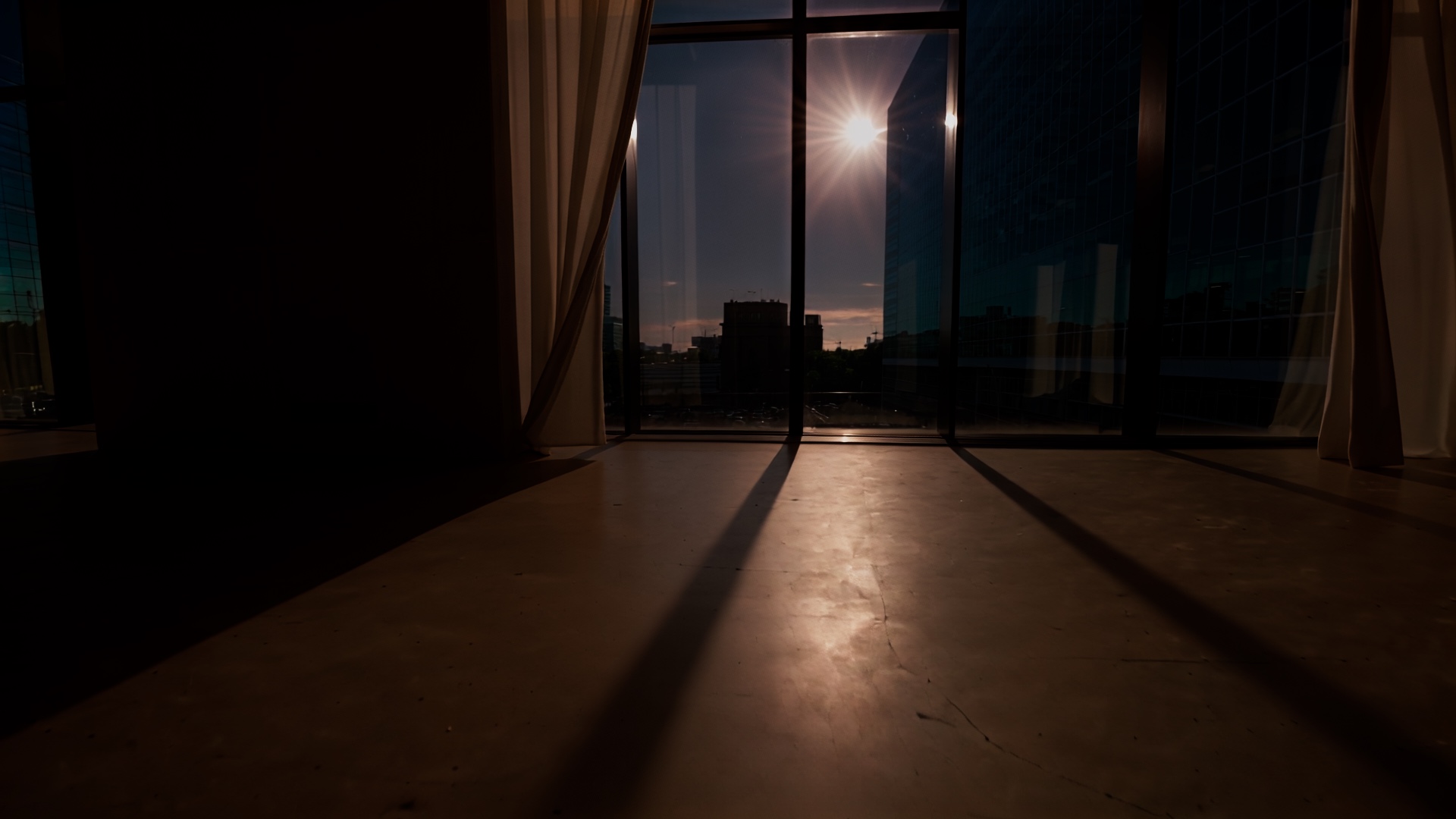} &
    \includegraphics[width=0.165\linewidth]{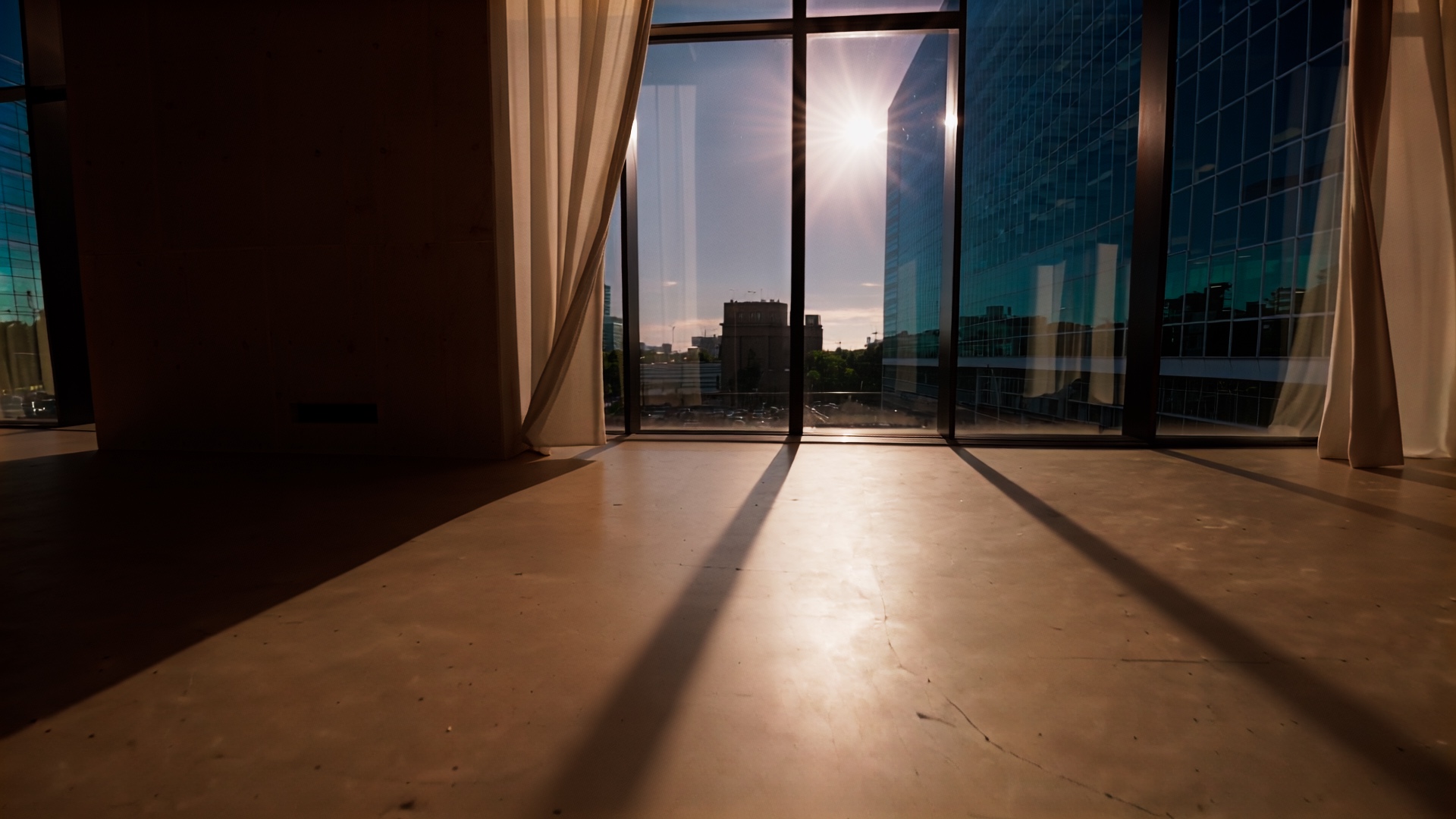} &
    \includegraphics[width=0.165\linewidth]{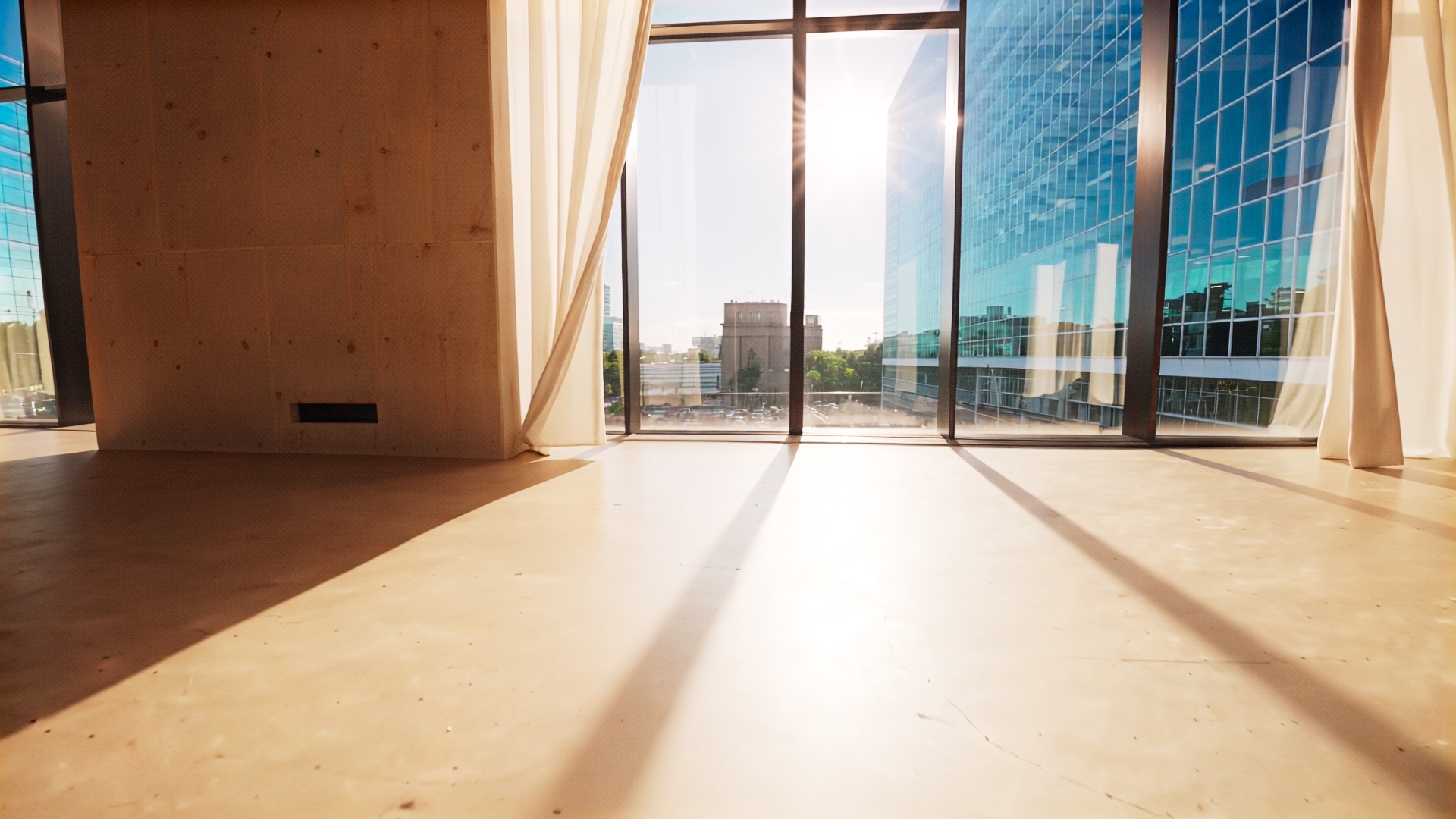}  \\  
    EV -4 & EV 0 & EV +4 && EV -4 & EV 0 & EV +4 \\
    \includegraphics[width=0.165\linewidth]{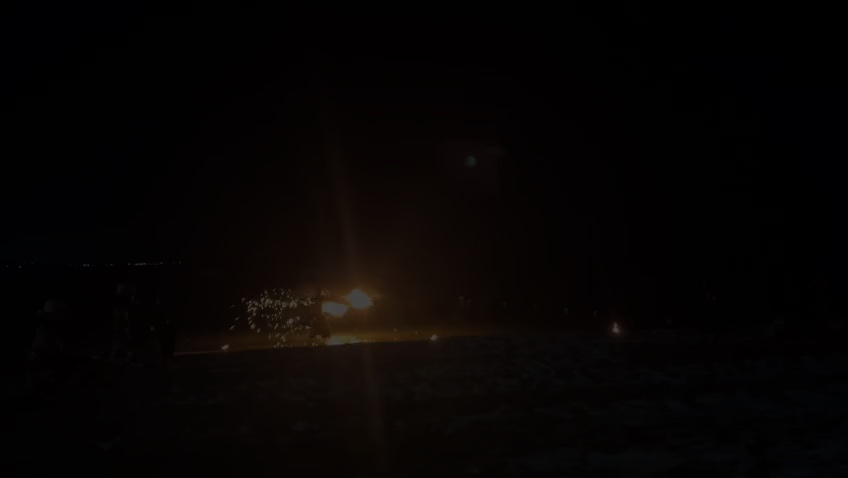} &
    \includegraphics[width=0.165\linewidth]{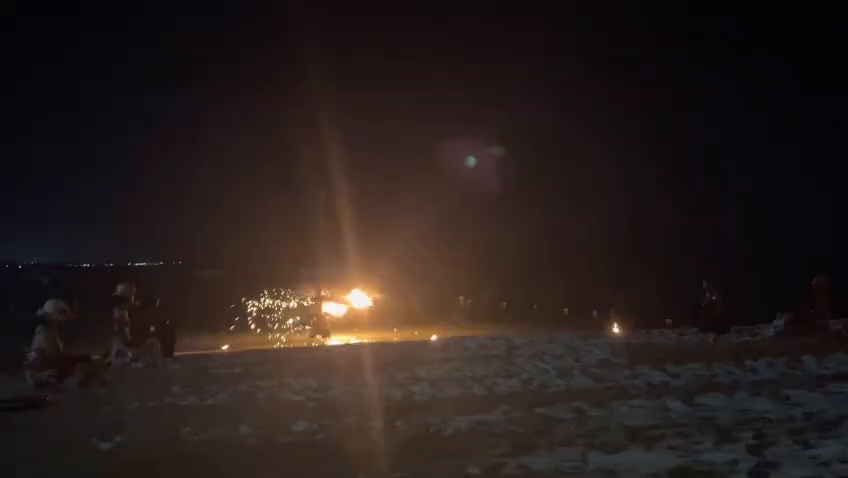} &
    \includegraphics[width=0.165\linewidth]{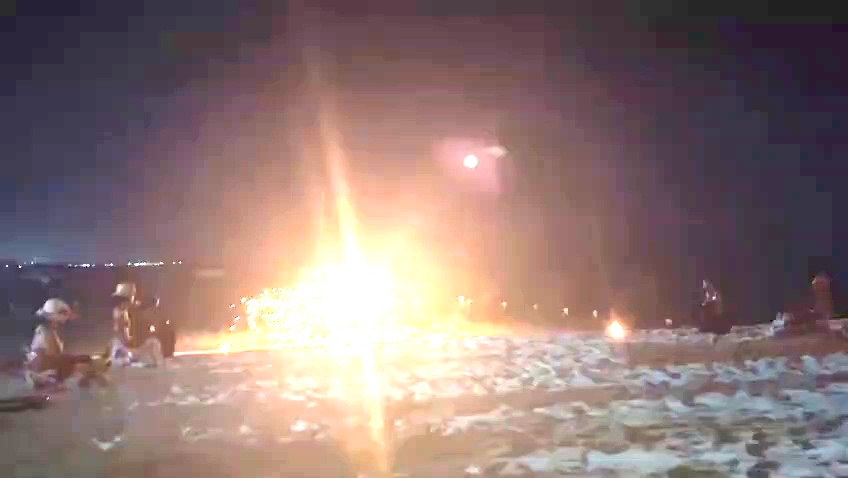} &
    { } &
    \includegraphics[width=0.165\linewidth]{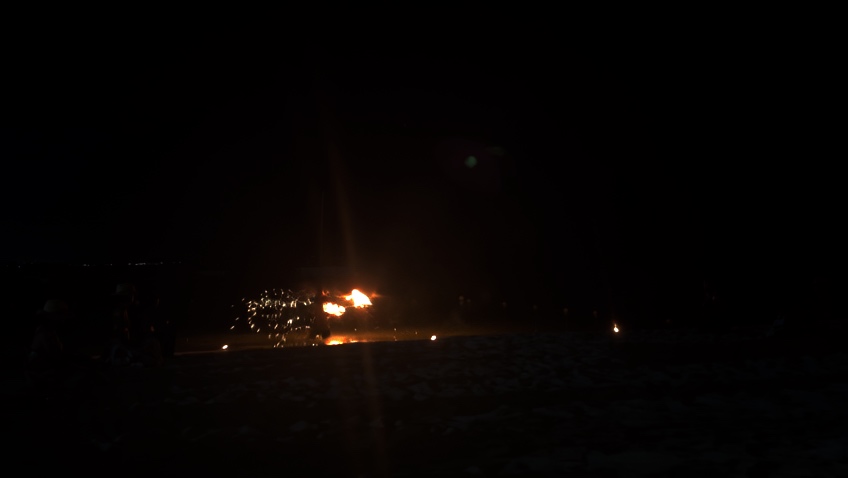} &
    \includegraphics[width=0.165\linewidth]{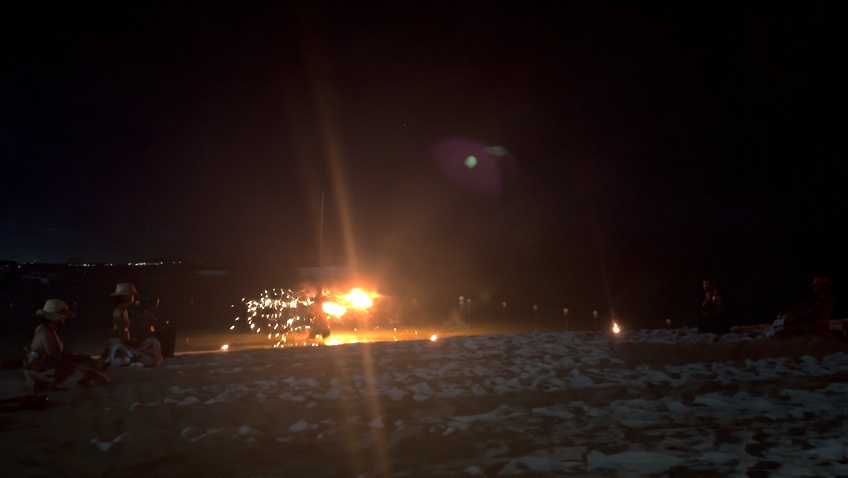} &
    \includegraphics[width=0.165\linewidth]{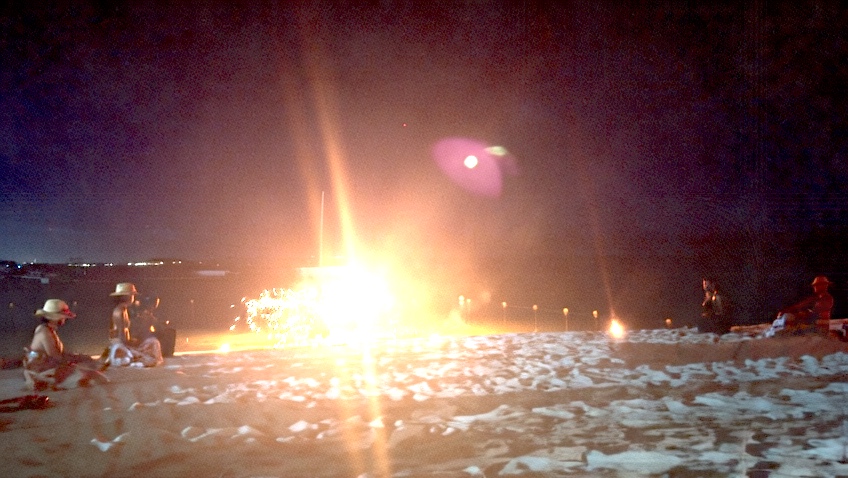}  \\ 

    \multicolumn{3}{c}{\normalsize{SDR (Input)}} && \multicolumn{3}{c}{\normalsize{HDR}}
    \end{tabular}

    \caption{
    LumiVid qualitative results: For each scene, we compare the input SDR (left) with our generated HDR (right) across multiple exposure values (EV). Our method effectively recovers high-frequency details and synthesizes plausible radiance in clipped regions.}
    \label{fig:video_results}
    
\end{figure*}

\subsection{Experimental Setup}
\label{sec:setup}

\textbf{Datasets.}
We evaluate \method{} on two benchmarks.
ARRI Cinema Benchmark ~\cite{arri_sample_footage} which consist of 50 professional cinema clips with authentic 12-bit HDR ground truth. Input SDR videos are generated by forward tone-mapping the raw HDR plates to 8-bit Rec.709. These clips contain real-world cinematography, including motion, and dynamic lighting transitions.
The second benchmark is In-the-Wild Benchmark: 66 diverse SDR video clips collected from YouTube. This benchmark tests generalization on real-world non-rigid motion (e.g., dynamic objects, human faces and motion, lighting transitions) without HDR ground truth. 

We compare \method{} against four state-of-the-art SDR-to-HDR baselines spanning both per-frame image models and native video diffusion backbones. 
\begin{enumerate}
    \item HDRTVNet++~\cite{chen2309towards}: A per-frame inverse tone-mapping network designed for SDRTV-to-HDRTV translation. Frames are processed independently and stacked into video sequences. HDRTVNet++ targets display-referred HDR10 delivery rather than 16-bit scene-linear EXR radiance files.
    \item X2HDR~\cite{x2hdr2025}: A generative single-image SDR-to-HDR diffusion model operating in PU21 perceptual space. Applied frame-by-frame to video sequences. 
    \item The closest native video baseline is DiffHDR~\cite{yu2026diffhdr}, employing a frozen VAE, log encoding, and masked radiance inpainting on a video DiT, based on a textual prompt. However, relying on luminance masks and text prompts reformulates the task into generative inpainting rather than direct SDR-to-HDR translation, adding manual overhead and risking visual inconsistency. 
    \item A concurrent SDR-to-HDR video diffusion method Tedla et al.~\cite{tedla2026generating} that fine-tunes Wan2.2 to predict linearized multi-exposure brackets for post merging. 
\end{enumerate}

All methods run at $720 \times 1280$ resolution. We run 49-frame on ARRI and 17-frame on in-the-wild clips, outputting Rec.709 scene-linear EXR files. Tedla et al., works only on 17 frames, thus when running inference on ARRI benchmark, we run three 17-frame windows, that are tiled at starting frame indices 0, 16, and 32 (overwriting 1-frame overlaps) to cover 49 frames.

\textbf{Evaluation Metrics.}
All generated EXR files are mapped to Rec.709 scene-linear space prior to metric calculation to ensure strict photometric alignment with the other baselines. We evaluate our method using common hdr video metrics, across four axes:
\begin{enumerate}
    \item{Ground-Truth Referenced Fidelity: Can be measured only in the ARRI benchmark. Standard PSNR on unbounded linear radiance over-penalizes absolute errors in extreme highlights where human visual sensitivity is low, while ignoring perceptually critical errors in the darker areas. We therefore report, PU21-PSNR / SSIM ~\cite{mantiuk2021pu21} that computes PSNR and SSIM after mapping linear radiance into the perceptually uniform PU21 space, ensuring errors in shadow and mid-tone regions, where the eye is far more sensitive, are properly reflected in the fidelity score. We also report ColorVideoVDP ~\cite{mantiuk2024colorvideovdp} measures perceptual video quality and visible artifacts by modeling human visual sensitivity across spatial, temporal, and chromatic dimensions across varying dynamic ranges, Outputting a score on a $0$--$10$ scale.}

    \item{Temporal Stability: To measure frame-to-frame coherence, we report F2F-PSNR, PSNR computed between consecutive PU21-encoded frames ($\mathbf{y}_t$ and $\mathbf{y}_{t+1}$). We also report $t$-LPIPS ~\cite{zhang2018lpips} - LPIPS between consecutive PU21 frames. Higher F2F-PSNR and lower $t$-LPIPS indicate stable, flicker-free video generation. }

    \item{No-Reference Quality (In-the-Wild): On clips without ground truth, we assess output quality using MUSIQ~\cite{ke2021musiq} a multi-scale transformer image quality assessment (IQA) model and CLIP-IQA+~\cite{wang2023exploring}. Since these models expect bounded 8-bit Rec.709 inputs, we evaluate them after applying explicit exposures. ($\mathrm{EV}{-3}$ to inspect recovered highlight detail and $\mathrm{EV}{+3}$ for shadow detail) followed by sRGB tone-mapping.}

    \item{Dynamic Range Expansion ($\text{Stops}$): To measure how far each method expands luminance above diffuse white, we compute scene-linear Rec.709 luminance $Y$ (where $Y=1.0$ represents diffuse white). In a scene-referred frame only a very small part of the pixels should reach extreme high values, a light, the sun, a specular, a sparkle, while the rest of the frame sits between $0-1$. We therefore read the tail with a high percentile rather than a statistic of the whole frame. We define the dynamic range  in stops as:
\begin{equation}
\text{Stops} = \log_2(Y_{99.9})
\label{eq:stops}
\end{equation}
where $Y_{99.9}$ is the 99.9th percentile of $Y$ over all pixels in a clip. The percentile is high enough to fall on that small highlight set.
We report the median over the $66$ clips.}
\end{enumerate}

Flicker, highlight recovery, and video quality are best judged in the
supplementary video and in a pairwise user study vs.\ HDRTVNet++, X2HDR and DiffHDR
(\S\ref{sec:user_study}).

\subsection{Results}
\label{sec:results_arri}

Table~\ref{tab:arri} reports ARRI GT-referenced fidelity and temporal stability. \method{} leads in PU21-PSNR, ColorVideoVDP, and F2F. While HDRTVNet++ achieves equal or slightly better results on $t$-LPIPS, PU21-SSIM, its PU21-PSNR is much worse, indicating that its output fails to achieve true HDR dynamic range. Qualitative crops at $\text{EV}{-3}$ (highlights) and $\text{EV}{+4}$ (shadows) are shown in Figure~\ref{fig:video_results}. In-the-wild baselines are presented in Figures~\ref{fig:car_comparison} and~\ref{fig:sunset_comparison}.

\begin{table}[t]
\centering
\caption{ARRI cinema, $n{=}50$ paired clips, 49 frames.
DiffHDR is $720{\times}1280$ (official crop); Tedla et al.\ is three 17-frame
$704{\times}1280$ windows (starts $0/16/32$). GT is area-downsampled to each
method. ColorVideoVDP is on a 1000-nit HDR display with 100 nits diffuse white.
second-best \underline{underlined}.}
\label{tab:arri}
\scriptsize
\setlength{\tabcolsep}{3.2pt}
\begin{tabular}{@{}lccccc@{}}
\toprule
Method & PU21-PSNR $\uparrow$ & PU21-SSIM $\uparrow$ & ColorVideoVDP $\uparrow$ & F2F $\uparrow$ & t-LPIPS $\downarrow$ \\
\midrule
HDRTVNet++ & 22.43 & \textbf{0.935} & \underline{7.29} & \underline{44.81} & \textbf{0.0057} \\
X2HDR & 18.27 & 0.818 & 3.19 & 44.60 & 0.0143 \\
DiffHDR & \underline{24.28} & 0.743 & 4.57 & 40.36 & 0.0075 \\
Tedla et al.~\cite{tedla2026generating} & 16.80 & 0.687 & 4.77 & 43.59 & 0.0070 \\
\textbf{\method{}} & \textbf{29.79} & \underline{0.915} & \textbf{7.92} & \textbf{54.28} & \underline{0.0059} \\
\bottomrule
\end{tabular}
\end{table}

Table~\ref{tab:wild} scores the in-the-wild benchmark. F2F / t-LPIPS for native
video; MUSIQ and CLIP-IQA+ at EV $-3$ / $+3$ for highlights and shadows.
\method{} leads those columns.

\method{} sits $+2.8$~stops above white at $Y_{99.9}$,  ($6.7\times$) much higher than the other baselines. HDRTVNet++ is a mild display lift ($1.7\times$).
Flicker, highlight recovery, and the exposure pulls are on the project page. The comparisons are best viewed on an HDR display.

\begin{table}[t]
\centering
\caption{In-the-wild SDR videos.
MUSIQ / CLIP-IQA+ at EV $-3$ (highlights); CLIP-IQA+$_{+3}$ at EV $+3$ (shadows).
Range: Rec.709 linear $Y$ ($1$ is white), median over clips.
Best in bold.}
\label{tab:wild}
\scriptsize
\setlength{\tabcolsep}{1.8pt}
\begin{tabular}{@{}lccccc cc@{}}
\toprule
& \multicolumn{5}{c}{Quality} & \multicolumn{2}{c}{Range} \\
\cmidrule(lr){2-6} \cmidrule(lr){7-8}
Method & F2F $\uparrow$ & t-LPIPS $\downarrow$ & MUSIQ$_{-3}$ $\uparrow$ & CLIP$_{-3}$ $\uparrow$ & CLIP$_{+3}$ $\uparrow$ & $Y_{99.9}$ $\uparrow$ & Stops $\uparrow$ \\
\midrule
HDRTVNet++ & 37.72 & 0.065 & 43.6 & 0.412 & 0.355 & 1.66 & $+0.73$ \\
X2HDR & 38.19 & 0.068 & 44.4 & 0.426 & 0.352 & 2.06 & $+1.04$ \\
DiffHDR & 38.84 & 0.065 & 40.8 & 0.425 & 0.385 & 0.84 & $-0.26$ \\
Tedla et al.~\cite{tedla2026generating} & 36.55 & 0.062 & 38.2 & 0.403 & 0.383 & 0.73 & $-0.45$ \\
\textbf{\method{}} & \textbf{45.15} & \textbf{0.059} & \textbf{46.3} & \textbf{0.434} & \textbf{0.417} & \textbf{6.74} & $\mathbf{+2.75}$ \\
\bottomrule
\end{tabular}
\end{table}

\subsection{Perceptual User Study}
\label{sec:user_study}

\begin{figure}[t]
\centering
\includegraphics[width=\linewidth]{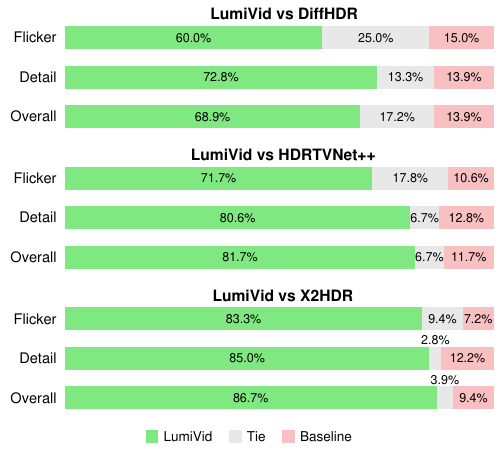}
\caption{Pairwise preference on flicker, detail, and overall quality.
Twenty one raters compared \method{} to each baseline on ten clips drawn at random
from the ARRI and in-the-wild test sets, on a 1000-nit MacBook display.}
\label{fig:user_study}
\end{figure}

Full-reference metrics like PU21-PSNR and ColorVideoVDP do not fully capture subjective visual quality or subtle temporal flickering. Thus, we conducted a pairwise 2AFC (Two-Alternative Forced Choice) perceptual user study comparing LumiVid against DiffHDR, X2HDR, and HDRTVNet++. A total of twenty one raters evaluated 30 video pairs each (10 scene pairs per baseline).
The study evaluated three key dimensions: (1) temporal stability, color preservation, and flicker; (2) spatial detail and highlight texture recovery; and (3) overall visual quality. The EXR files were displayed as hlg files in three exposures, -3, 0, 3. To ensure good visual evaluation, all ratings were collected on a 1000-nit MacBook Pro display operating in native HDR mode. Figure~\ref{fig:user_study} is flicker, detail, and overall quality.
On this HDR display, raters prefer \method{} to every baseline.

\subsection{Ablation Studies}
\label{sec:ablations}

We conduct ablation experiments to isolate two design choices in our framework: 
(1) whether the generative diffusion prior of the Video DiT is necessary for high quality HDR recovery, or if a fine-tuned VAE decoder is sufficient; and 
(2) whether camera-mimicking condition degradations are required to prevent shortcut copying of 8-bit SDR pixel structures.

\textbf{Isolating the Video DiT Backbone (VAE Decoder Ablation):}
A fundamental question is whether SDR-to-HDR video translation requires a generative diffusion process, or if it can be solved as a lightweight latent-to-pixel decoding task. Training a standalone encoder-decoder network from scratch would discard the rich representations learned by the pretrained VAE model and demand large-scale HDR video datasets that do not exist. Operating directly in the VAE's latent space preserves these representations.

Furthermore, as established in \S\ref{sec:acescct}, mapping scene-linear HDR into ACEScct allows the frozen VAE to represent log radiance without altering the latent manifold expected by the Video DiT (Table~\ref{tab:vae_roundtrip}). This raises a compelling question: if the frozen SDR VAE encoder already produces latents that map well to ACEScct, can we simply fine-tune a VAE decoder to decode those latents directly into HDR frames in a single feed-forward pass?

To test this, we construct a VAE decoder-only baseline. We fine-tune the pixel decoder directly on ACEScct targets while keeping the VAE encoder frozen, completely omitting the Video DiT. At inference, SDR video is encoded and decoded to HDR in a single pass. This isolates whether the Video DiT is required for temporal consistency and high-magnitude radiance synthesis, or if log encoding plus a fine-tuned pixel decoder is sufficient.
\begin{figure*}[t]
\setlength{\tabcolsep}{1pt}
\centering
\small
\begin{tabular}{ccccc}
\raisebox{8pt}{\rotatebox{90}{Input (SDR)}} &
  \includegraphics[width=0.22\linewidth]{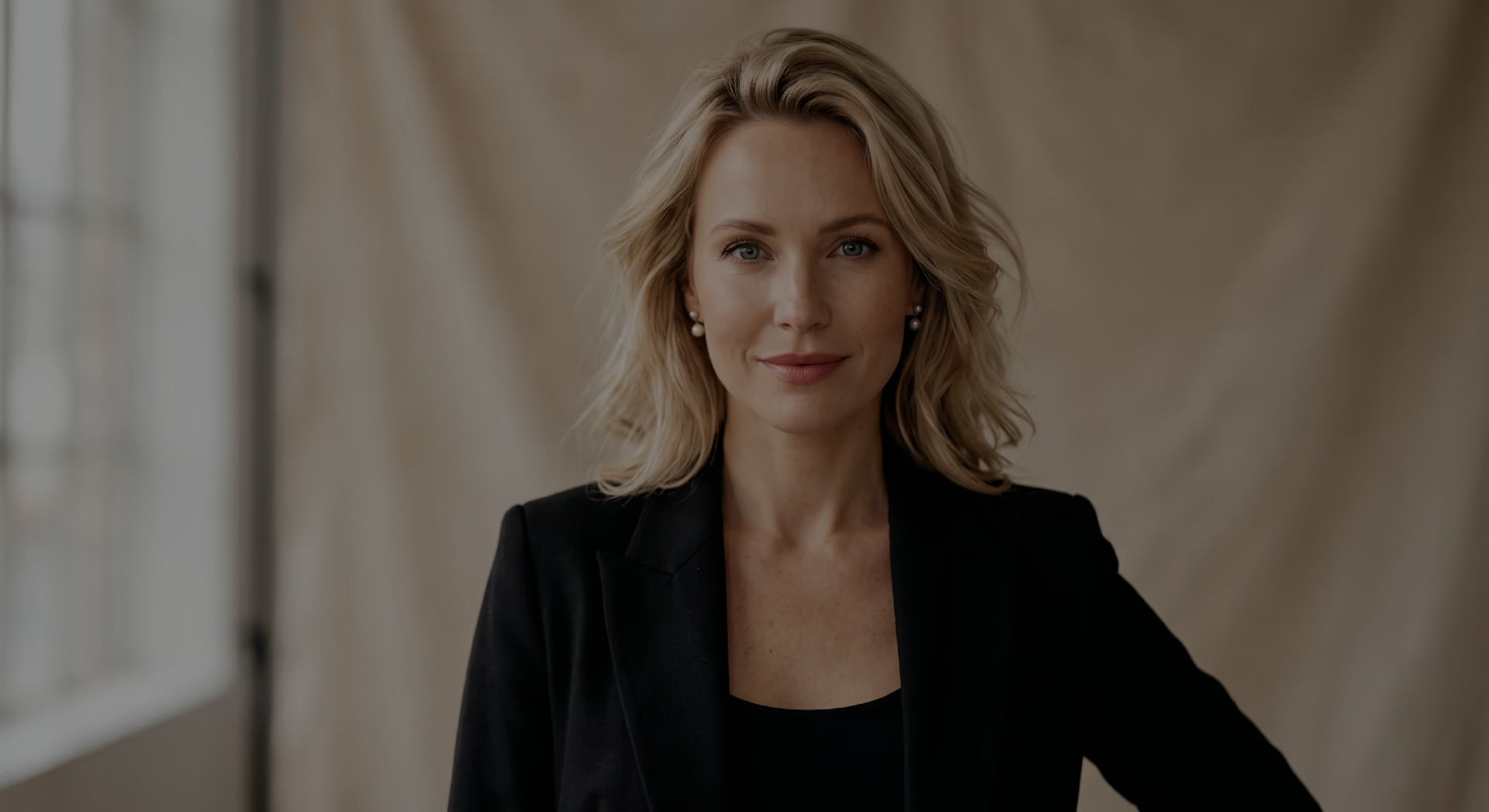} &
  \includegraphics[width=0.22\linewidth]{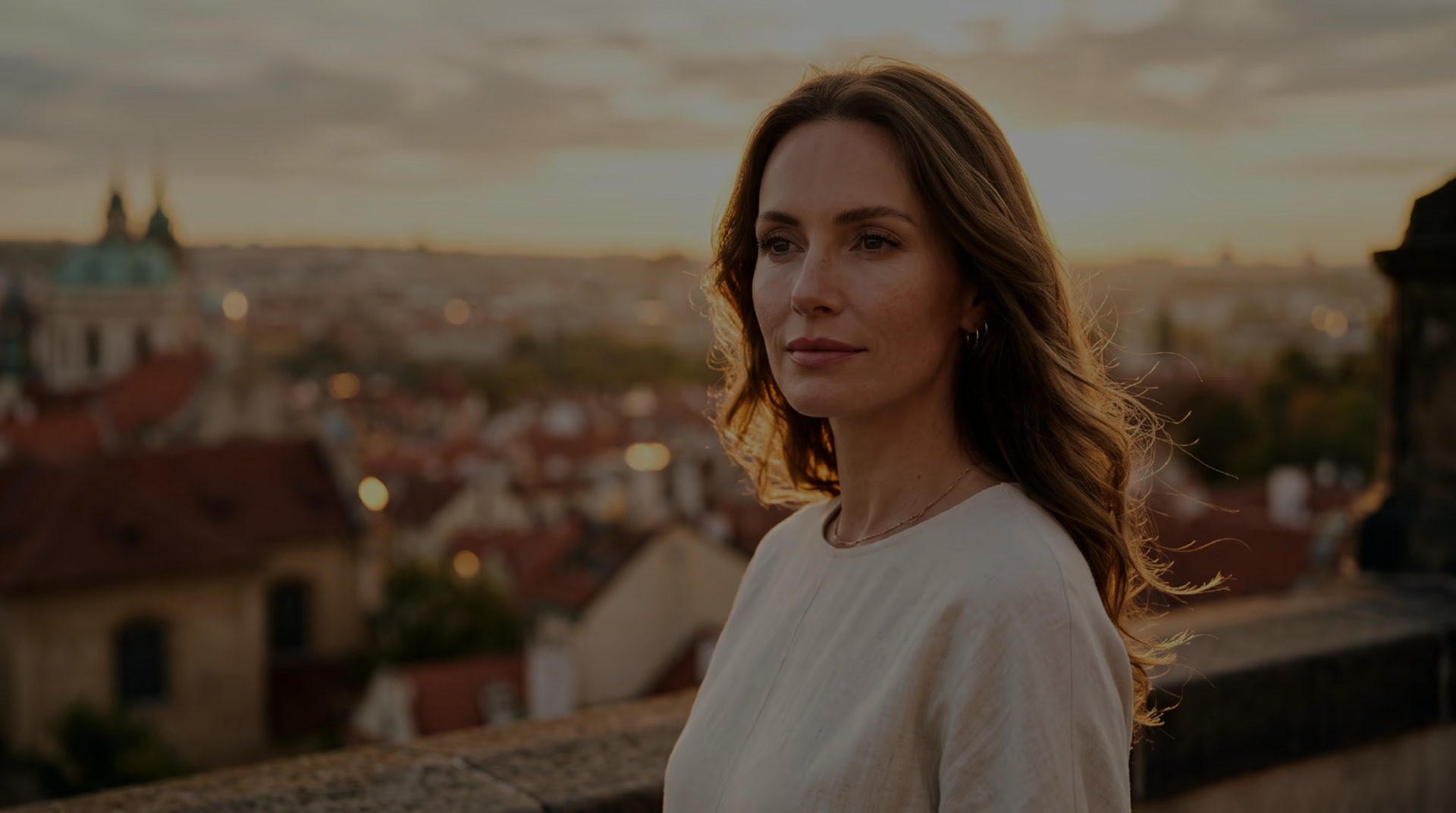} &
  \includegraphics[width=0.22\linewidth]{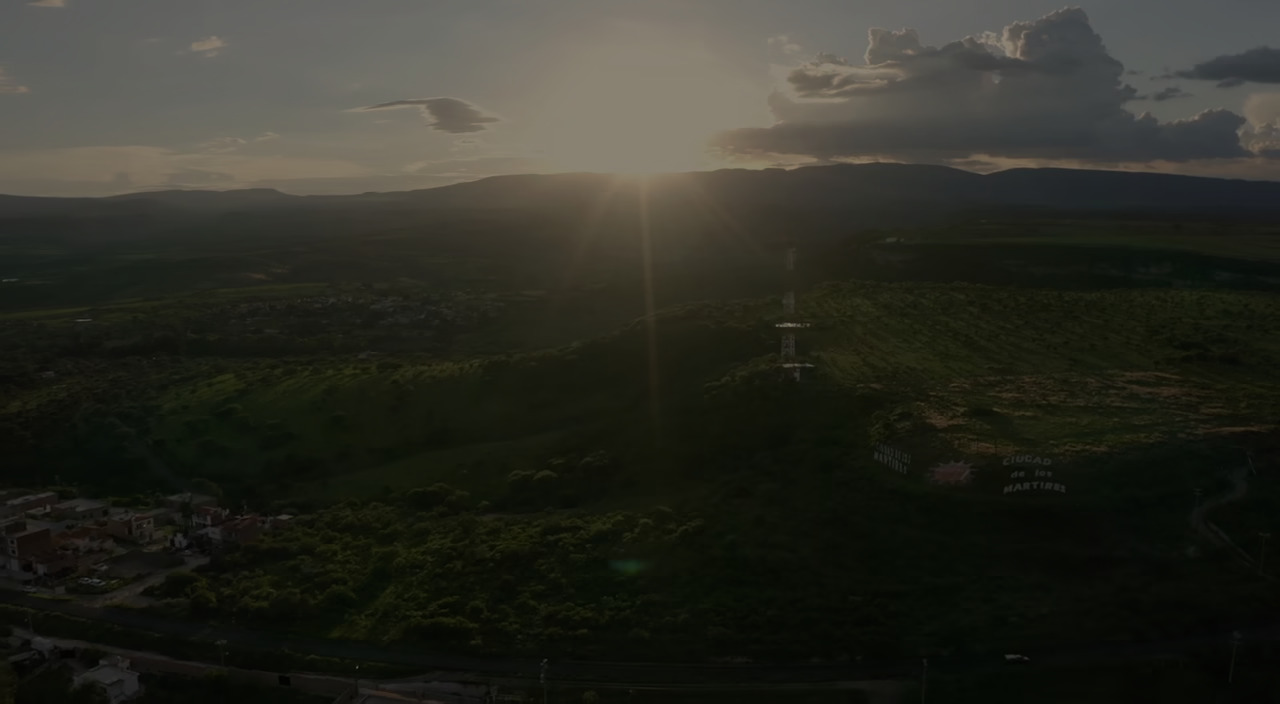} &
  \includegraphics[width=0.22\linewidth]{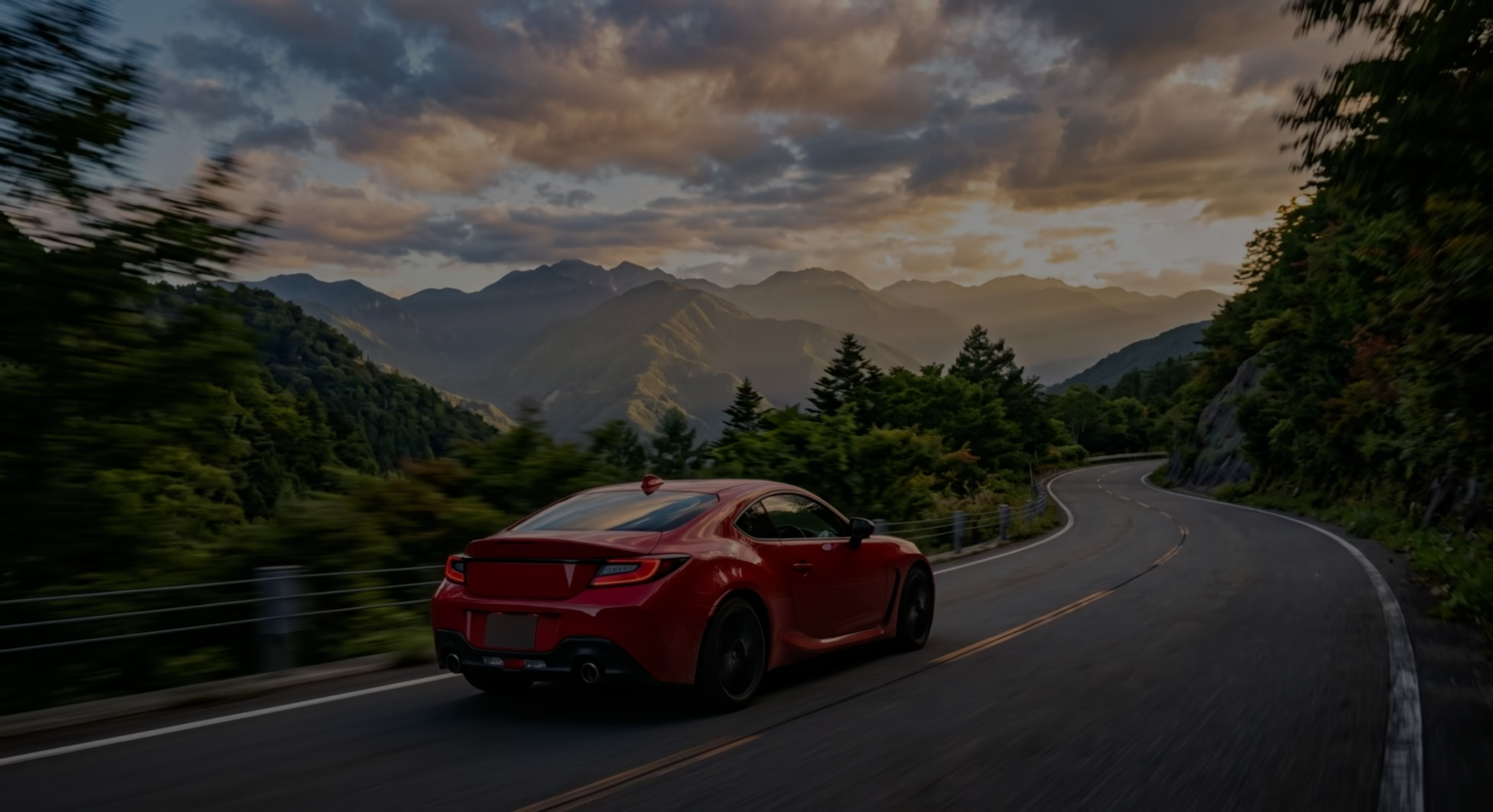} \\
\raisebox{2pt}{\rotatebox{90}{VAE decoder}} &
  \includegraphics[width=0.22\linewidth]{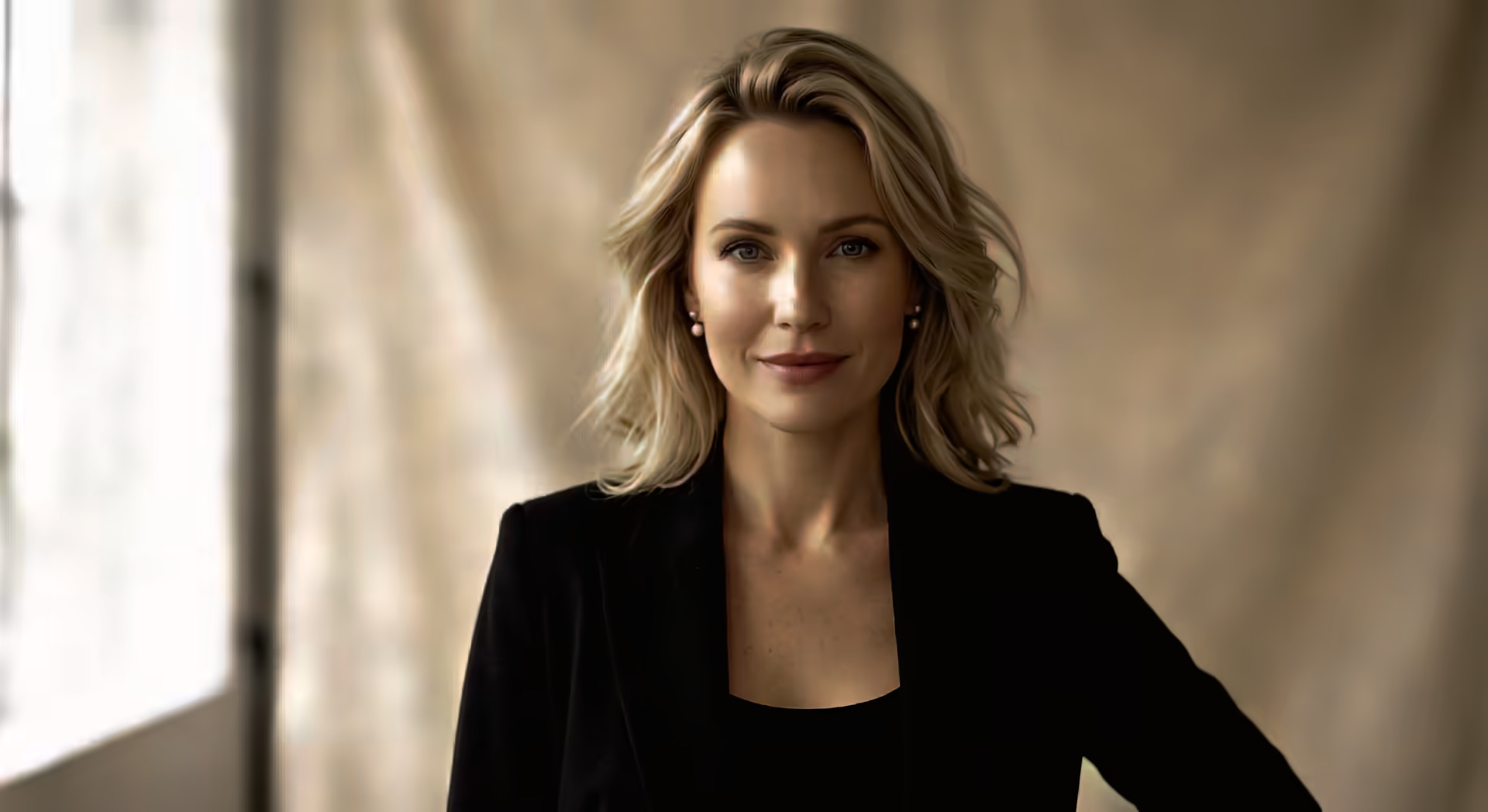} &
  \includegraphics[width=0.22\linewidth]{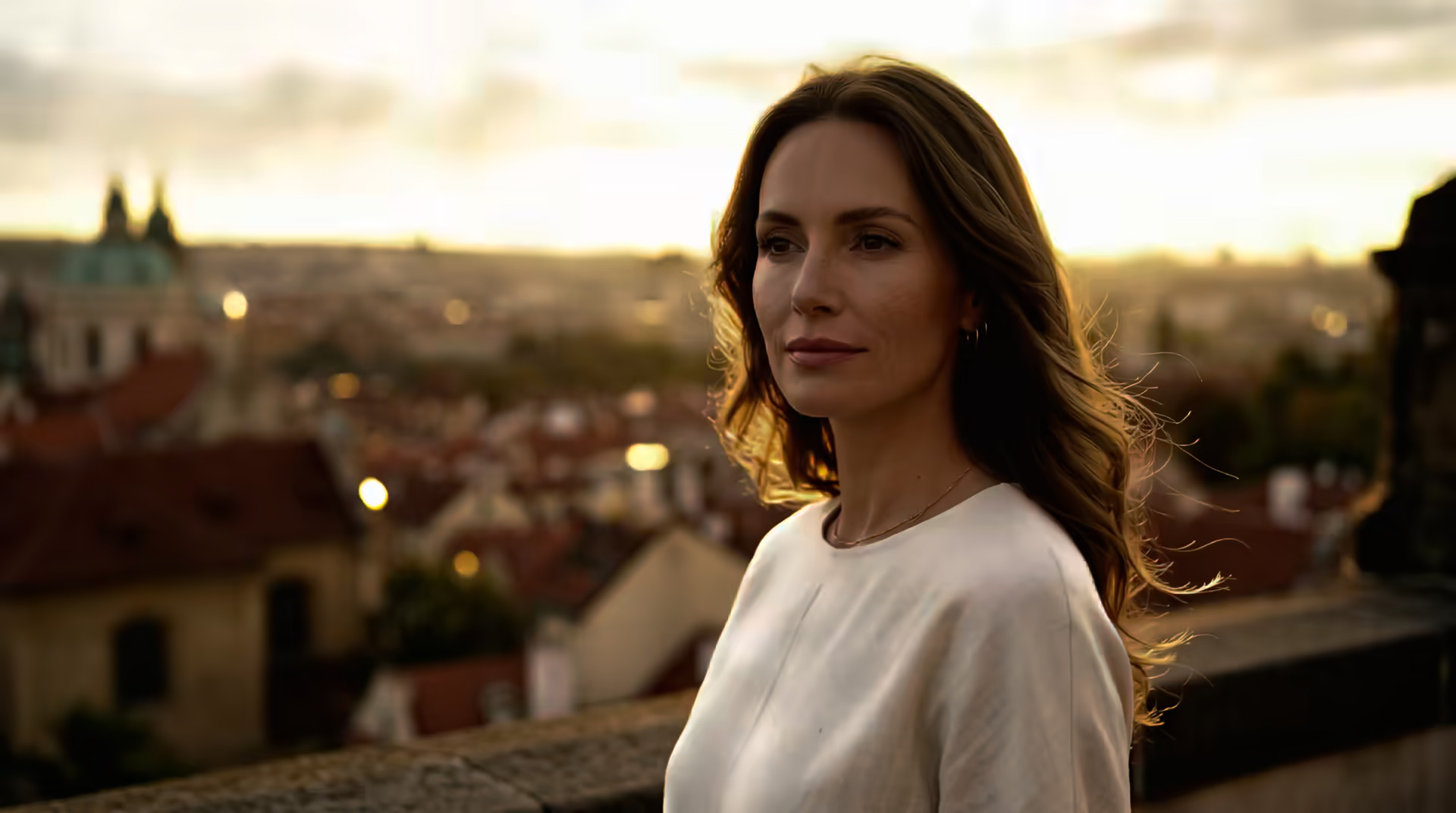} &
  \includegraphics[width=0.22\linewidth]{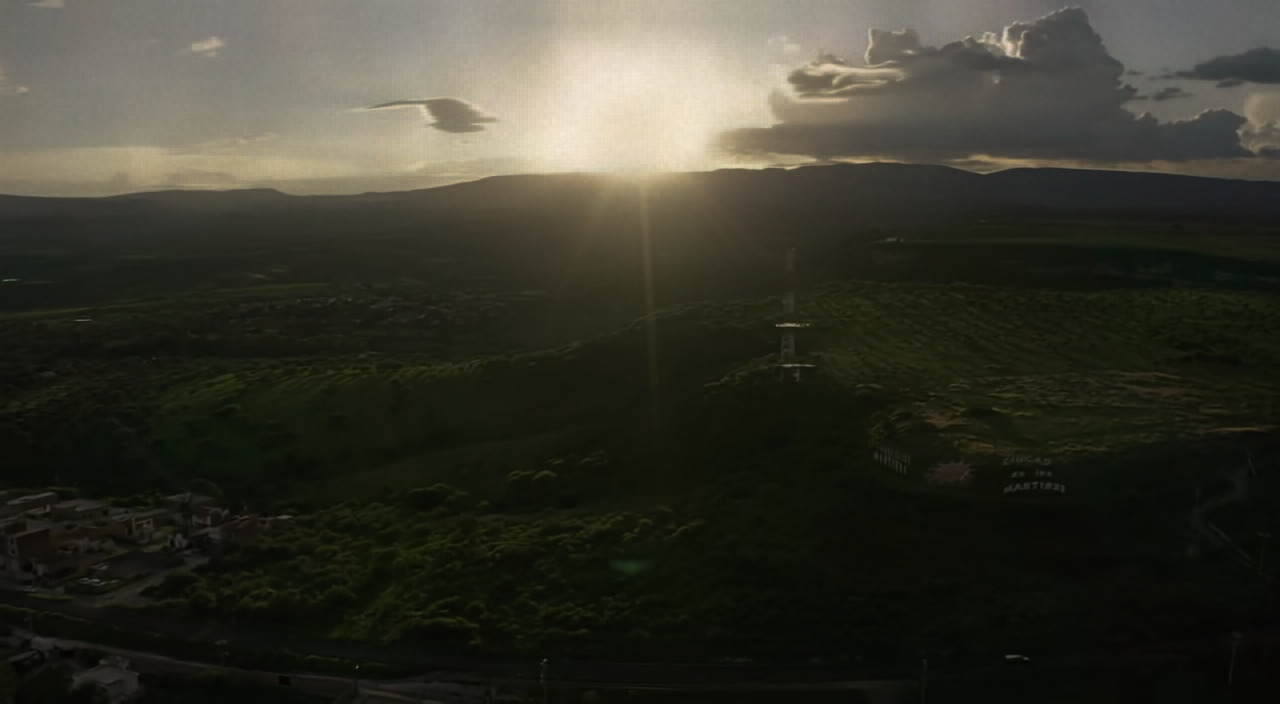} &
  \includegraphics[width=0.22\linewidth]{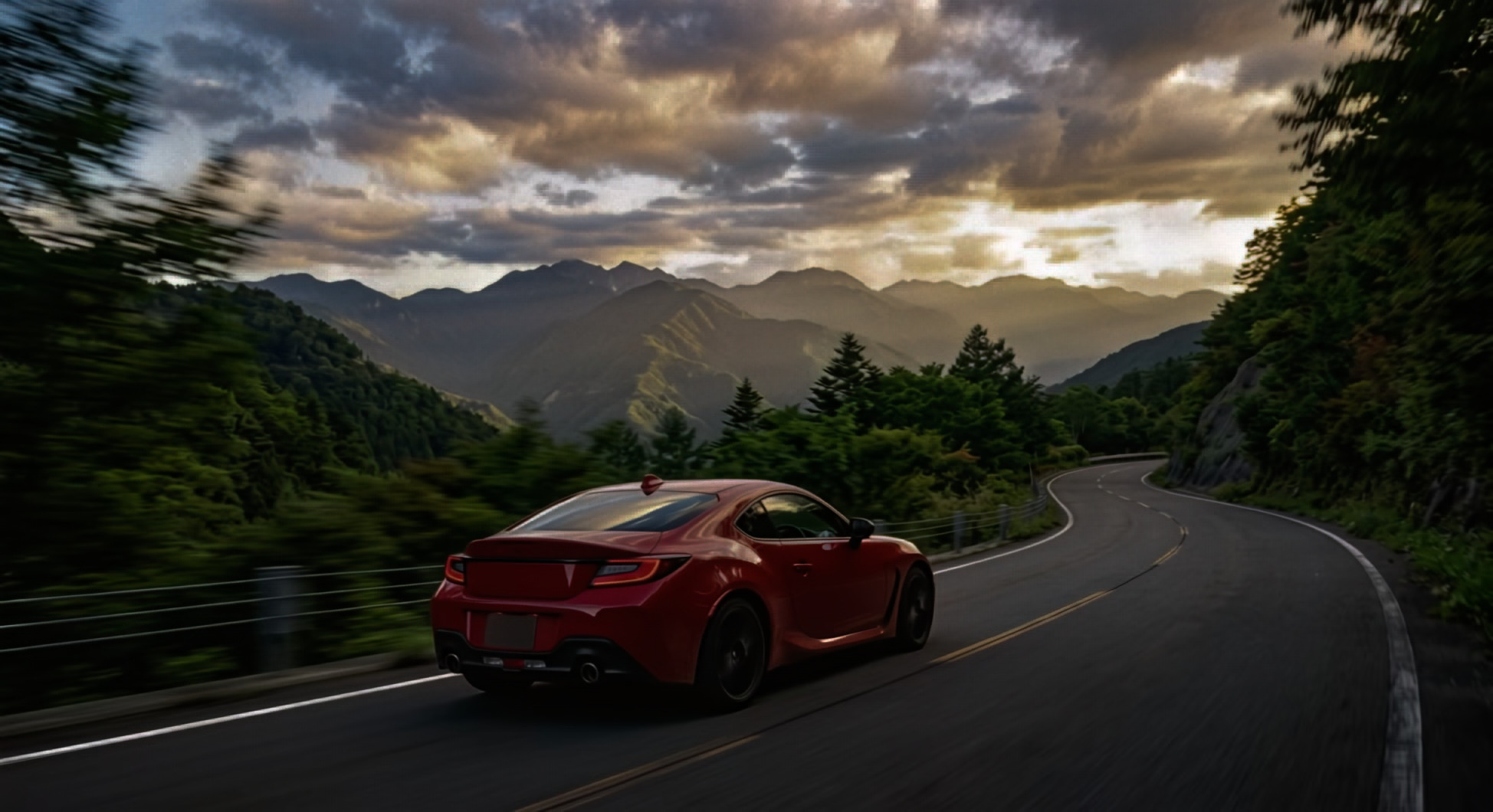} \\
\raisebox{10pt}{\rotatebox{90}{\method{}}} &
  \includegraphics[width=0.22\linewidth]{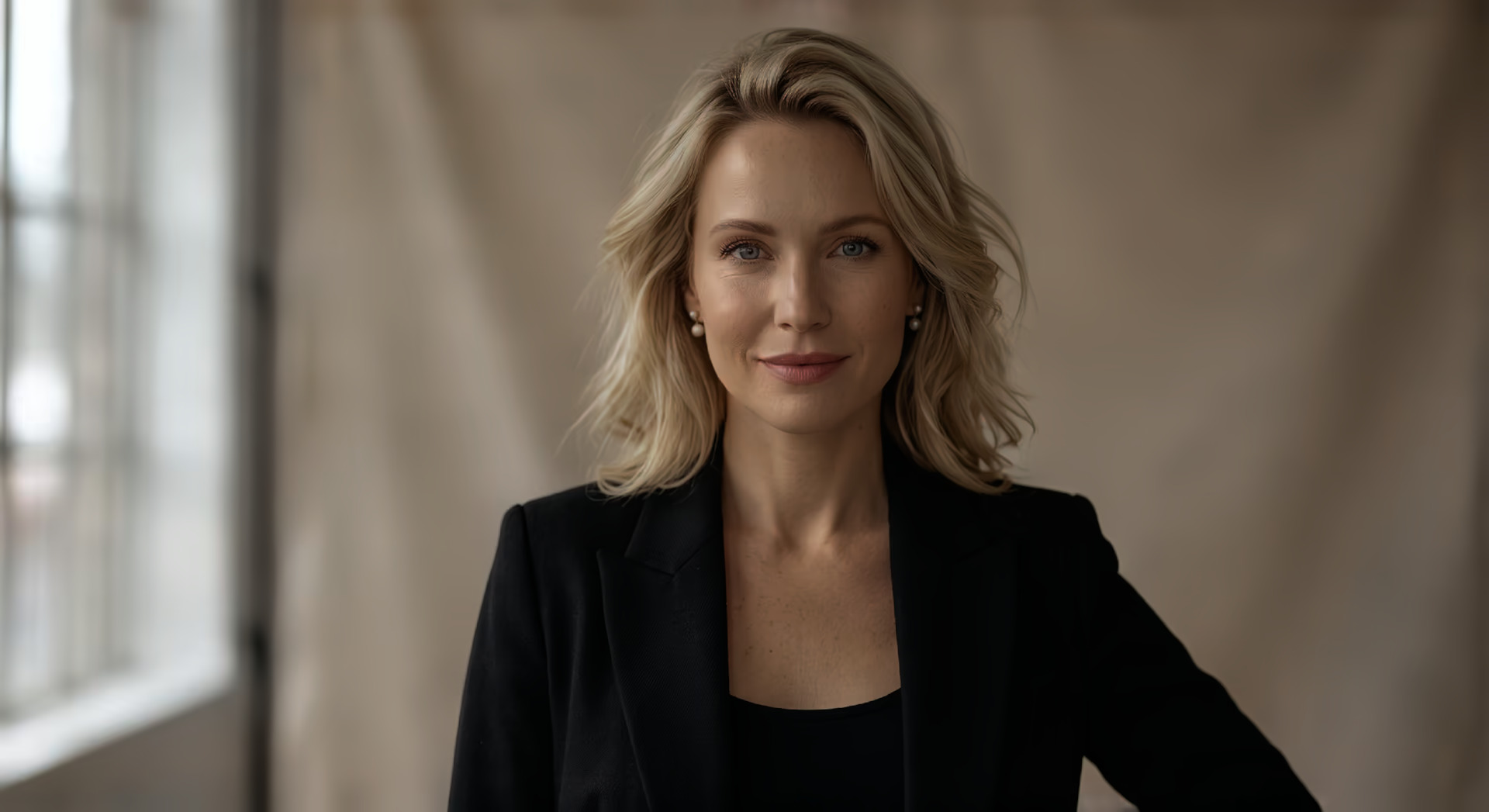} &
  \includegraphics[width=0.22\linewidth]{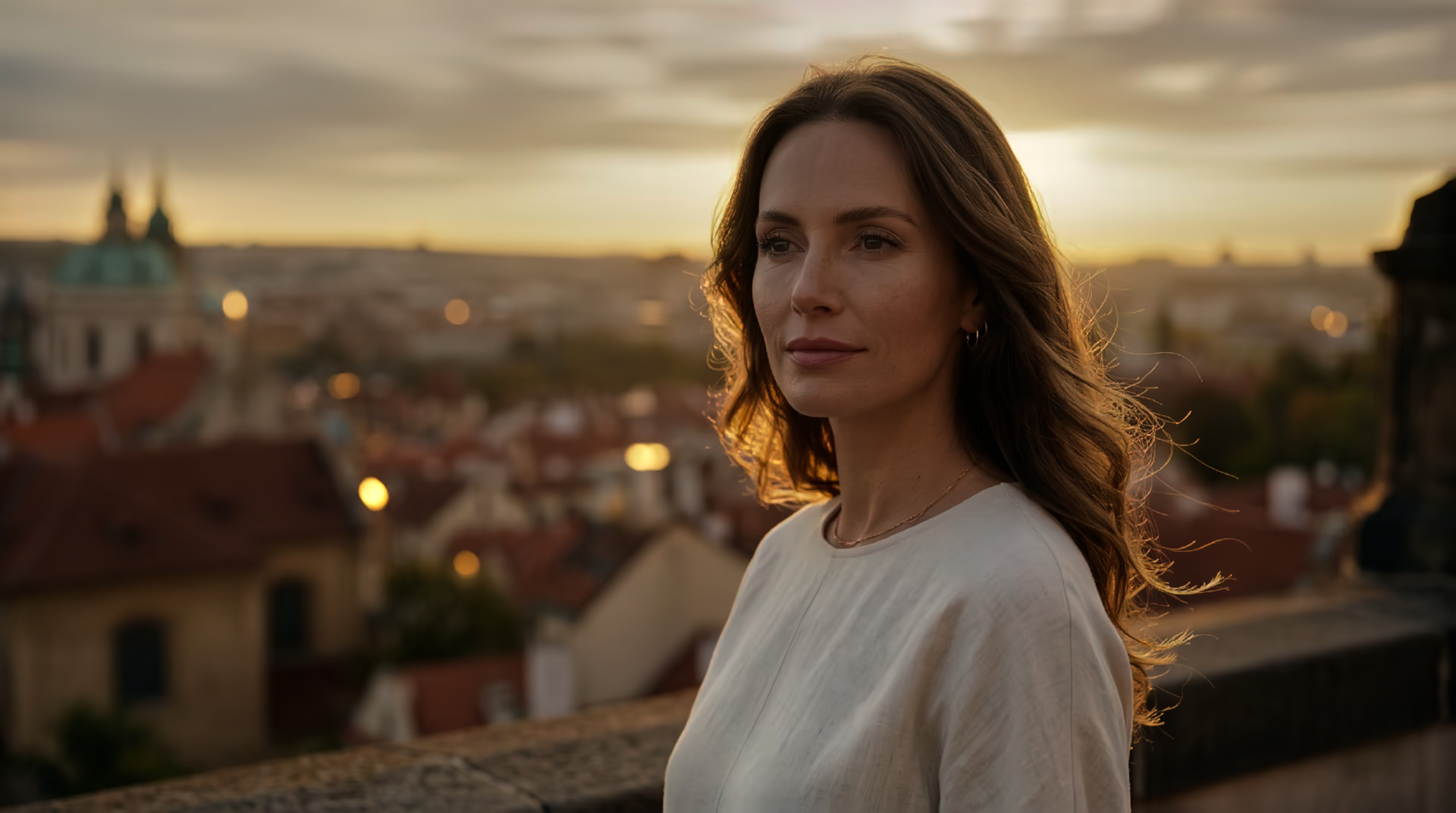} &
  \includegraphics[width=0.22\linewidth]{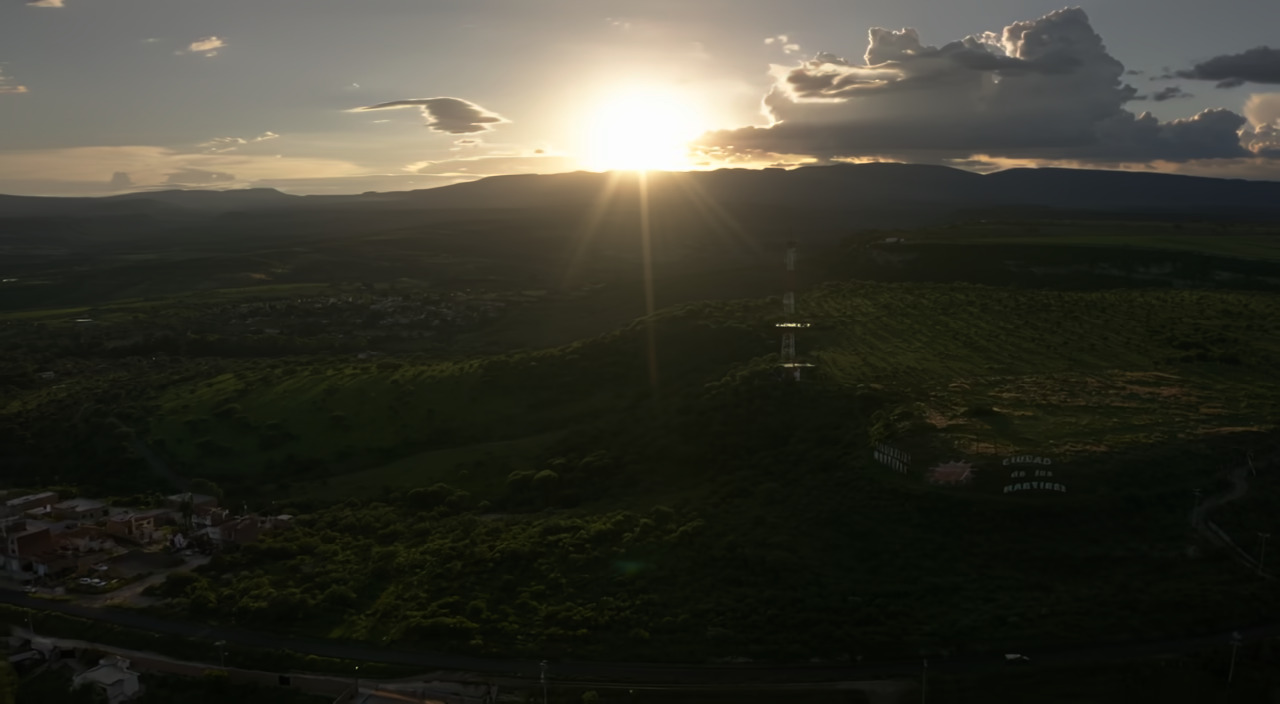} &
  \includegraphics[width=0.22\linewidth]{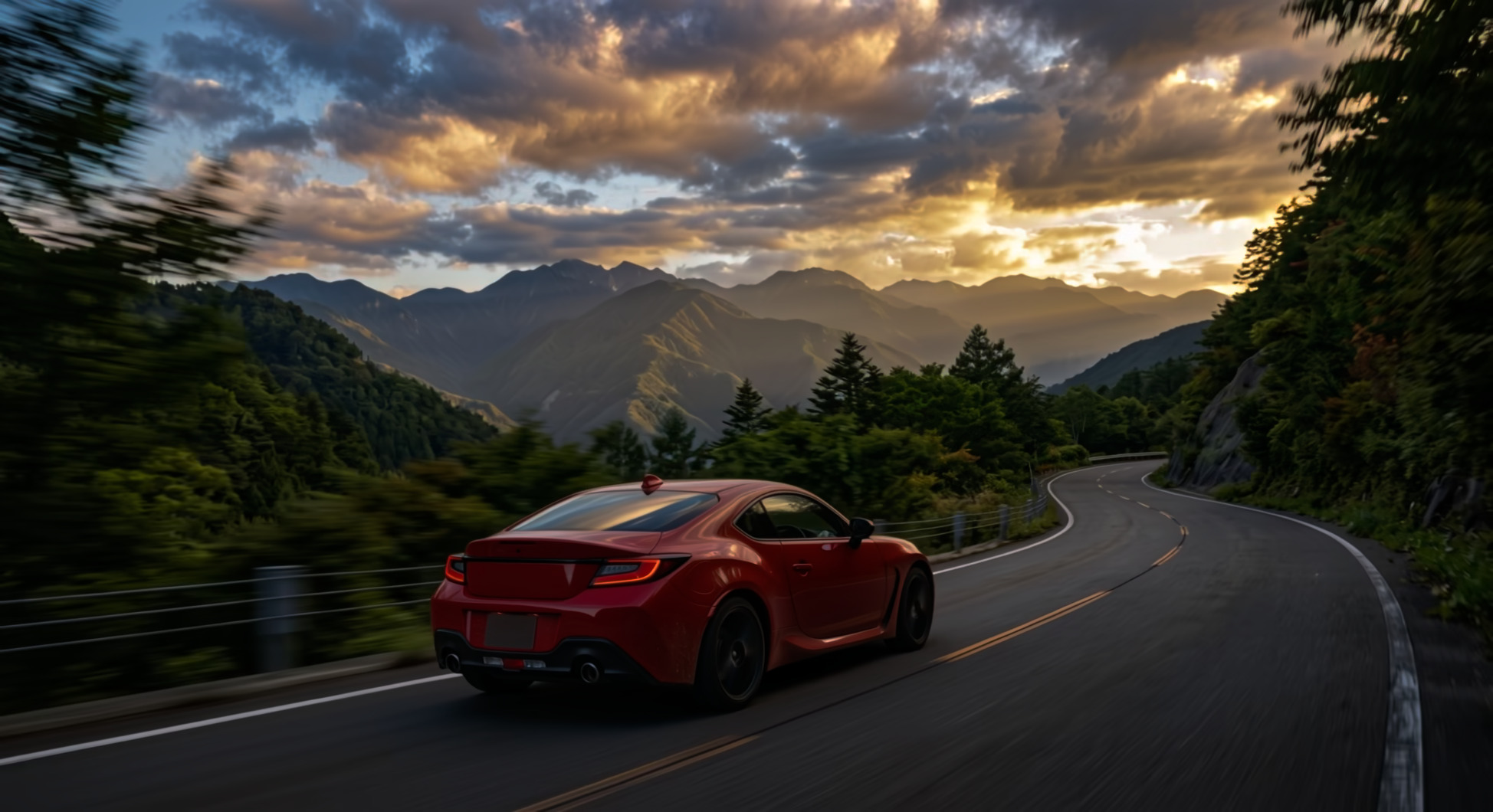} \\
\end{tabular}
\caption{Decoder-only ablation vs.\ the main LoRA on four plates
(Rec.709 preview of the HLG files, EV~$0$).
The VAE decoder can lift range in the window, sky, and sun; but it does not put structure
there. Faces, cloth, and foliage smooth, and those highlights clip.
\method{} is adding the HDR details and structure on top of the SDR input. Best to view these results in the attached project page.}
\label{fig:vae_decoder}
\end{figure*}

\begin{table}[t]
\centering
\caption{Quantitative comparison between the fine-tuned VAE decoder baseline and our main dual-canvas LoRA across ARRI ($n{=}50$) and in-the-wild ($n{=}66$) benchmarks. No-reference perceptual metrics ($\text{MUSIQ}$ and $\text{CLIP-IQA+}$) are evaluated under $\text{EV}-3$ highlight pulls and $\text{EV}+3$ shadow pushes.}
\label{tab:vae_decoder}
\scriptsize
\setlength{\tabcolsep}{4pt}
\begin{tabular}{@{}llcc@{}}
\toprule
Split & Metric & VAE Decoder & \textbf{\method{} (Main)} \\
\midrule
\multirow{4}{*}{ARRI ($n{=}50$)}
& PU21-PSNR $\uparrow$ & 29.43 & \textbf{29.79} \\
& ColorVideoVDP $\uparrow$ & 7.60 & \textbf{7.92} \\
& F2F $\uparrow$ & 46.35 & \textbf{54.28} \\
& t-LPIPS $\downarrow$ & 0.0082 & \textbf{0.0059} \\
\midrule
\multirow{5}{*}{In-the-Wild ($n{=}66$)}
& F2F $\uparrow$ & 43.33 & \textbf{45.15} \\
& t-LPIPS $\downarrow$ & 0.062 & \textbf{0.059} \\
& $\text{MUSIQ}_{-3} \uparrow$ & 43.2 & \textbf{46.3} \\
& $\text{CLIP-IQA+}_{-3} \uparrow$ & 0.407 & \textbf{0.434} \\
& $\text{CLIP-IQA+}_{+3} \uparrow$ & 0.380 & \textbf{0.417} \\
\bottomrule
\end{tabular}
\end{table}

Fine-tuning only the VAE decoder for SDR-to-HDR translation fails to deliver high-fidelity, temporally stable HDR video. While a spatial VAE decoder can predictably re-map diffuse mid-tones, it lacks the generative capacity of the Video DiT backbone to synthesize unclipped radiance or enforce cross-frame temporal coherence. As detailed in Table~\ref{tab:vae_decoder}, the limitations of this decoder-only baseline are evident across both ground-truth-referenced and in-the-wild evaluations.

Under a 3-stop highlight pull ($\text{EV}{-3}$), the VAE decoder fails to synthesize plausible structure in overexposed regions, producing flat, featureless clipping plateaus. This is directly reflected in in-the-wild no-reference metrics, where $\text{MUSIQ}_{-3}$ drops from $46.3$ (\method{}) to $43.2$ and $\text{CLIP-IQA+}_{-3}$ falls from $0.434$ to $0.407$. Symmetrically, pushing shadow regions up by 3 stops ($\text{EV}{+3}$) causes $\text{CLIP-IQA+}_{+3}$ to drop from $0.417$ to $0.380$, demonstrating the decoder's inability to reconstruct details in crushed darks.
The VAE decoder exhibits significant temporal instability. On the ARRI benchmark, frame-to-frame fidelity (F2F) drops sharply from $54.28$ to $46.35$, while temporal flicker ($t$-LPIPS) worsens from $0.0059$ to $0.0082$. Qualitatively (Figure~\ref{fig:vae_decoder}), whereas \method{} leverages diffusion priors to selectively generate extreme peak radiance dictated by scene semantics, the VAE decoder merely expands pixel values uniformly—causing sky regions to oversmooth, specular highlights to smear into halos, and fine geometric edges to vanish. These results confirm that temporally consistent dynamic-range expansion requires the full generative capacity of the Video DiT backbone.

\textbf{Impact of Camera-Mimicking Degradations:}
To force the model to infer unclipped radiance rather than copy input pixels, we apply contrast clipping, compression, and highlight/shadow blur to the SDR conditioning canvas during training. Table~\ref{tab:ablations} evaluates the impact of these degradations.
Training with no augmentation lowers PU21-PSNR by nearly $1.0$~dB ($28.83$ vs.\ $29.79$~dB) and degrades ColorVideoVDP ($7.45$ vs.\ $7.92$). Without contrast clipping, the SDR reference retains quantized highlight steps; the network learns a shortcut by simply stretching 8-bit steps into linear space, resulting in severe posterization and temporal flickering. Combining contrast clipping, highlight/shadow blur, and joint exposure shifts, forces the Video DiT to rely on its diffusion prior to synthesize continuous, unclipped radiance.

\begin{table}[t]
\centering
\caption{LoRA camera-aug ablations}
\label{tab:ablations}
\scriptsize
\setlength{\tabcolsep}{3.5pt}
\begin{tabular}{@{}lccc@{}}
\toprule
Config & PU21-PSNR $\uparrow$ & ColorVideoVDP $\uparrow$ & F2F $\uparrow$ \\
\midrule
Full augs (main) & \textbf{29.79} & \textbf{7.92} & \textbf{54.28} \\
No augmentation & 28.83 & 7.45 & 49.14 \\
Blur only (no contrast) & 29.11 & 7.39 & 49.95 \\
\bottomrule
\end{tabular}
\end{table}

\section{Conclusion}
\label{sec:conclusion}

We presented \method{},
an SDR-to-HDR method that generates temporally coherent wide range HDR video from a single SDR input via a video diffusion model backbone.
We showed that the barrier is not model capacity but how we align HDR content with pretrained models.
A video diffusion transformer trained only on SDR data already possesses rich knowledge about light transport and dynamic range.
Unlocking this knowledge requires solving two problems: mapping HDR into the frozen VAE's learned distribution, 
and forcing the model to learn to reconstruct extreme-luminance regions from its learned priors.
With these in place, a LoRA adapter adding less than 1\% parameters,
trained for only 5{,}000 steps on ${\sim}$300 clips (${\sim}$8 hours, single GPU),
produces temporally coherent float16 HDR video that outperforms the strongest dedicated SDR-to-HDR baseline.
Several extensions are natural.
Scaling the training data to include real-world HDR video captures would strengthen generalization.
Incorporating perceptual HDR metrics such as HDR-VDP-3~\cite{mantiuk2023hdrvdp3} or ColorVideoVDP~\cite{mantiuk2024colorvideovdp} as training objectives could further improve perceptual quality.

Our logarithmic alignment preserves the backbone's native latent space, exciting direction for future work is extending this framework to other pretrained video capabilities, such as HDR video editing, HDR inpainting, and fine-grained visual control.

\section*{Acknowledgments}
We would like to thank Alon Yaar and Alon Alpert from the LTX R\&D group, and Omer Rabinovoch, Michel Angielczyk, and Yoav Lutzky from the LTX Creative group, for their
guidance and support throughout the development of LumiVid.
We are especially grateful to Jean Dominique Vandenberg (JD) for sharing his expertise in color science. 

\bibliographystyle{eg-alpha}
\bibliography{references}

\clearpage
\appendix
\section{Supplementary media}
\label{app:media}

The project page, including the overview video, is the intended way to judge the results: \url{https://hdr-lumivid.github.io/}

\section{Color working space}
\label{app:color}

\begin{figure}[t]
\centering
\includegraphics[width=0.78\linewidth]{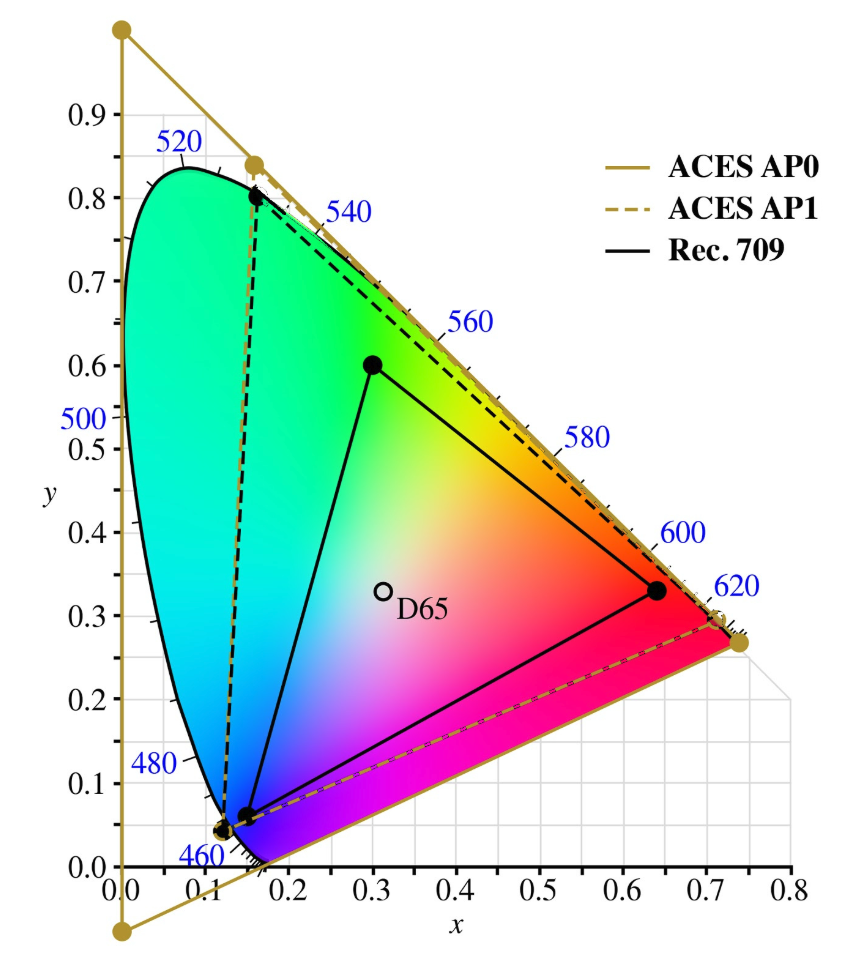}
\caption{CIE $xy$ chromaticity.
Rec.709 is the small SDR triangle (D65): the gamut of almost all web and
broadcast video.
ACEScg is AP1 (dashed gold), a wider working triangle inside the AP0 container
(solid gold).
AP0's blue primary is imaginary (below the spectral locus), so the container
can wrap Rec.709 and AP1 without clipping chromaticity.
The diagram shows chromaticity only; light strength (how far above white) is
not plotted.}
\label{fig:color-range}
\end{figure}

\paragraph{Human color gamut, SDR, and HDR.}
The \emph{human color gamut} is the boundary of human color vision, drawn as
the CIE 1931 chromaticity diagram (Figure~\ref{fig:color-range}).
The triangles inside it, spanned by red, green, and blue primaries, are the
smaller subsets that a given screen, camera, or printer can actually reproduce.

\emph{Gamut} is which chromaticities exist at all.
Most SDR video is stored as Rec.709: a small triangle inside the full locus.
Saturated LEDs, neon, and many camera-native colors lie outside it and are
clipped onto the Rec.709 sides.
An HDR working space uses a wider triangle so those chromaticities remain
distinct instead of collapsing to the Rec.709 primaries.

\emph{Quantization} is whether those values are a continuum or a palette.
Consumer SDR is 8-bit integer per channel after a gamma curve: 256 discrete
codes.
Nearby radiances become the same code; an exposure pull or a grade shows
banding.
HDR is stored as scene-linear floating point (or 10--12-bit PQ on a display
curve).

\emph{Light strength} is how bright the sample is relative to a white surface
in the scene.
SDR is display-referred and clipped to a peak, historically about $100$~nits.
The sun, a headlamp, and a white wall can share the same code.
HDR is scene-referred: we take $x{=}1$ as diffuse white and let highlights be
many times that.

Most videos on the web are Rec.709 since Rec.709 was specified for CRT-era
television: D65 white, those three primaries, and a peak the display could
actually make.
Cameras, broadcast, discs, and later phones and web codecs (H.264/H.265 SDR,
YouTube) copied that contract because the device at the end of the chain was a
Rec.709 display with an 8-bit pipeline.
HDR panels exist now, but not at a scale close to Rec.709 video.
The pretrained video models we adapt were trained on that Rec.709 corpus, so
their VAE expects Rec.709-like statistics.

\paragraph{Rec.709 vs.\ ACEScg.}
Rec.709 is a display/camera encoding: D65, the small triangle in
Figure~\ref{fig:color-range}.
ACEScg is a scene-linear \emph{working} space from the Academy Color Encoding
System~\cite{aces2014}: AP1 primaries -wider than Rec.709, inside AP0. (Rec.709 primaries lie inside AP1.), ACES
white near D60, and \emph{linear} RGB with $x{=}1$ equal to diffuse white and no
upper bound.
AP0 is the ACES2065-1 \emph{container} gamut; its blue primary is imaginary
(below the spectral locus), which is why the container can wrap Rec.709, AP1,
and several camera encodings without clipping chromaticity.
AP1 is the working gamut we actually compute in.
ACEScg is not a display format: it is unbounded linear light
on a wide triangle.

At inference we decode the SDR file with the sRGB EOTF and convert Rec.709
linear to ACEScg and then to ACEScct.
The reverse path writes ACEScg EXR; metrics convert ACEScg to Rec.709 linear so
every method is scored in the same space as the ARRI ground truth
(Appendix~\ref{app:protocol}).

\paragraph{ACEScct.}
Linear ACEScg is a poor input to a frozen SDR VAE: the distribution is
heavy-tailed and part of the pixels sits far above the $[0,1]$ codes the VAE saw
in training.
ACEScct is a per-channel log of AP1.

Let $x$ be an ACEScg linear RGB channel.
The forward map is
\begin{equation}
  T(x) =
  \begin{cases}
    10.540237742 x + 0.072905534 & x \le 0.0078125, \\
    \dfrac{\log_2(x) + 9.72}{17.52} & x > 0.0078125.
  \end{cases}
\end{equation}
The inverse is
\begin{equation}
  T^{-1}(y) =
  \begin{cases}
    \dfrac{y - 0.072905534}{10.540237742} & y \le 0.15525114155, \\
    2^{y \cdot 17.52 - 9.72} & y > 0.15525114155.
  \end{cases}
\end{equation}
$T(1){\approx}0.555$ (diffuse white) and $T^{-1}(1){\approx}223$.
The linear toe below $x{=}2^{-7}$ avoids $\log 0$ in deep shadows.

Before the VAE we clamp ACEScct.
On the SDR canvas the upper clamp is $0.55$, just below $T(1)$, so the condition
cannot present highlights above white ($T^{-1}(0.55){\approx}0.94$).
On the HDR canvas the upper clamp is $1.0$ (${\approx}223\times$ white).

The $0.55$ vs.\ $1.0$ split stops the LoRA from copying values that would
otherwise sit in both ranges.
Camera-mimicking degradations are applied on the SDR canvas after this clamp. ACEScct is a production log color space with a documented inverse.

\section{Training and Inference Implementation Details}
\label{sec:appendix_c}

\paragraph{Dual-Canvas LoRA Architecture}
Following AVControl~\cite{ben2026avcontrol}, we construct an in-context dual-canvas framework by concatenating the SDR reference canvas with the HDR target canvas in the spatial-temporal sequence dimension. The SDR condition canvas is encoded once by the frozen VAE and injected as reference tokens, while the HDR canvas contains the noisy target latents. Noisy latents are not concatenated along the channel dimension. For text-to-HDR generation, the SDR condition canvas is simply dropped (Appendix~\ref{app:t2hdr}). Training the LoRA in this dual-canvas fashion is exceptionally data-efficient because the backbone does not need to learn generative video synthesis from scratch; it merely learns a targeted dynamic-range translation from the ACEScct SDR sub-space to the full ACEScct HDR range.

\paragraph{Optimization Protocol}
We train a rank-128 LoRA adapter on the frozen LTX-2.5 Video DiT backbone using flow matching for 5,000 steps on $\sim$300 paired 49-frame clips. The VAE encoder and decoder remain completely frozen throughout training. Optimization takes approximately eight hours on a single NVIDIA A100 GPU. The trainable LoRA parameters represent less than 1\% of the total backbone parameters. At inference time, we use 8 denoising steps with standard flow matching sampling.

\paragraph{Inference Color Pipeline}
During inference, the input SDR video file is decoded using the standard sRGB Electro-Optical Transfer Function (EOTF) to obtain scene-linear values, converted from Rec.709 color primaries to ACEScg scene-linear space, mapped into the log ACEScct SDR range $[0, 0.55]$ via $\mathcal{T}(x)$, and encoded into latent space by the frozen VAE. The Video DiT with rank-128 LoRA adapters performs iterative denoising directly in ACEScct latent space. Following VAE decoding, output latents in ACEScct $[0, 1.0]$ are inverted back to scene-linear ACEScg via $\mathcal{T}^{-1}(y)$ and exported as 16-bit half-float EXR files. For metric evaluation, ACEScg EXRs are transformed into Rec.709 scene-linear space to ensure photometric alignment with ground-truth ARRI footage.

\section{Training data}
\label{app:data}

Paired clips come from Poly Haven camera orbits~\cite{polyhaven}.
Both sides start from sRGB Rec709 converted to ACEScg scene-linear frames, mapped through ACEScct $T$.
The HDR canvas is clamped at $1.0$.
The SDR canvas is clamped at $0.55$, then quantized, and passed through the
camera-mimicking degradations of 
We use about 300 pairs.
No ARRI footage is in the training set.
The ARRI test set is 50 clips$\times$49 frames at $1280X720$; in-the-wild is 66 YouTube
SDR clips with no HDR ground truth.

\section{Camera-Mimicking Degradation Implementation Details}
\label{app:degradations}

This section details the hyperparameter distributions, ACEScct masking formulas, and application probabilities for the SDR condition degradations described in Section~\ref{sec:degradations}.

\subsection{Luminance-Targeted Spatial Blur in ACEScct Space}
Spatial blur degradation is applied with probability $p = 1$.
All degradations operate directly in ACEScct logarithmic space. For the SDR reference canvas, ACEScct values are clamped to $y \in [0, 0.55]$, where $y_{\text{white}} \approx 0.55$ corresponds to diffuse white ($x = 1.0$). 

The highlight degradation $B_{\text{hi}}(I)$ applies a Gaussian blur with standard deviation $\sigma \sim \mathcal{U}(5, 12)$ and kernel size $k = 2\lfloor 2\sigma \rfloor + 1$. The shadow degradation $B_{\text{sh}}(I)$ downsamples the log image by $2\times$ using area interpolation followed by cubic upsampling. The degraded ACEScct image is blended back.

\subsection{Anchored Contrast Scaling}
Contrast scaling is applied with probability $p = 0.30$. The scaling factor $f$ is uniformly sampled from $\mathcal{U}(1.0, 1.5)$, anchored around the ACEScct midpoint $0.5$:
\begin{equation}
    I_{\text{contrast}} = \text{clamp}\left(0.5 + (I_{\text{ACEScct}} - 0.5) \cdot f, \, 0, \, 0.55\right).
\end{equation}

\subsection{H.264 Recompression}
All SDR reference sequences undergo H.264 video encoding. To simulate web video compression, $30\%$ of reference clips are encoded with an elevated Constant Rate Factor $\text{CRF} \sim \text{UniformDiscrete}(23, 35)$ (mean $\text{CRF} \approx 29.3$), while the remaining $70\%$ use a high-quality baseline of $\text{CRF} = 12$.

\section{Photometric protocol}
\label{app:protocol}

All methods are scored in Rec.709 scene-linear.
ARRI ground truth is ARRI Wide Gamut converted to Rec.709.
Our EXRs are ACEScg converted to Rec.709.
DiffHDR~\cite{yu2026diffhdr} and Tedla et al.~\cite{tedla2026generating} already emit Rec.709 linear.

\section{Evaluation metricks}

\begin{enumerate}
    \item 

Ground-Truth Referenced Fidelity: 
PU21-PSNR/SSIM~\cite{mantiuk2021pu21} PU21-PSNR/SSIM is PSNR (and SSIM) between ground-truth and generated frames after those frames are encoded with PU21. That encoding makes the PSNR better matched to the eye. In both frames the brightest values are very high; the eye is not very sensitive to differences in that range, and much more sensitive in the darks. Encoding to PU21 therefore gives a larger error to mistakes in the more sensitive parts of the frame.

ColorVideoVDP. JOD~\cite{mantiuk2024colorvideovdp} is a comparison to ground-truth, of how the clips would look on an HDR display rather than in encoded pixels. It models a 1000-nit panel, with 1.0 in the EXR shown as 100 nits (diffuse white), and scores short video windows (9 frames at 24 fps) on a 0–10 JOD scale. As with PU21, small errors in the brightest highlights count less than errors in darker regions; here that is built into the display model instead of into a PSNR.

\item Temporal metrics.
F2F is PSNR between consecutive PU21 frames.
t-LPIPS is mean LPIPS between consecutive PU21 frames.

\item No-reference (in-the-wild).
MUSIQ and CLIP-IQA+ are reported at EV $-3$ because EV~$0$ is saturated on many
clips.
CLIP-IQA+ at EV $+3$ is a shadow check.
There is no HDR ground truth on this split.
\end{enumerate}

\section{Baseline recipes}
\label{app:baselines}

\paragraph{DiffHDR.}
Official 49-frame, $720\times 1280$ recipe on ARRI; 17-frame windows on
in-the-wild.
EXRs are Rec.709 linear.
We used the published video recipe.

\paragraph{HDRTVNet++ and X2HDR}
Run frame-wise on the same SDR frames, then stacked into a clip for video
metrics.
HDRTVNet++ targets HDR10 delivery, not scene-linear EXR.

\paragraph{Tedla et al.}
On in-the-wild we run the official 17-frame, $704\times 1280$ recipe (empty prompt, 50 steps, seed 1).
On ARRI we tile three windows at starts $0/16/32$ (1-frame overlaps overwritten)
to cover 49 frames; in-the-wild is a single 17-frame window.
EXR's are Rec.709 linear.

\section{In-the-wild baseline comparison}
\label{app:car}

Figure~\ref{fig:car_comparison} is an ActionVFX garage plate (no HDR ground truth)
with the same methods as Tables~\ref{tab:arri}--\ref{tab:wild}.
SDR, \method{}, and DiffHDR are baked from scene-linear EXRs
(ACEScg$\to$Rec.709 for \method{}; Rec.709-linear for DiffHDR) via
$sRGB(Y\cdot 2^{\mathrm{EV}})$.
LEDiff is an image checkpoint (not a 49-frame video model).

\section{Text-to-HDR}
\label{app:t2hdr}

\begin{figure*}
    \setlength{\tabcolsep}{1pt}
    \centering
    \small
    \begin{tabular}{cccc}
        \raisebox{15pt}{\rotatebox{90}{LumiVid}} &
            \includegraphics[width=0.32\linewidth]{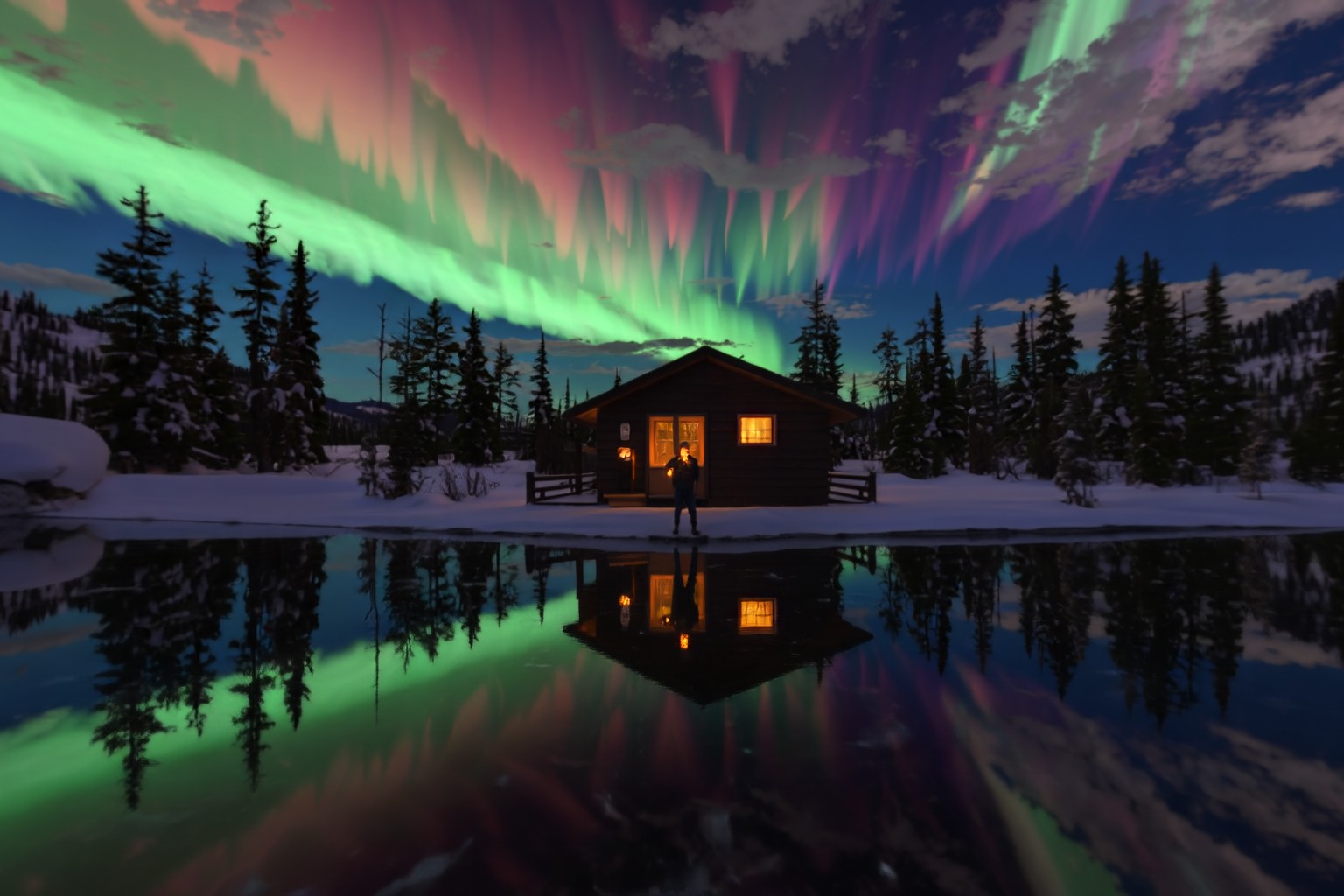} & \includegraphics[width=0.32\linewidth]{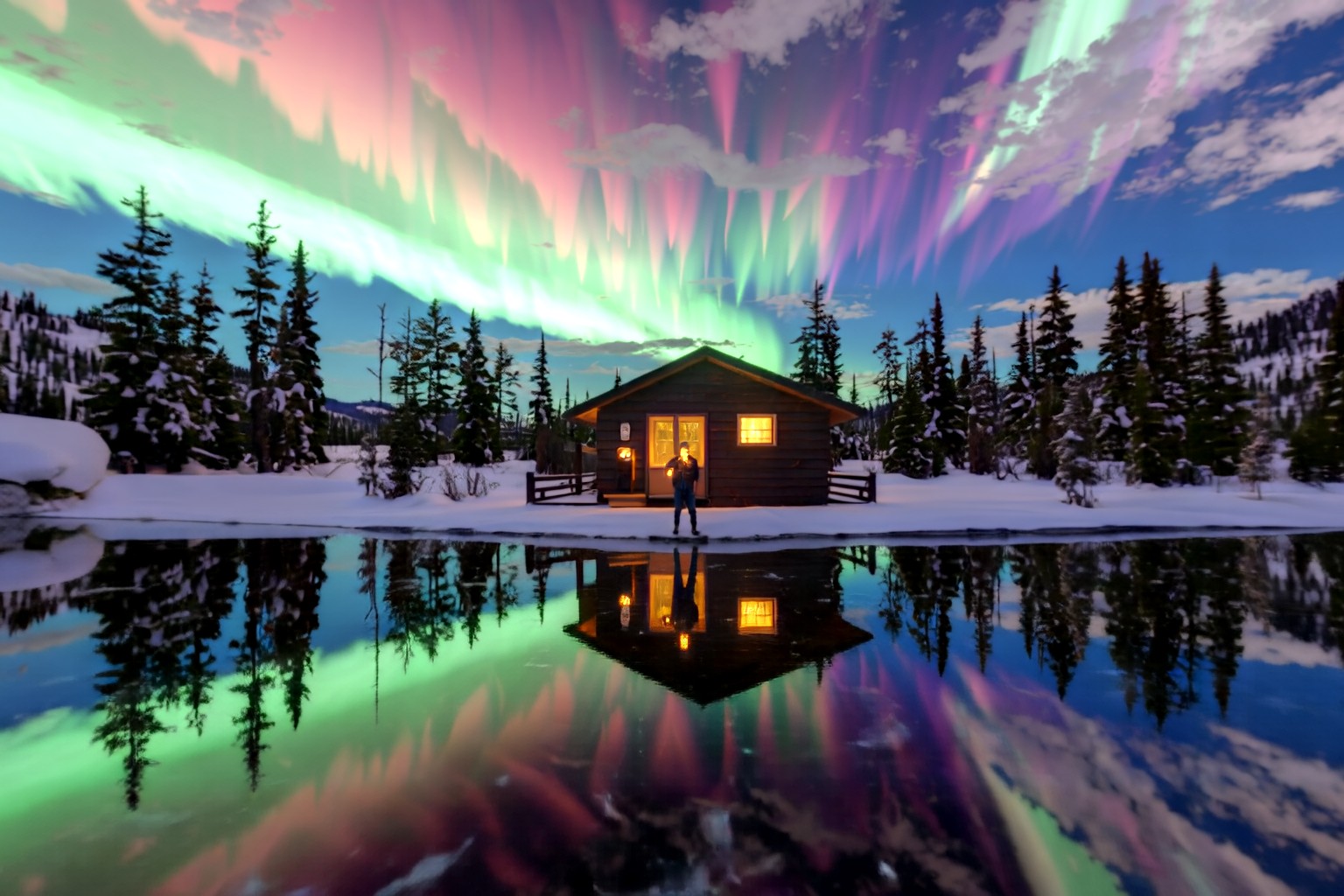} & \includegraphics[width=0.32\linewidth]{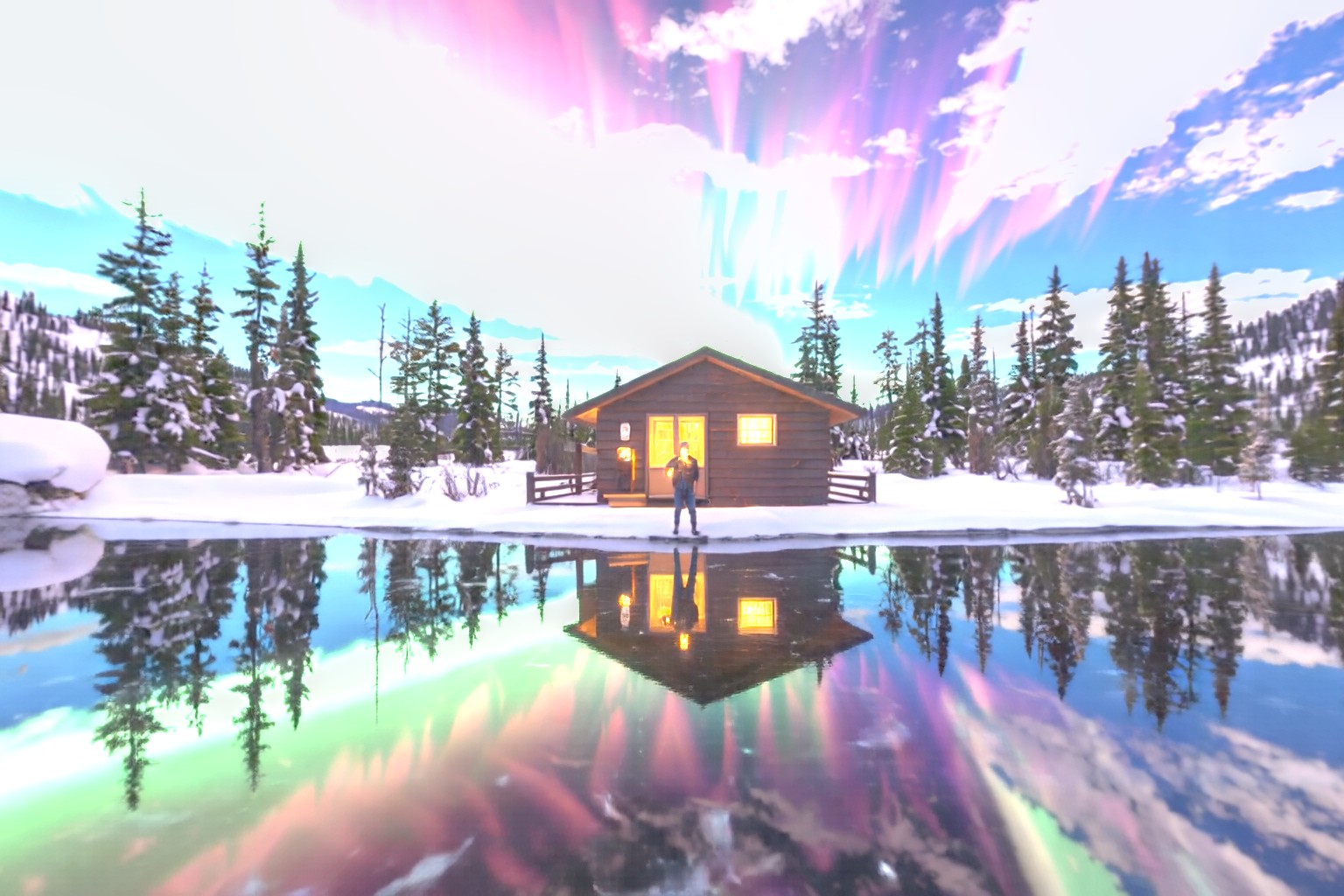} \\
        \raisebox{12pt}{\rotatebox{90}{X2HDR}} &
            \includegraphics[width=0.32\linewidth]{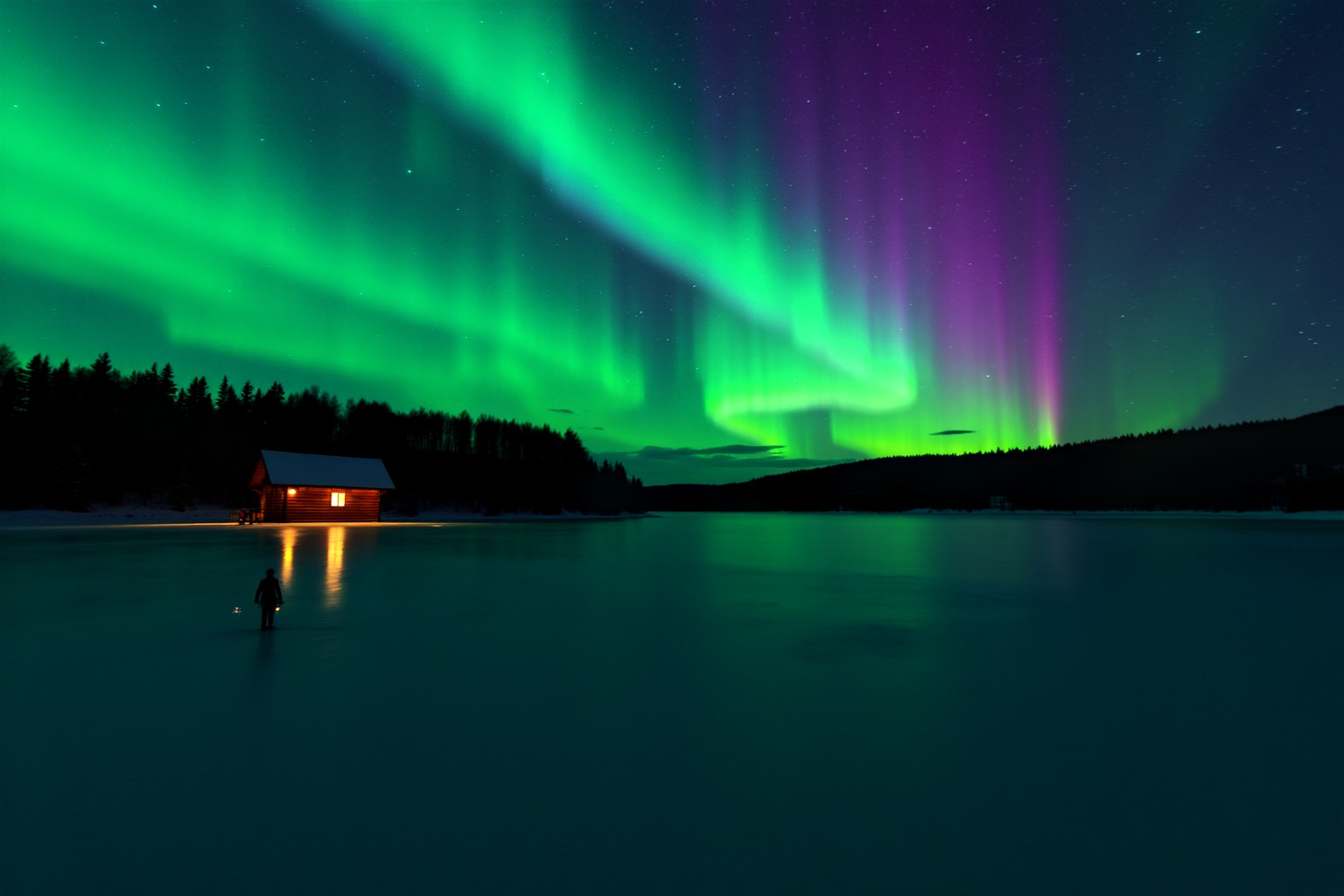} & \includegraphics[width=0.32\linewidth]{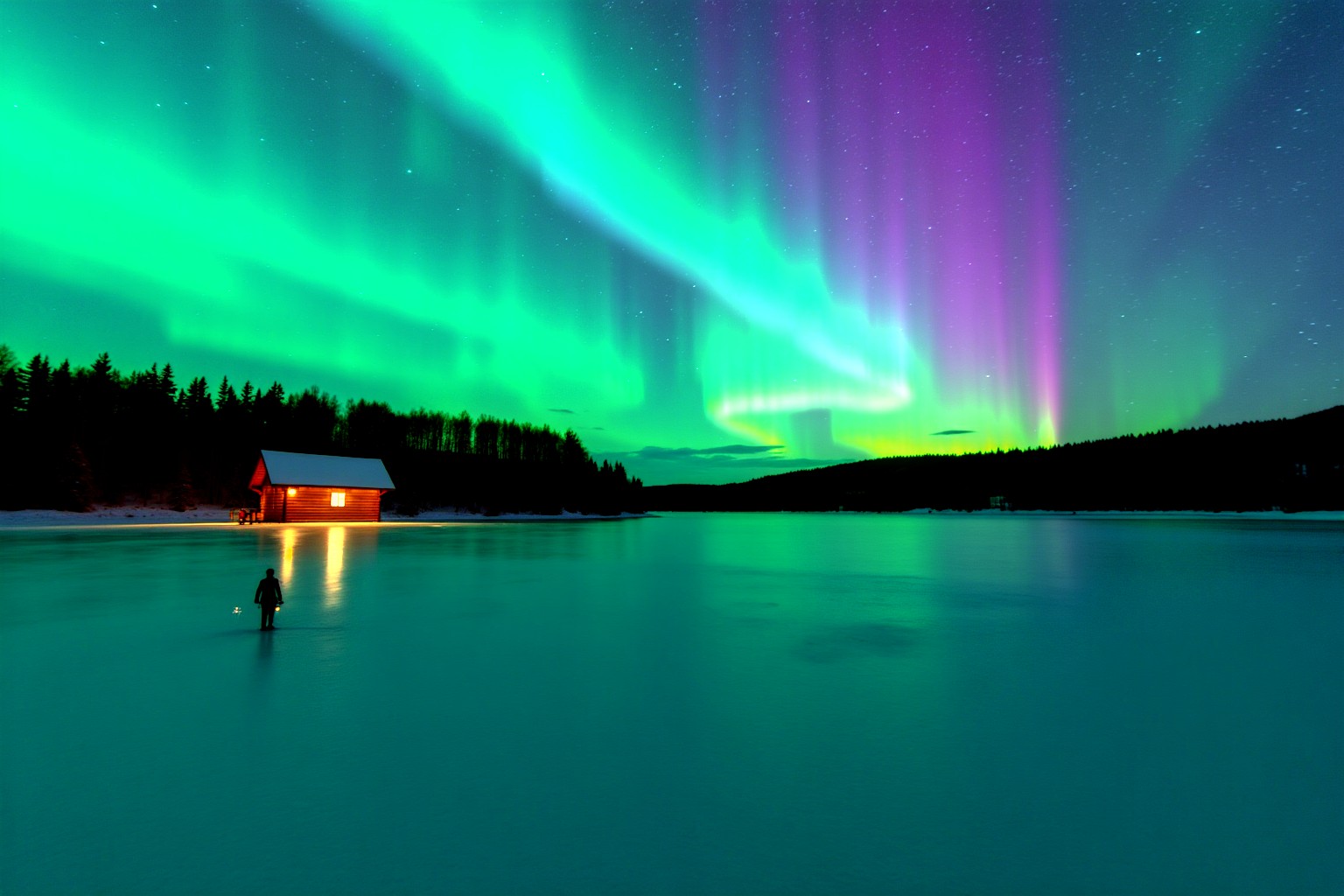} & \includegraphics[width=0.32\linewidth]{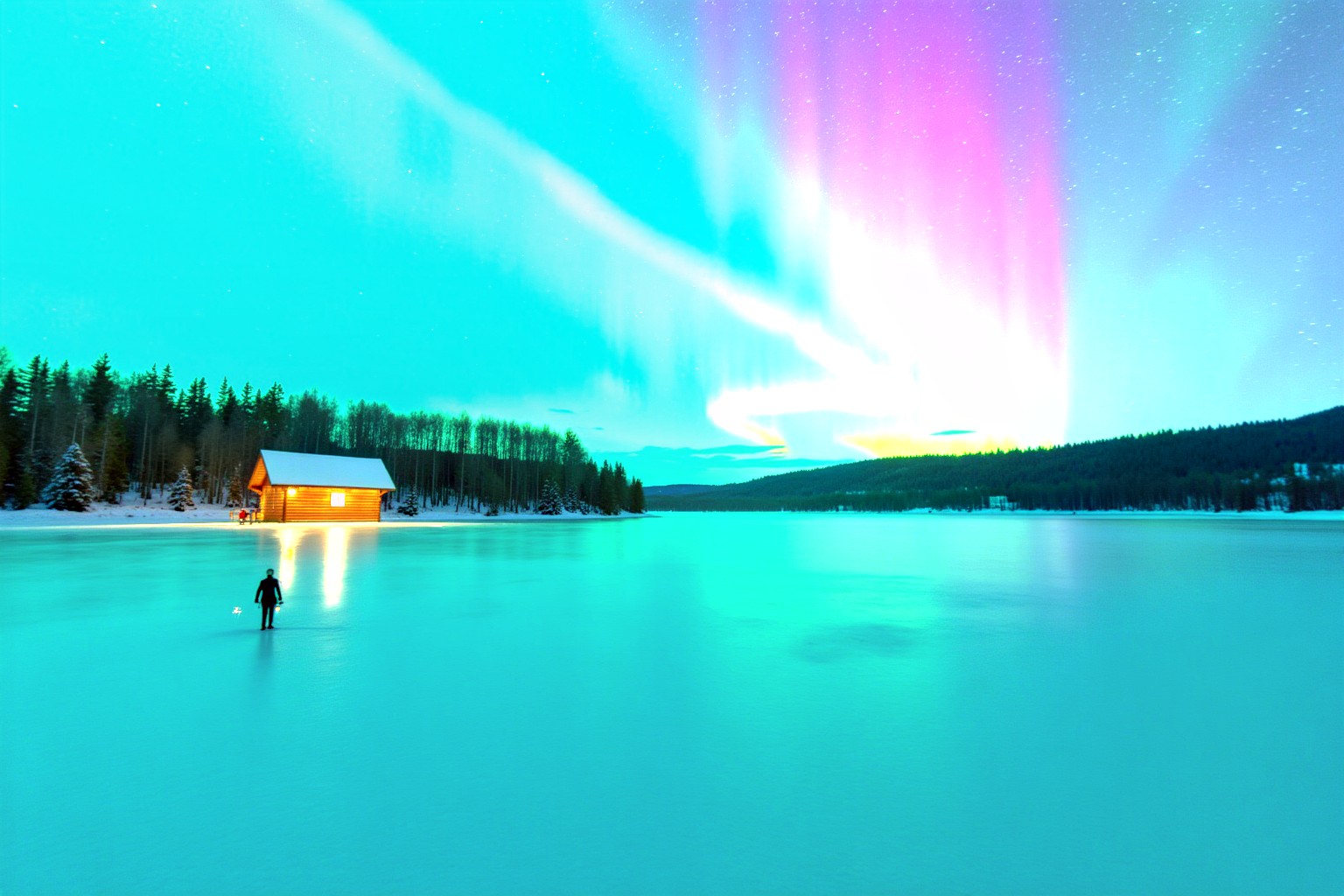} \\
        \raisebox{15pt}{\rotatebox{90}{LumiVid}} &
            \includegraphics[width=0.32\linewidth]{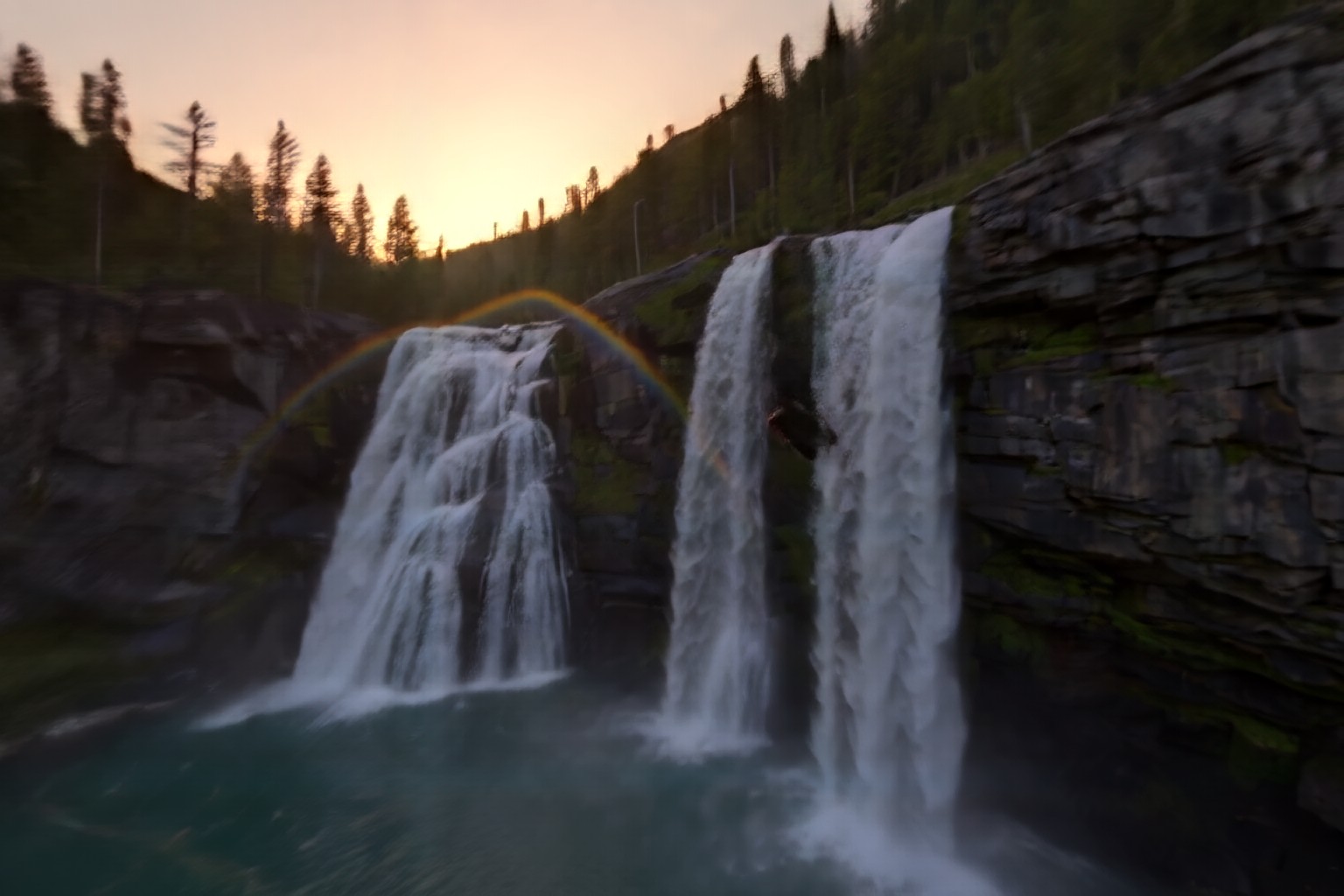} & \includegraphics[width=0.32\linewidth]{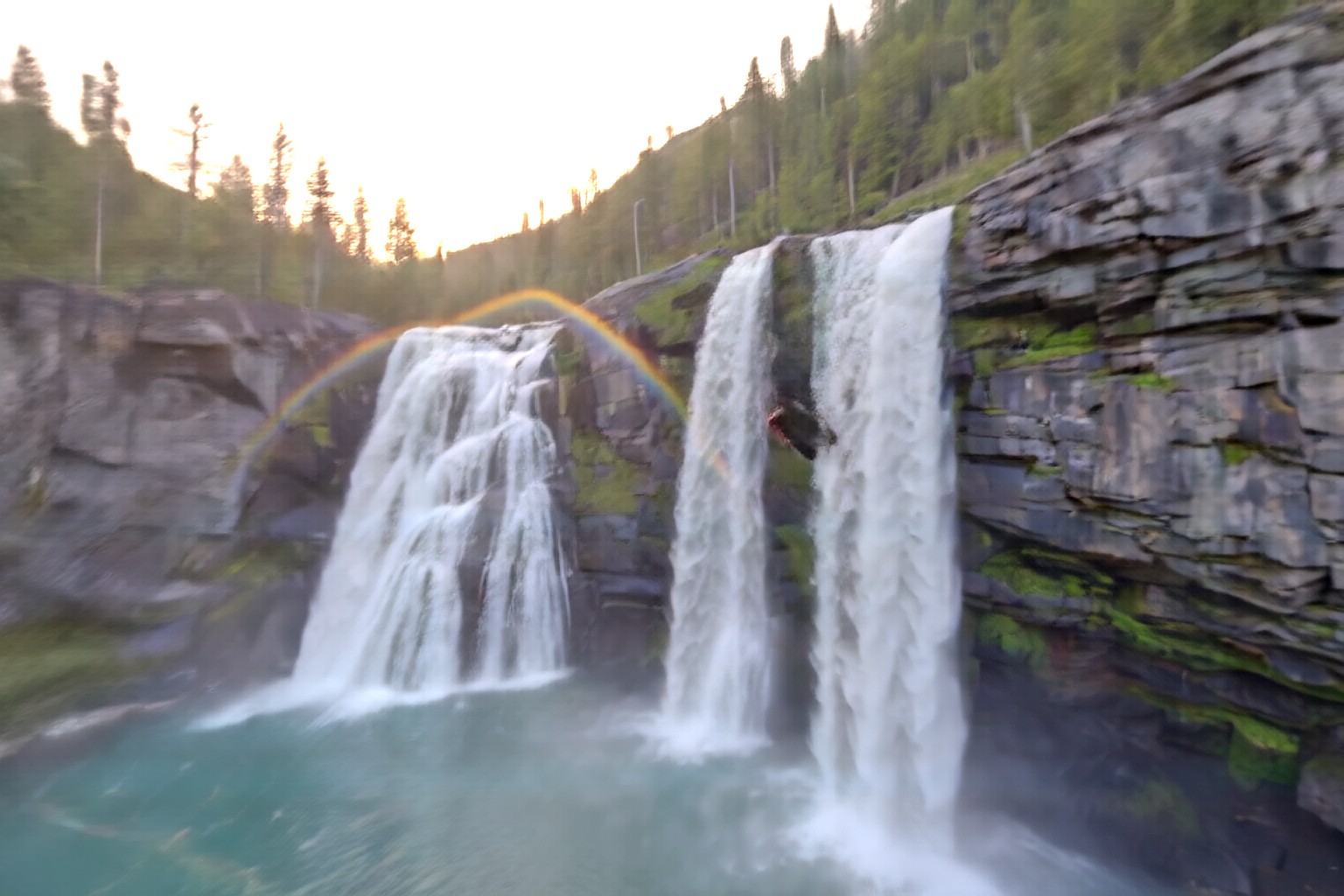} & \includegraphics[width=0.32\linewidth]{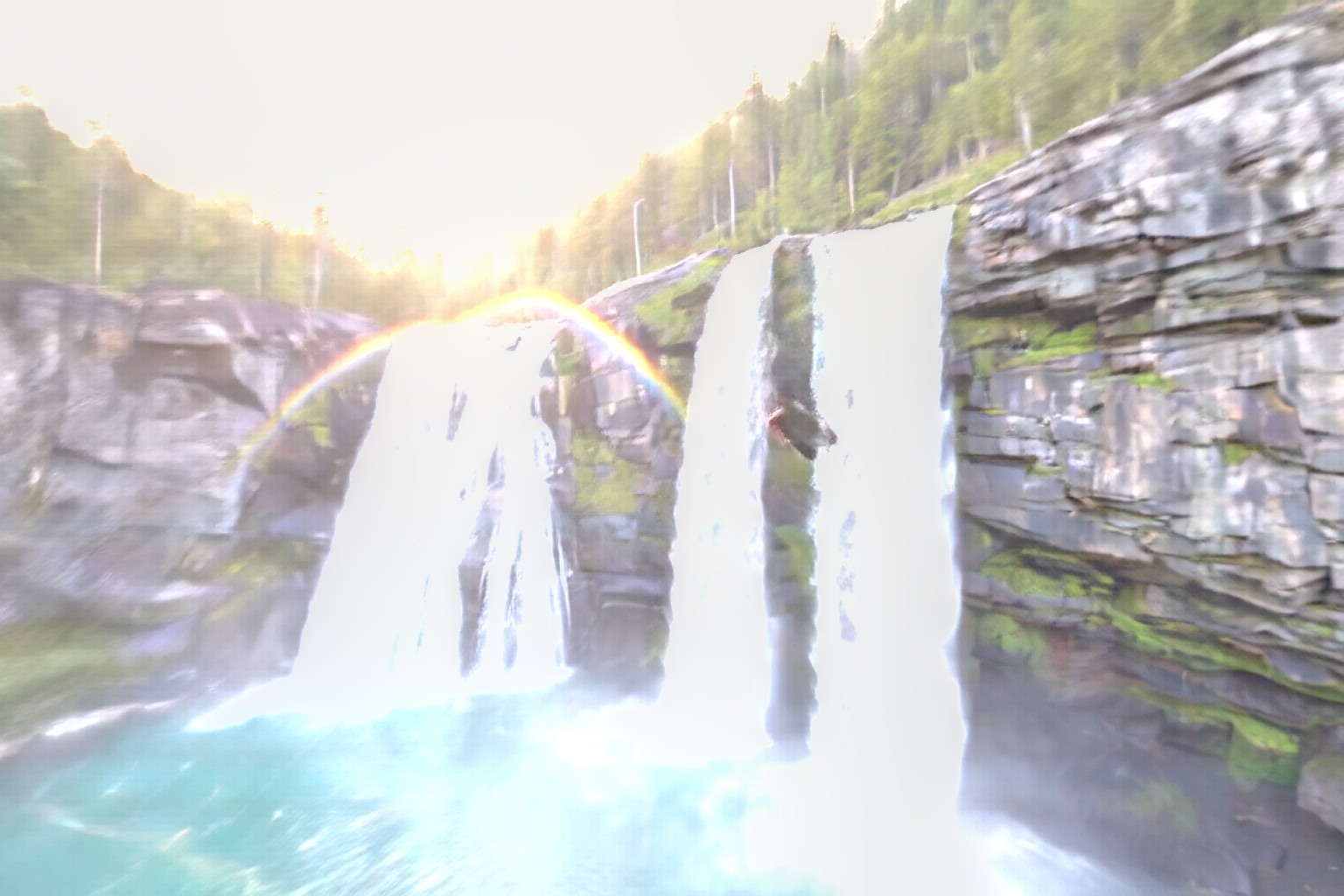} \\
        \raisebox{12pt}{\rotatebox{90}{X2HDR}} &
            \includegraphics[width=0.32\linewidth]{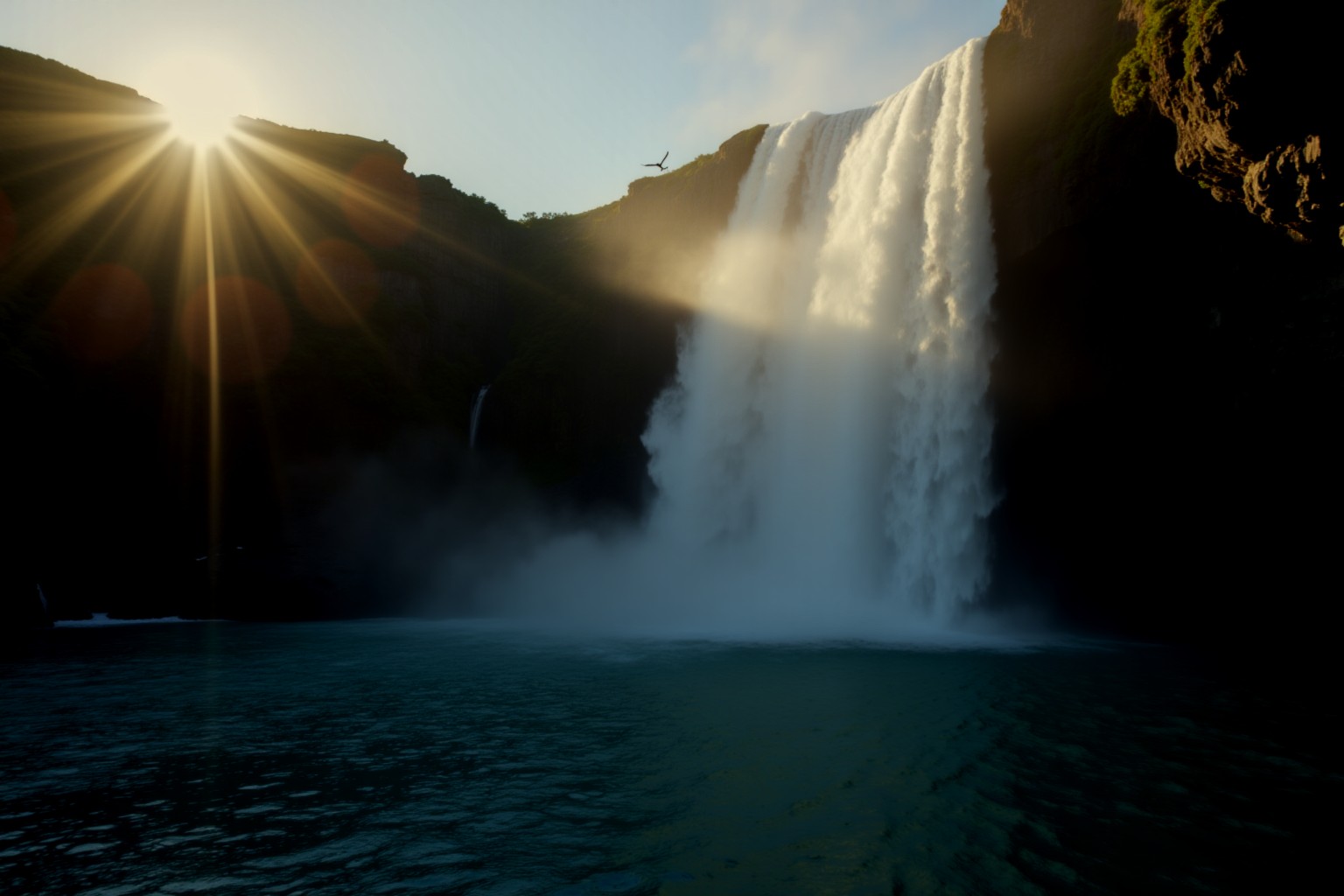} & \includegraphics[width=0.32\linewidth]{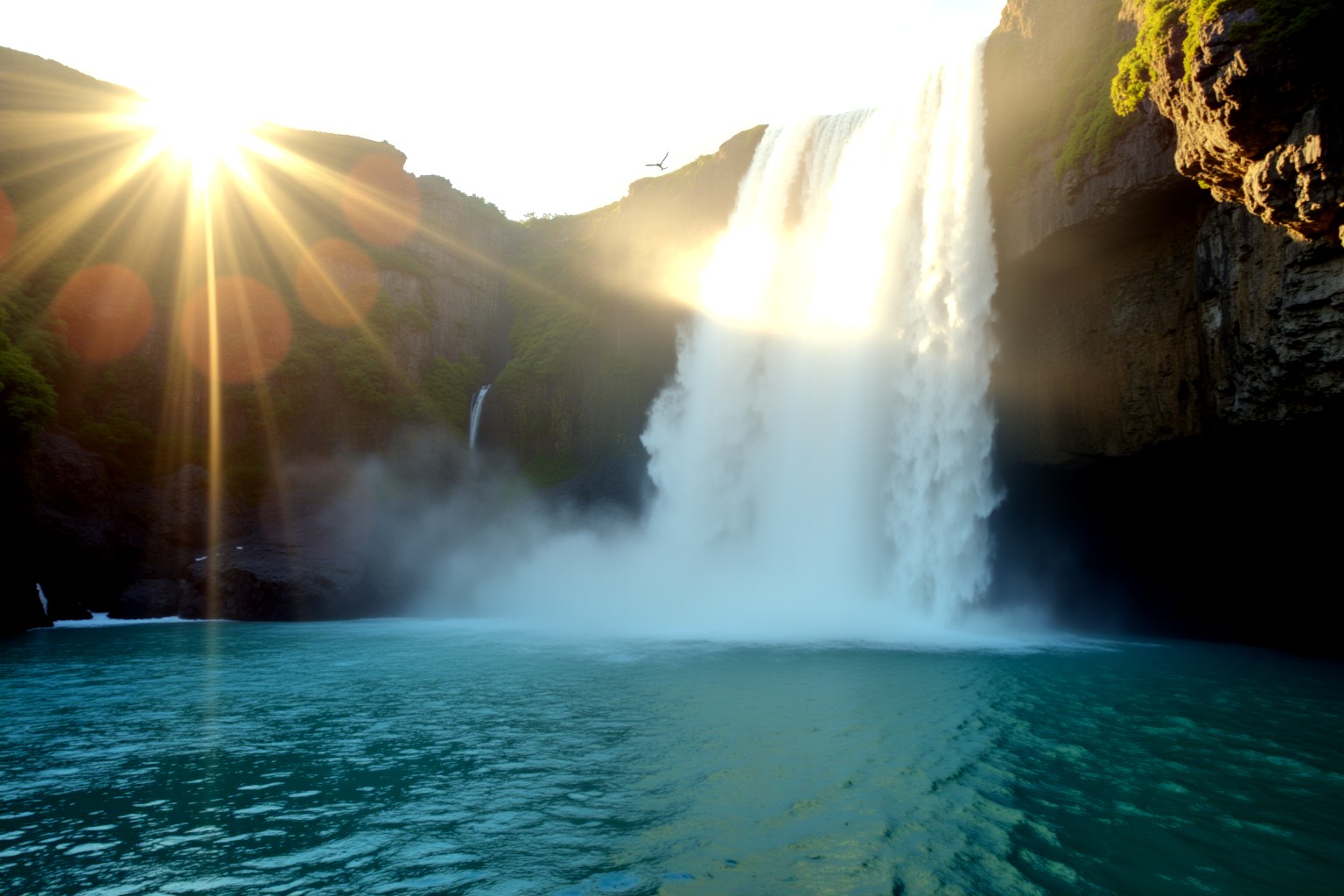} & \includegraphics[width=0.32\linewidth]{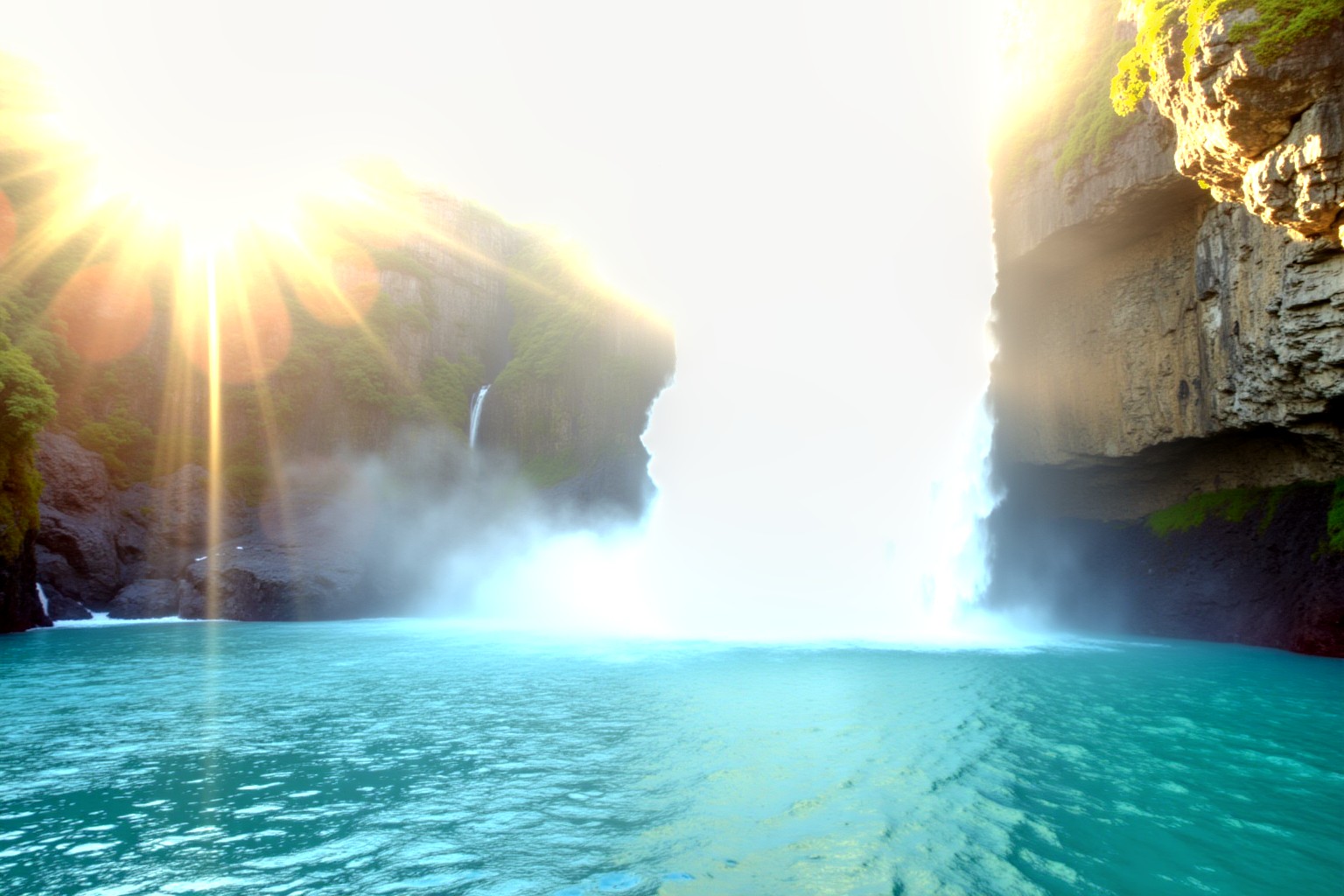} \\
        \raisebox{15pt}{\rotatebox{90}{LumiVid}} &
            \includegraphics[width=0.32\linewidth]{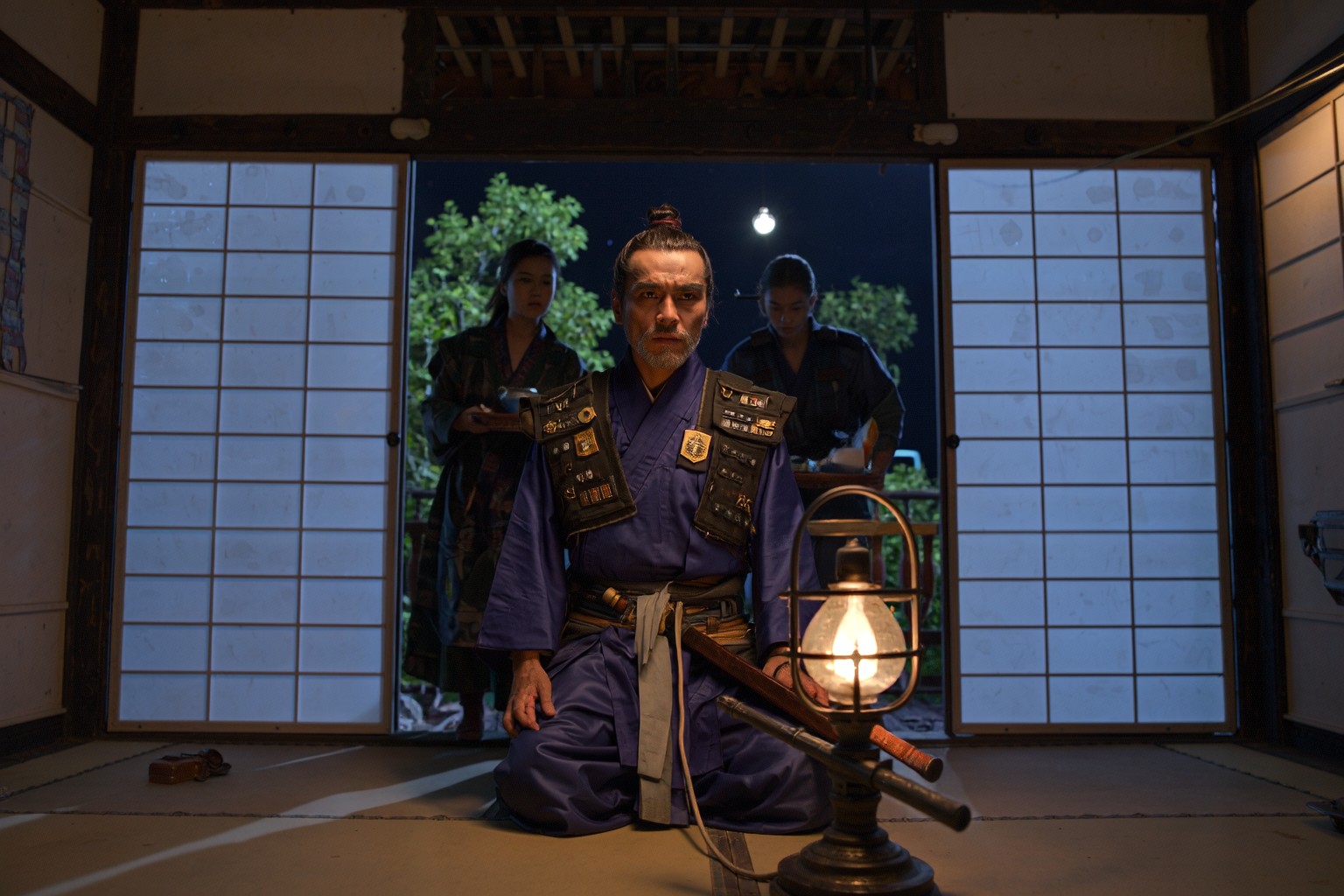} & \includegraphics[width=0.32\linewidth]{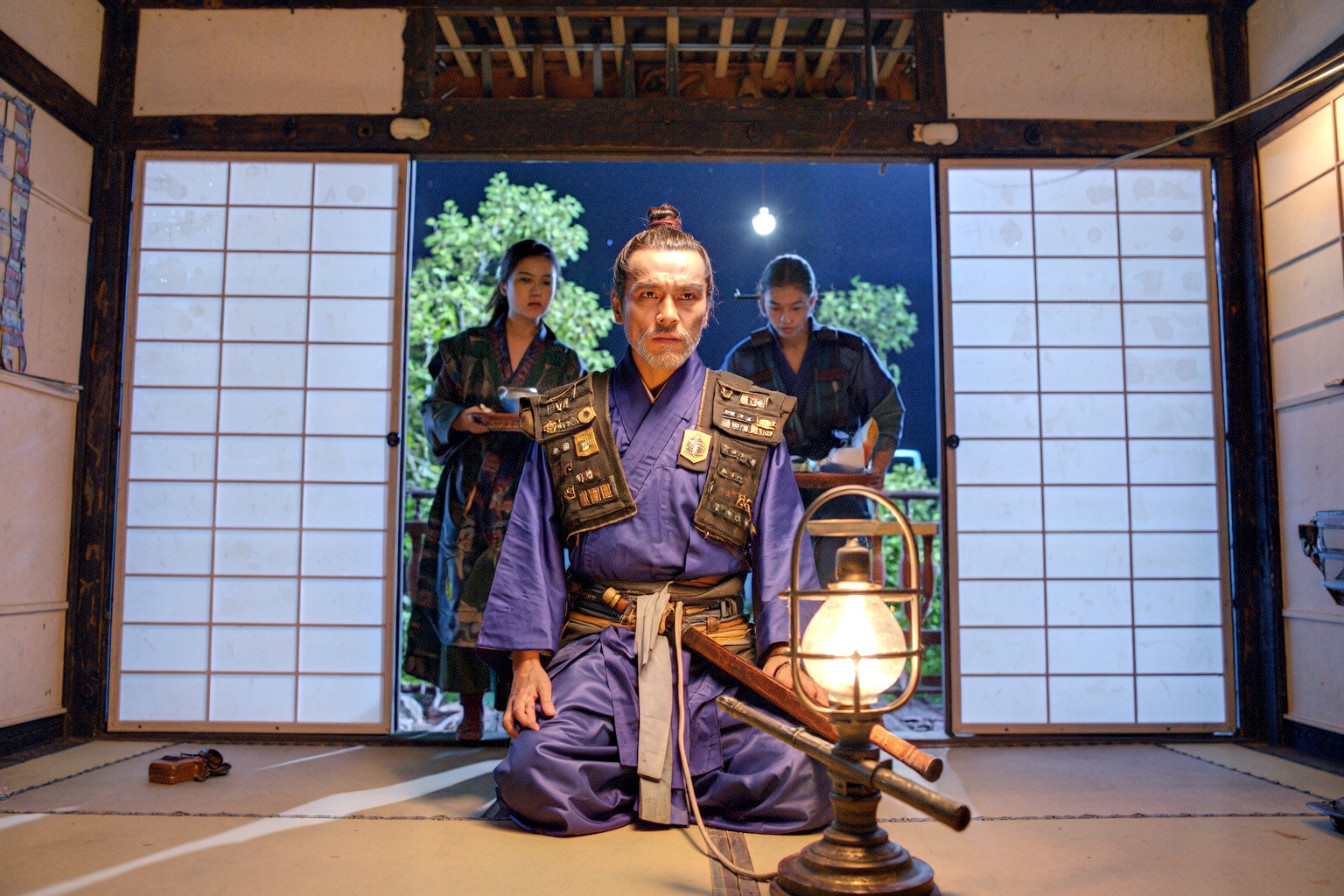} & \includegraphics[width=0.32\linewidth]{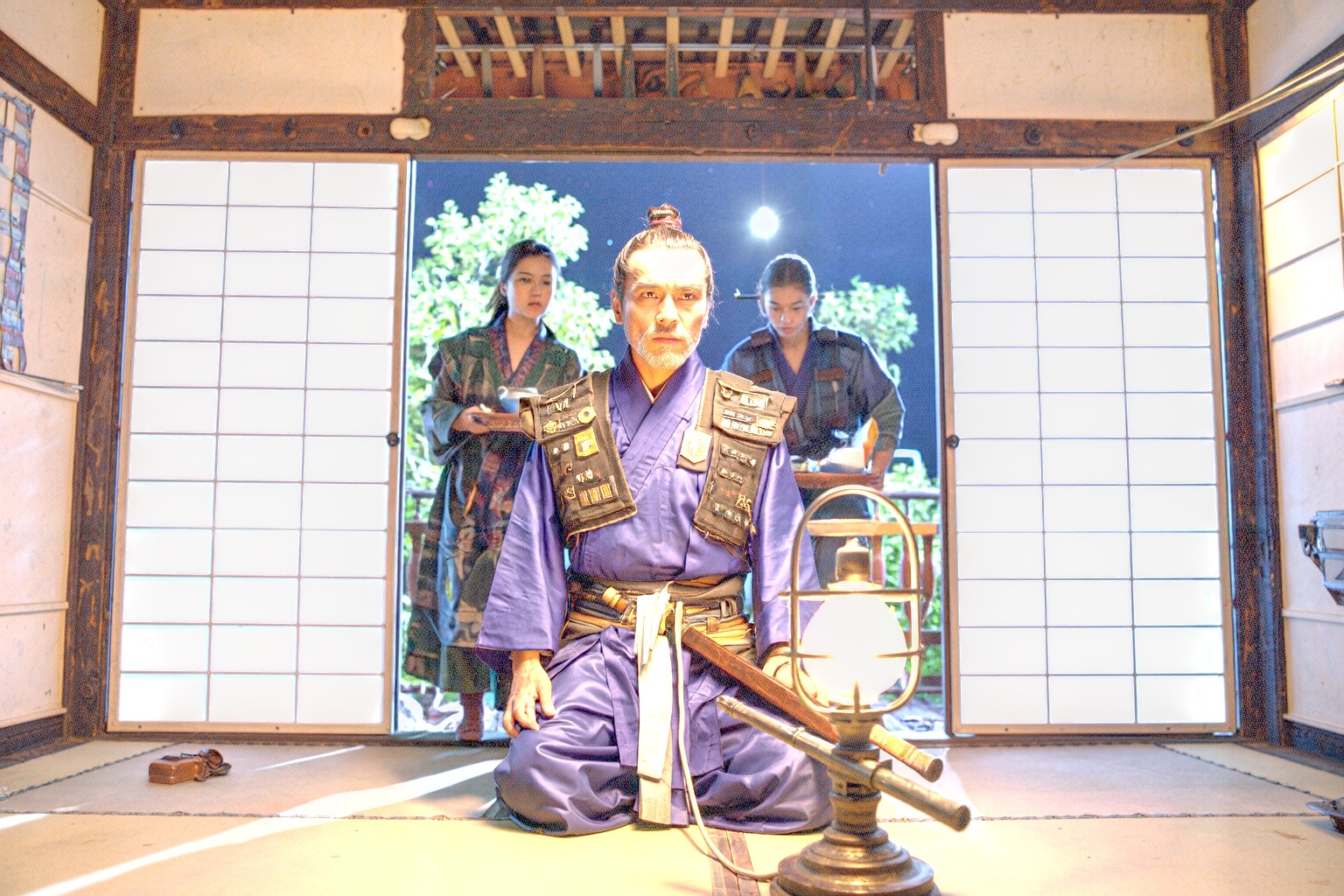} \\
        \raisebox{12pt}{\rotatebox{90}{X2HDR}} &
            \includegraphics[width=0.32\linewidth]{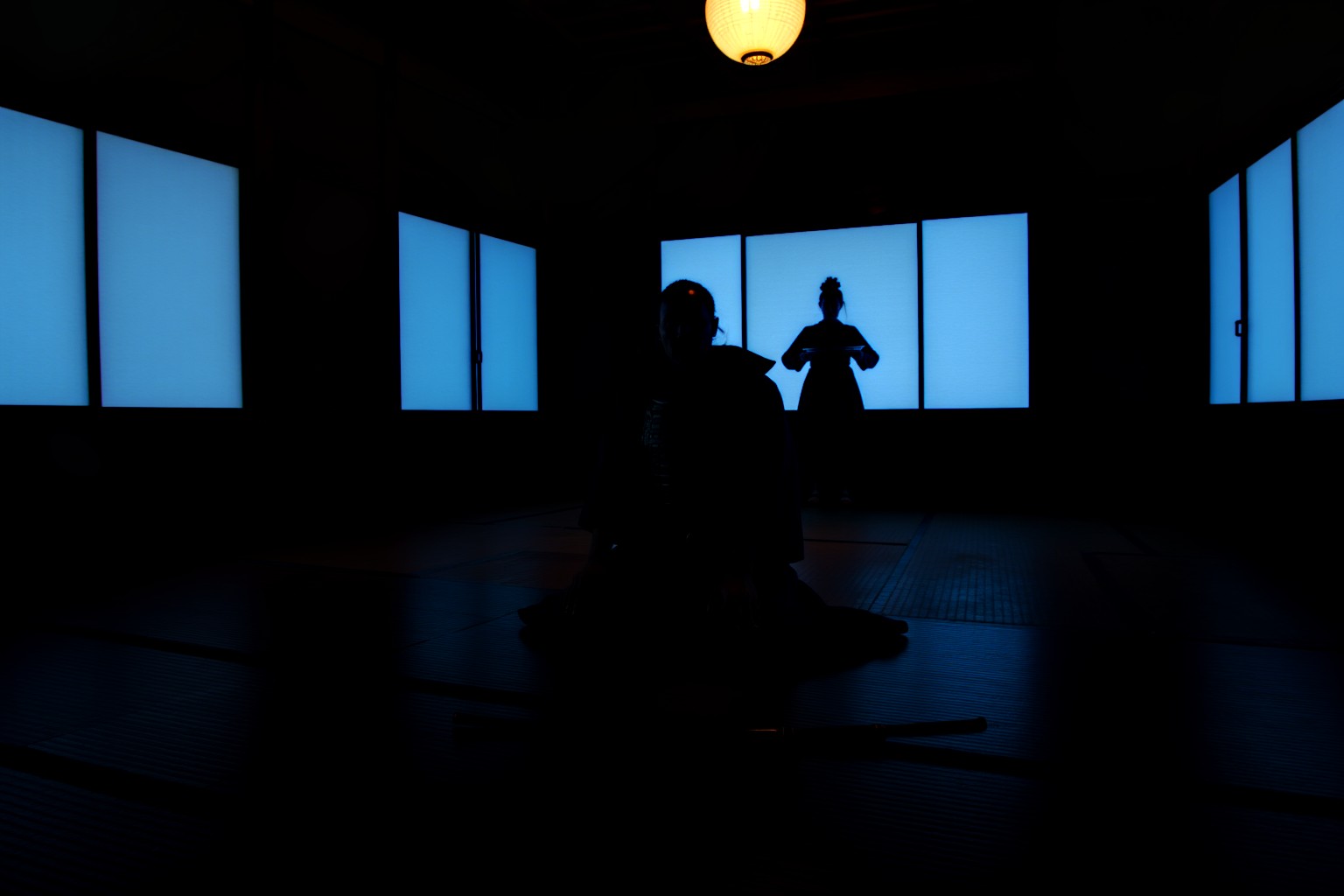} & \includegraphics[width=0.32\linewidth]{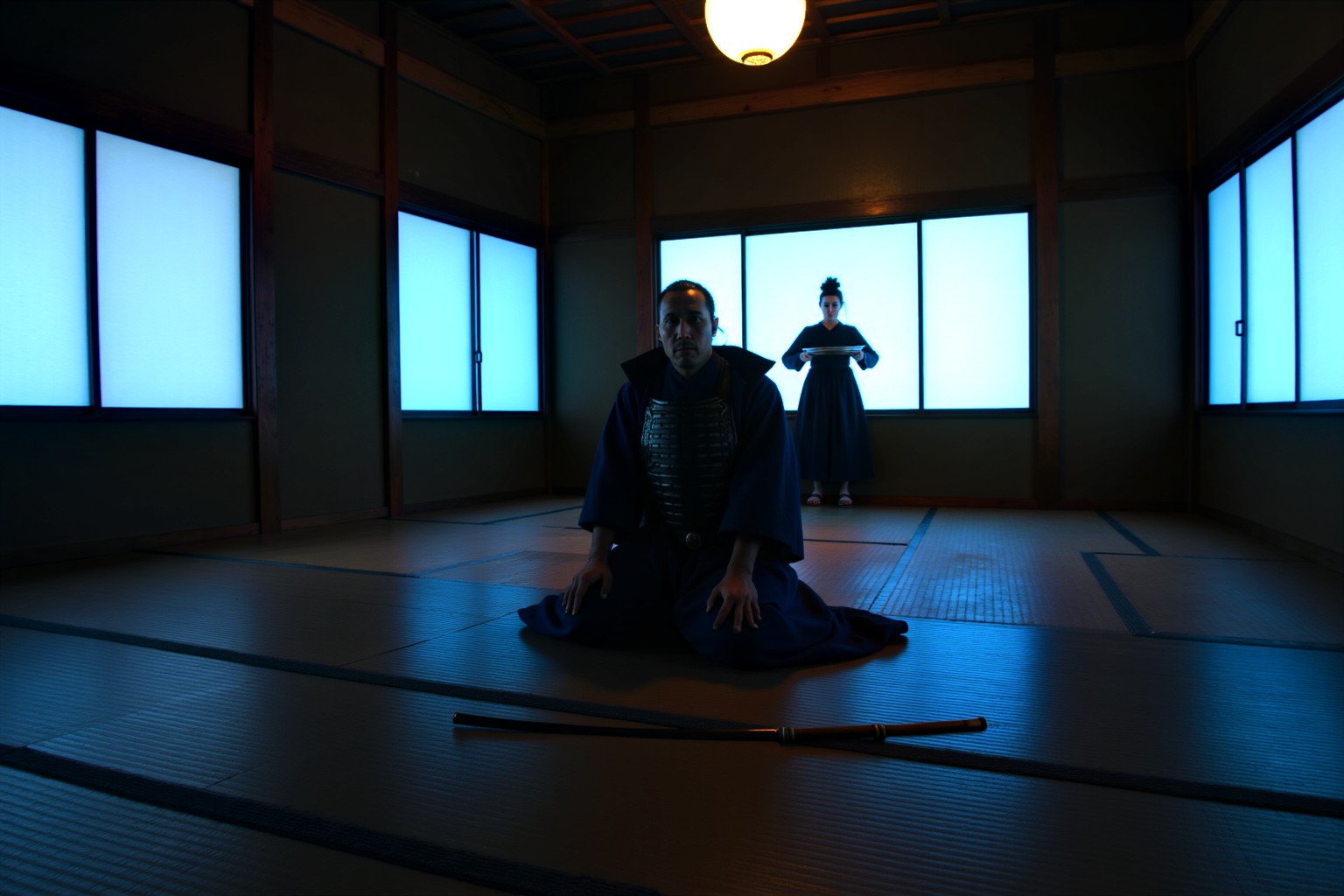} & \includegraphics[width=0.32\linewidth]{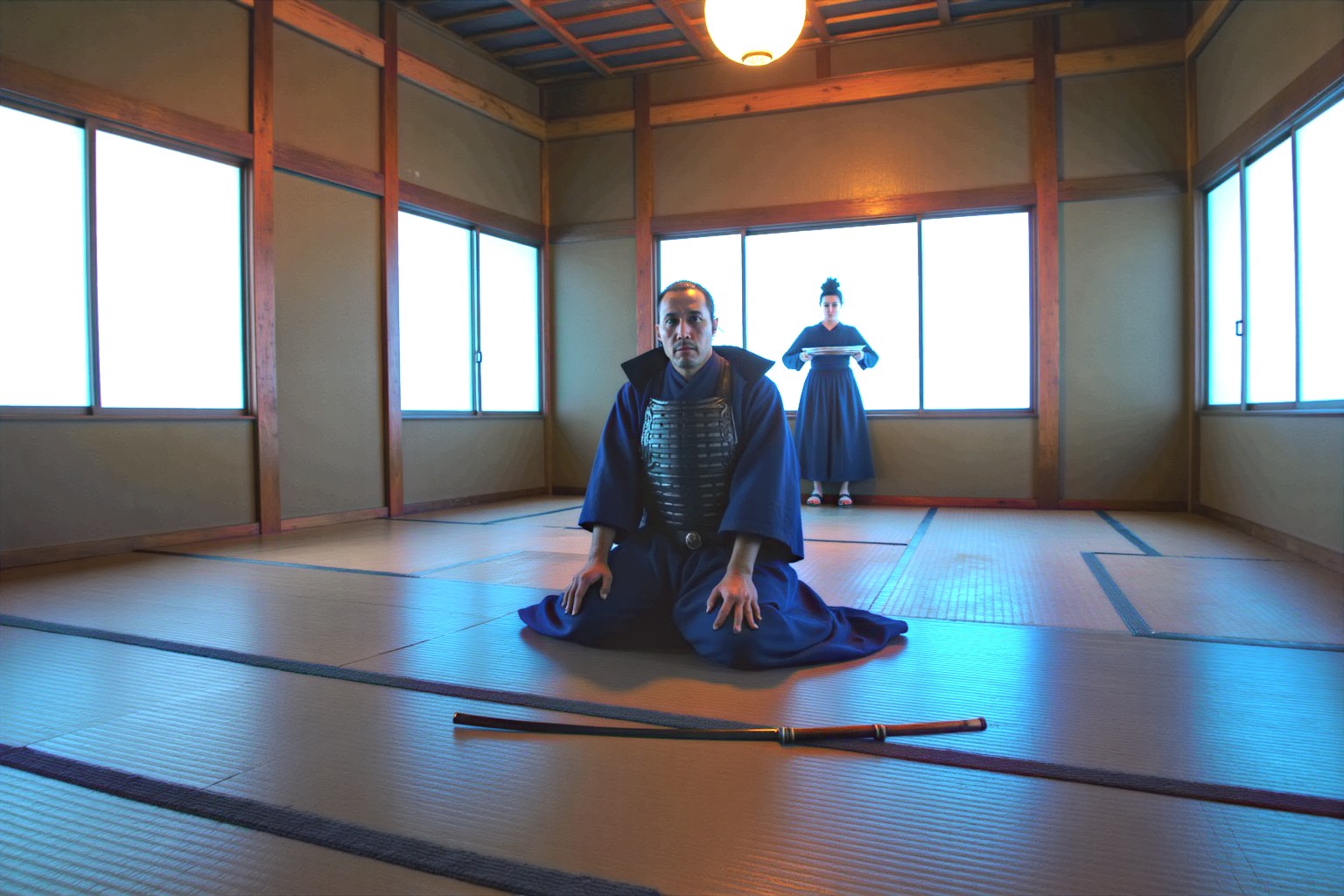} \\
        & EV $-4$ (highlights) & EV $0$ (base) & EV $+4$ (shadows)
    \end{tabular}
    \caption{%
        \textbf{Detail revealed under exposure shifts} text-to-HDR 
        representative scenes. For each scene we show LumiVid (top
        row) and X2HDR (bottom row) at three exposures: EV~$-4$
        (highlights revealed), EV~$0$ (base), and EV~$+4$ (shadows
        revealed). At EV~$+4$,
        LumiVid preserves real spatial structure in dark and highlight regions, while X2HDR's shadows and highlights
        collapse to a featureless plateau, brightening reveals less
        additional details. The
        EV~$-4$ column shows a consistent advantage in
        highlight texture (sky, clouds, lit windows).
    }
    \label{fig:shadow-detail-revealed}
\end{figure*}

Text-to-HDR uses the same LoRA with the SDR canvas dropped.
The model denoises ACEScct tokens from text alone; we invert $T$ to ACEScg EXR.
Figure~\ref{fig:shadow-detail-revealed} compares \method{} to X2HDR~\cite{x2hdr2025} in a text-toHDR mode,
at EV $-4$, $0$, and $+4$.
X2HDR is an image model, so each frame is independent.

\clearpage

\end{document}